\documentclass{article}

\usepackage{arxiv}

\usepackage[utf8x]{inputenc} 
\usepackage[T1]{fontenc}    
\usepackage{hyperref}       
\usepackage{url}            
\usepackage{booktabs}       
\usepackage{amsfonts}       
\usepackage{nicefrac}       
\usepackage{microtype}      
\usepackage{lipsum}
\usepackage{makeidx}         
\usepackage[bottom]{footmisc}

\usepackage{glossaries}
\usepackage[colorinlistoftodos,prependcaption,textsize=medium]{todonotes}
\usepackage[linesnumbered,ruled,lined,commentsnumbered]{algorithm2e}
\usepackage{multicol}
\usepackage{amsmath}
\usepackage{amssymb}
\usepackage{mathtools}
\usepackage{bm}
\usepackage{cleveref}
\usepackage{graphicx}
\usepackage{array}
\usepackage{multirow}
\usepackage{csquotes}
\usepackage{ctable}
\usepackage{subcaption}
\usepackage{float}
\usepackage{setspace}
\usepackage{gensymb}
\emergencystretch=\maxdimen
\newtoggle{CAMERA}
\toggletrue{CAMERA}
\newtoggle{Thesis}
\togglefalse{Thesis}
\newcommand{\erik}{\iftoggle{CAMERA}{ERIK}{AnonIK}}
\newcommand{\ejm}{\iftoggle{CAMERA}{EJM}{AJM}}

\newcommand{\urlDate}{(accessed January 12, 2019)}

\newcommand{\uvecX}{{\bm{\hat{X}}}}
\newcommand{\uvecY}{{\bm{\hat{Y}}}}
\newcommand{\uvecZ}{{\bm{\hat{Z}}}}

\newcommand{\figDir}{fig}

\title{Expressive Inverse Kinematics Solving in Real-time\\ for Virtual and Robotic Interactive Characters}

\author{
  Tiago Ribeiro\thanks{\protect\url{www.tiagoribeiro.pt}} \\
  INESC-ID \& \\Instituto Superior T\'{e}cnico\\
  University of Lisbon\\
  Portugal \\
  \texttt{me@tiagoribeiro.pt} \\
   \And
 Ana Paiva \\
  INESC-ID \& \\Instituto Superior T\'{e}cnico\\
  University of Lisbon\\
  Portugal \\
  \texttt{ana.paiva@inesc-id.pt} \\
}

\begin{document}
\maketitle

\begin{abstract}
With new advancements in interaction techniques, character animation also requires new methods, to support fields such as robotics, and VR/AR.
Interactive characters in such fields are becoming driven by AI which opens up the possibility of non-linear and open-ended narratives that may even include interaction with the real, physical world.
This paper presents and describes \erik, an expressive inverse kinematics technique aimed at such applications.
Our technique allows an arbitrary kinematic chain, such as an arm, snake, or robotic manipulator, to exhibit an expressive posture while aiming its end-point towards a given target orientation.
The technique runs in interactive-time and does not require any pre-processing step such as e.g. training in machine learning techniques, in order to support new embodiments or new postures.
That allows it to be integrated in an artist-friendly workflow, bringing artists closer to the development of such AI-driven expressive characters, by allowing them to use their typical animation tools of choice, and to properly pre-visualize the animation during design-time, even on a real robot.
The full algorithmic specification is presented and described so that it can be implemented and used throughout the communities of the various fields we address.
We demonstrate \erik\ on different virtual kinematic structures, and also on a low-fidelity robot that was crafted using wood and hobby-grade servos, to show how well the technique performs even on a low-grade robot.
Our evaluation shows how well the technique performs, i.e., how well the character is able to point at the target orientation, while minimally disrupting its target expressive posture, and respecting its mechanical rotation limits.

\end{abstract}


\newacronym{hri}{HRI}{Human-Robot Interaction}
\newacronym{cgi}{CGI}{Computer-Graphics}
\newacronym{ai}{AI}{artificial intelligence}
\newacronym{ik}{IK}{inverse kinematics}
\newacronym[longplural={Degrees of Freedom}]{dof}{DoF}{Degree of Freedom}

\newcommand{\sgn}{\text{sgn}}

\newcommand{\parallelVec}{\mathbin{\|}}

\newcommand{\abs}[1]{\lvert #1\rvert}
\newcommand{\norm}[1]{\lVert #1\rVert}
\newcommand{\proj}[2]{\text{proj}_{#1}#2}
\newcommand{\round}[1]{\lfloor #1\rceil}
\newcommand{\swap}[2]{\text{Swap}(#1,#2)}

\newcommand{\parent}[1]{{#1_{\text{Parent}}}}
\newcommand{\child}[1]{{#1_{\text{Child}}}}
\newcommand{\Root}[1]{{#1_{\text{Root}}}}
\newcommand{\Superpoint}[1]{{#1_{\text{SuperPoint}}}}
\newcommand{\EE}[1]{{{#1}_{\text{EE}}}}
\newcommand{\skEE}{{\Lambda}_{\Sk_{\text{EE}}}}
\newcommand{\Sk}[0]{{\text{Sk}}}
\newcommand{\RA}[1]{#1_{RA}}
\newcommand{\OA}[1]{#1_{OA}}
\newcommand{\POA}[1]{#1_{POA}}
\newcommand{\Eta}{\text{H}}

\newcommand{\isRoot}[1]{#1_{\text{IsRoot}}}
\newcommand{\isEE}[1]{#1_{\text{IsEndPoint}}}
\newcommand{\IsTwister}[1]{\text{IsTwister}(#1)}

\newcommand{\solGO}[1]{\Theta_{#1_Q}}
\newcommand{\solLO}[1]{\Theta_{#1_L}}
\newcommand{\solGOmat}[1]{\Theta_{#1_{Q_M}}}
\newcommand{\solAng}[1]{\Theta_{#1_\theta}}
\newcommand{\solAngEE}{\Theta_{EE_\theta}}
\newcommand{\solJOmat}[1]{\Theta_{#1_{\Omega_M}}}
\newcommand{\solJOaxis}[2]{\Theta_{#1_{\Omega_#2}}}
\newcommand{\solJD}[1]{\Theta_{{#1}_d}}
\newcommand{\solJO}[1]{\Theta_{#1_\Omega}}
\newcommand{\solPos}[1]{\Theta_{#1_\rho}}

\newcommand{\postAng}[1]{\Psi_{#1_\theta}}
\newcommand{\postPos}[1]{\Psi_{#1_{\vec{\rho}}}}
\newcommand{\postBasis}[1]{\Psi_{#1_Q}}

\newcommand{\segment}[1]{#1_{\vec{\sigma}}}
\newcommand{\segmentNorm}[1]{{#1}_{\hat{\sigma}}}
\newcommand{\JointDir}[1]{\Theta_{#1_d}}
\newcommand{\PitchRA}[1]{\text{PitchRA}(#1)}
\newcommand{\VAngle}[2]{\text{VecAngle}(#1,#2)}
\newcommand{\VAngleRA}[3]{\text{VecAngle}(#1,#2,#3)}

\newcommand{\TBasis}[2]{\text{TBasis}(#1,#2)}
\newcommand{\TBasisRoll}[2]{\text{TBasisRoll}(#1,#2)}
\newcommand{\QVRot}[2]{\text{VDiffAsQ}(#1,#2)}
\newcommand{\QDiff}[2]{\text{QDiff}(#1,#2)}
\newcommand{\QAA}[2]{\text{QAA}(#1,#2)}
\newcommand{\RotVQ}[2]{\text{RotVQ}(#1,#2)}

\newcommand{\EmptySolution}[1]{\text{EmptySolution}(#1)}
\newcommand{\SafeAngle}[2]{\text{SafeAngle}(#1, #2)}
\newcommand{\SafeAngleit}[2]{\text{SafeAngle}(#1, #2)}
\newcommand{\SafeAngleCycle}[3]{\text{SafeAngle}(#1, #2, \text{True})}
\newcommand{\SetOriFromParent}[2]{\text{SetOriFromParent}(#1, #2)}
\newcommand{\SetFrameFromParent}[2]{\text{SetFrameFromParent}(#1, #2)}
\newcommand{\ApplyFK}[2]{\text{ApplyFK}(#1, #2)}
\newcommand{\LALUT}[2]{\text{LALUT}(#1, #2)}
\newcommand{\LALUTit}[2]{\text{LALUT}(#1, #2)}
\newcommand{\TargetLatitude}[2]{\text{TargetLatitude}(#1, #2)}
\newcommand{\EPA}[2]{\text{EPA}(#1, #2)}
\newcommand{\AvoidJointEdge}[2]{\text{AvoidJointEdge}(#1, #2)}
\newcommand{\AvoidPostureJointEdges}[2]{\text{AvoidPostureJointEdges}(#1, #2)}
\newcommand{\NonConversionDetected}[2]{\text{NonConversionDetected}(#1, #2)}
\newcommand{\NonConvOffsetTrick}[2]{\text{NonConvOffsetTrick}(#1, #2)}
\newcommand{\SelectBestSolution}[1]{\text{SelectBestSolution}(#1)}
\section{Introduction}
When we think of animated characters, what immediately comes to our mind are the characters seen on TV and in movies. 
These characters were artistically crafted either by hand or using computer graphics (CGI) and design techniques, in order to convey the illusion that they are alive (e.g. \cite{thomas1995illusion,JohnCanemaker1996,Goldberg2008}). 
Currently however, we can find characters that \textit{live} out of the big screens, and are becoming more interactive, powered by artificial intelligence (AI), immersive media such as virtual reality (VR) \cite{Narang2016} and augmented reality (AR) \cite{Cimen2017}, and even through robots.

While the virtual characters used in VR and AR applications are more familiar to the animation community, social robots in particular are posing as a new form of animated characters. 
In order to be used in social applications such as education, entertainment, assisted living or in public spaces, they are endowed with the most powerful AI capabilities, and the most advanced robotics hardware technology.

However what is common to all applications of AI and social agents to physical and immersive characters, is the need to be expressive while also being fully interactive.
As a need, this is also what distinguishes these interactive characters, from the linear characters created for TV and movies \cite{Tomlinson2005}.
Such virtual and robotic animated characters must be able to respond to users in various situations and environments, without appearing stiff or pre-scripted, in order to become believable interactive characters, ones that convey the illusion of life \cite{Lasseter1987,Bates1994,Hoffman2014}.

Our goal is to turn what used to be regarded as a machine, into life-like animated characters, by focusing on the software technology and animation techniques that would allow a socially interactive AI to expressively control a complex, physical embodiment during an autonomous interaction with human users.
As such this paper places a stronger focus on robot animation, while linking to and from CGI character animation.
In particular, our long-term goal is to connect the world of character animation and robotics, and to allow animation artists to take a central role in the development of social robots, by providing expert body knowledge that can be used by an AI-driven character.
On that topic, we have previously described many of the \textit{Nutty-based robot animation} theories and practices \cite{ribeiro2019nutty}.
Please refer to that reference for more details on our holistic view and approach regarding robot animation.

Through this paper, we do contribute to our goal with a novel inverse kinematics technique that allows interactive expressive characters, either virtual or robotic, to convey an expressive posture while facing towards any given direction (e.g. as when gazing an object, or tracking a person).
Our contribution is expected to establish new boundaries in the fields we address, by allowing such interactive characters to become more expressive and less \textit{scripted}, thus becoming more believable, and capable of exhibiting the illusion of life.

While this paper focuses on robot animation, we consider that our approach, as being heavily based on CGI character animation, could further be applied to other types of interactive characters in which immersion and interaction with the physical users and their world is key, such as in VR and AR applications.

\subsection{Expressive Interactive Characters and Robots}

Animating autonomous social robots introduces concerns that are not typically addressed by CGI animation. 
However CGI animation has already solved many problems and introduced many techniques that allow artists to remain central to the character and animation design process.
Therefore we have been looking both at techniques from CGI character animation and from robotics and human-robot interaction (HRI), in order to understand how these worlds can be bridged together.
By addressing both worlds (virtual and robotic), we expect to uncover new paradigms that will allow such animated characters to become fully expressive and seamlessly interactive regardless of the form or physicality of its embodiment.

One of the basic expressive behaviours expected from an interactive character is the ability to face towards people or particular directions, either using its face when one exists (gazing), or using its body, or part of it (e.g. pointing using an arm, a tail), or the whole body if e.g. the character is a snake or a worm.
In social robots in particular, this behaviour is cornerstone to the illusion of life, as it is through gazing that the robot is able to convey the illusion of thought, and of being aware of others, itself, and of its surrounding environment \cite{Takayama2011,Ribeiro2012,pereira2014,Hoffman2014}.

In order to implement expressive gazing behaviour, such socially interactive robots generally contain either very simple expressive traits, or anthropomorphic features that allow the gaze-blending process to become quite straightforward, by containing specific degrees of freedom (DoFs), such as a neck, which can be dedicated to such behaviour \cite{Fong2003,Hoffman2014,Ribeiro2013,Ribeiro2016,jibo}.
For more complex and articulated embodiments, one could use inverse kinematics (IK), however the existing techniques focus on specifying a target position and/or orientation for the end-point, and do not care for the expressivity of the pose that results from the IK solution.
In other cases where some level of control over the posture is permitted, the techniques require full motion planning before being executed, which hinders their degree of interactivity (e.g. \cite{Dragan2015})

The problem of conveying an expressive posture towards a given target orientation is complex because the whole embodiment must be used to shape the posture (designed by an animator), while simultaneously using the same degrees of freedom to warp that posture towards a given orientation (e.g. gaze-tracking direction), while disrupting the posture as little as possible, so that the character conveys the same emotion or gesture as the animator had initially intended.
Furthermore such a problem must be solved in real-time for interactive use, and the resulting postures should allow to provide a smooth and natural motion while solving through different continuously changing postures and orientation.
To add up to the problem, we also argue that the solution to the problem should work without going through major tweaking or training through examples or demonstrations, which could heavily embarrass and deprive the creative process.
As such, the solution must be able to do its best to take what the animator wanted to convey, and use it interactively with minimal disruption.


This section presents and describes \erik, a heuristic inverse kinematics technique that allows a virtual or robotic character with an arbitrary articulated embodiment to convey and hold a given expressive posture while facing a given direction during an interaction.
It is able to provide many solutions per second using a standard computer, allowing it to be used for interactive applications.

The effort to design expressive behaviours for interactive characters using \erik\ is also minimal.
Animators can design single front-facing postures for any given embodiment, which are used as input to the algorithm, with no pre-computing or offline training required for any new posture or embodiment.
The algorithm is then able to take that posture and warp it in real-time, so that a given end-point is facing any given orientation, while respecting the embodiment's kinematic constraints, and while attempting as best to hold the overall shape of the posture.

Furthermore, \erik\ was also developed to support its use with robots.
As such, its output consists of a list of rotation angles, one for each joint, which can be used either in virtual or robotic applications.
The solver computes on a \textit{per-frame} basis in order to easily fit into a typical animation cycle, i.e., it produces one full-body solution at a time, and not a pre-planned motion trajectory.

\begin{figure}[!htbp]
	\centering
	\includegraphics[width=0.6\columnwidth]{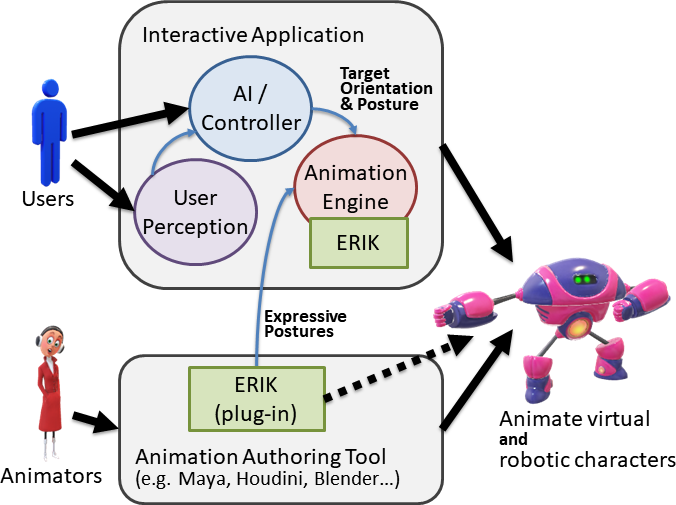}
	\caption[The \erik\ workflow, illustrated as a particular version of the Nutty Pipeline]{The \erik\ workflow, illustrated as a particular version of the Nutty Pipeline \ifboolexpr{togl {Thesis}}{}{\cite{ribeiro2019nutty}}.
	Animators can create expressive postures using typical animation tools. 
	Those expressive postures can be selected by an AI or character controller in an Interactive Application to drive an animation engine which, through the use of \erik, is able to perform the selected posture towards a selected orientation (e.g. from a user-perception component). 
	Alternatively (dotted arrow), ERIK can also be used by a plug-in for the animation tool, to allow live authoring of postures on real robots i.e., the animators are able to test the result of the postures in a given robot directly and in real-time.}~\label{fig:ERIKWorkflow}
\end{figure}

\ifboolexpr{togl {Thesis}}{
	\textbf{Figure \ref{fig:ERIKWorkflow}} illustrates the work-flow of a Nutty-ERIK system.
}{
	\textbf{Figure \ref{fig:ERIKWorkflow}} illustrates the work-flow of a Nutty-ERIK system \cite{ribeiro2019nutty}.
}
\erik\ can either be used as a component of an Interactive Application such as a game, VR/AR application, or a robotic AI, or alternatively, it can be used as a plug-in for an animation authoring tool.
In the latter case, due to its real-time nature, it allows artists to creatively explore the design of expressive postures for real robots in real-time, and directly in the real, physical embodiments.
This animator-inclusive workflow follows on the work initially proposed by \cite{Ribeiro2012,Ribeiro2013}.

\erik\ can be used to create tools and character animation engines directed at animation artists, so that they can take a stronger role in the development of autonomous, interactive, computer-animated characters, be them virtual or robotic.
By bringing the artists closer to the AI - or the AI closer to the artists - we expect \erik\ to prove as a strong technological contribution for the creation of better and more life-like interactive characters, in particular within immersive and emergent applications such as the ones based on VR, AR and robotics.

In many cases, the algorithm will be solving for two constraints that are not simultaneously satisfiable: the expressive posture of the character, i.e. the configuration of angles for each DoF that results in the given posture; and the global orientation of the end-point node, i.e. the configuration of angles for each DoF such that the end-point node faces towards a given orientation (in world coordinates).
This means that depending on the character's embodiment, on the target posture, and on the target orientation, the resulting pose may either fully satisfy both goals, or fully satisfy the orientation goal while partially satisfying the posture goal.
Due to mechanical limitations, there will be many cases in which it is physically impossibly to solve both goals.
Given that \erik\ aims at autonomous characters that interact with humans, it will prefer a pose that complies with the given orientation target (where a human is expected to be), while allowing the posture to fall short of the target expression.
This design decision makes \erik\ most appropriate for situations in which the characters perform a merely expressive role, where the control of its embodiment is not crucial for safety or successful completion of tasks, such as robotic manipulation of real objects.

From our evaluation, we found that \erik\ does in fact solve most cases successfully for both the posture and orientation goals, as long as the embodiment contains enough DoFs to achieve it.

\section{Related Work}
\erik\ was built based on existing models, techniques and ideas from the fields of robotics, HRI, interactive virtual agents (IVAs) and from CGI animation.
This sections presents not all of them, but the ones we found most relevant for grounding work and inspiration.
\subsection{Robotics and HRI}
The AUR is a robotic desk lamp with 5 DoFs and an LED lamp which can illuminate in a range of the RGB colour space \cite{Hoffman2008}. 
It is mounted on a workbench and controlled through a hybrid control system that allows it to be used for live puppeteering, in order to allow the robot to be expressive while also being responsive. 
In AUR, the motion is controlled by extensively trained puppeteers, and was composed through several layers. 
The bottom-most layer moves each DoF based on a pre-designed animation that was made specifically for the scene of the play. 
If the robot was set to establish eye contact, several specific DoFs would be overridden by an ik solution using CCD. 
A final \textit{animacy} layer added smoothed sinusoidal noise, akin to breathing, to all the DoFs, in order to provide a more lifelike motion to the robot.

Weinberg has dedicated to the creation of robotic musical companions, such as Shimon, a gesture based musical improvisation robot created along with Hoffman \cite{HoffmanWeinberg2010}. Shimon plays a real marimba. 
Its behaviour is a mix between its functionality as a musician, and being part of a band.
Using its head, it is able to perform expressive behaviour by gazing towards its band mates during the performance.

Various interactive social robots have been created at MIT's MediaLab \cite{Gray2010}, in particular the AIDA\footnote{\protect\url{http://robotic.media.mit.edu/portfolio/aida} \urlDate}, which is a friendly driving assistant for the cars of the future. 
AIDA interestingly delivers an expressive face on top of an articulated neck-like structure to allow to it move and be expressive on a car's dashboard.

Walt\footnote{\protect\href{http://robovision.be/offer/\#airobots}{http://robovision.be/offer/\#airobots} \urlDate} is a social collaborative robot that that helps factory workers assemble cars. 
Walt uses a screen to exhibit an expressive face, icons or short animations. 
Its body is a concealed articulated structure that allows it to gaze around at its co-workers.

The use of animation principles was explored by Takayama et al. using the PR-2\footnote{\protect\url{http://www.willowgarage.com/pages/pr2/overview} \urlDate} robot \cite{Takayama2011}.
This is a large mobile robot with two arms, that can navigate in a human environment. 
A professional animator from Pixar Animation Studios collaborated on the design of the expressive behaviour so that the robot could exhibit a sense of \textit{thought}, by clearly demonstrating the intention of its actions.
\textit{Thought} and \textit{Intention} are two concepts that are central in character animation, and in the portrayal of the illusion of life.
In this work, the authors argue for the need of both functional and expressive behaviours, i.e., that some of the robot's behaviours would be related with accomplishing a given task (e.g. picking up an object; opening a door), and that another part would concern its expressiveness in order to convey thought and emotion.

Nutty Tracks \cite{Ribeiro2013} is an animation engine and pipeline, aimed at providing an expressive bridge between an application-specific artificial intelligence (AI), the perception of user and environment, and a physical, animated embodiment.
It is able to combine and blend multi-modal expressions such as gazing towards users, while performing pre-designed animations, or overlaying expressive postures over the idle- and gazing- behaviour of a robot\footnote{\protect\url{http://vimeo.com/67197221} \urlDate}. 
This has allowed the researchers to explore the use of animation principles in such autonomous interactions with humans by focusing, however, on the behaviour selection and management mechanisms, and on pre-designing particular animations that were solely selected and played back on the robots. 
Nutty Tracks therefore stands as an animation engine that is aimed at interactive social robots, which however, through the use of simple animation blending, lacks the ability to compute expression for more complex embodiments, i.e., when techniques such as IK are required.

The challenge of providing legible and predictable motion to autonomous collaborative robots has also been addressed by Dragan et al., which demonstrates the benefits of including such properties into motion planners\cite{Dragan2015}.
Their technique however focuses on these two properties in particular, which become embedded into the planner through a more mathematical solution that does not provide an actual free-form expressive control.

\subsection{Virtual Characters and Computer Animation}

The challenge of dynamically generating expressive and meaningful behaviour in the field of virtual characters is not new.
Most work devoted to bridging AI and virtually animated characters focuses on the animation of human figures (e.g. \cite{Badler1997}), 
and is based on standardizing behaviours through markup language specifications that allow such characters to either be pre-scripted,
or to have their scripted behaviour generated by an AI during interaction.

This method is widespread across the field of IVAs, with authors focusing on how to specify and generate e.g. emotions \cite{Reilly1996}, gestures \cite{Cassell2001}, facial expressions \cite{Carolis2004, Niewiadomski2009} or even more general full-body behaviour \cite{PerlinGoldberg1996, Prendinger2004, Kopp2006, Heloir2009}.
By relying heavily on scripting however, it makes it difficult to generalize behaviour to any type of embodiment (most are specifically made for virtual humans), and also makes it difficult to specify certain \textit{nuances} for particular characters and situations.
This is because most of the behaviour tags are directed at specifying particular communicative behaviours used by humans such as hand gestures, head movement, or facial expressions based on FACS \cite{Ekman1978} or FAPs \cite{Pandzic2003}.

Some works do focus more on motion generation, in order to generate appropriate e.g. gestures in a more procedural way.

For the video-game \textit{Spore}\footnote{\protect\url{https://www.spore.com/} \urlDate}, the authors had to create an innovative animation system that allowed creatures with custom morphologies to behave coherently by performing locomotion and animated actions in a procedurally generated world \cite{Hecker2008}.
They preserved a traditional animation workflow so that artists could take a central role in the development process.
The \textit{Spore} engine is aimed specifically at the types of creatures used in the game, which contain leg groups and arm groups, and perform a set of pre-determined actions.
The piece of this work that became most relevant and inspiring to us is the use of the Particle IK Solver, which will be described in the next subsection.

Smartbody is a popular procedural animation system in the virtual humans field \cite{Thiebaux2008}. 
It takes a BML \cite{Kopp2006} specification of behaviour as input in order to control any type of embodied agent. 
This behaviour is scheduled and executed in several motion controllers, which are combined in each frame to generate a set of skeletal joint rotations and translations. 
Smartbody procedurally generates and adapts gestures using an example-based motion synthesis technique for locomotion, reach and object manipulation \cite{Feng2012}.
That technique takes a large set of example postures of a given embodiment e.g. reaching towards different directions, and is then able to produce a grasping pose for any direction by blending the previously authored examples.
This approach was very inspiring to us, as its end-result seems similar to what we would like to achieve in the field of robot animation, however we want to remove the need to provide a large set of example postures from the workflow.

\subsection{Forward and Inverse Kinematics}

\begin{figure}[h]
	\centering
	\includegraphics[width=0.5\columnwidth]{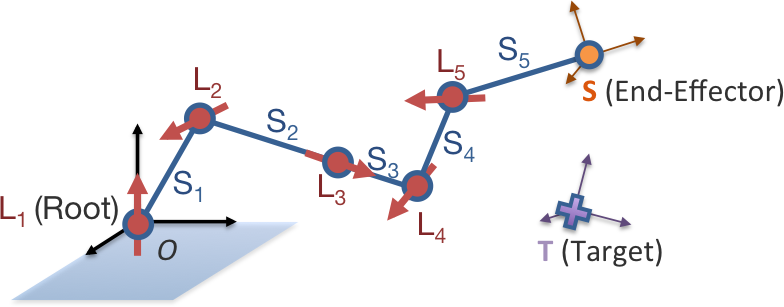}
	\caption{An articulated structure (kinematic chain) as used in both Forward Kinematics (FK) and Inverse Kinematics (IK). Also shown is a given target T that is to be reached by the end-point S.}
	\label{fig:IK}
\end{figure}

In general, computing motion of an articulated structure is done through Forward Kinematics (FK) and Inverse Kinematics (IK). Here we briefly introduce some fundamental concepts and techniques regarding these processes. Figure \ref{fig:IK} provides a visual guide on each of the elements that compose an FK/IK problem. 
The kinematic chain is a sequence of segments connected through links, starting at a segment $S_1$ at the Origin ($O$) of the world-frame through link $L_1$ (or $L_{\text{Root}}$). 
The tip of the last segment is designated as the \textit{End-Point} (or End-Effector). 
Each link $L_i$'s world-space location is $P_i$, and it allows for a rotation $\alpha$ about an arbitrary axis $R_i!=\vec{0}$ with angular limits such that \mbox{$\text{min}_{\alpha_i} \le \alpha \le \text{max}_{\alpha_i}$}.
A \textit{posture} is a set of world-space positions \mbox{${P_1, ..., P_N}$} for each joint. A \textit{Kinematic Solution (KS)} is a set of angles \mbox{${\alpha_1, ..., \alpha_N}$} to be applied to each link \mbox{${L_1, ..., L_N}$}. 
FK allows to compute the final \textit{Posture} achieved from a given \textit{Kinematic Solution}, while IK computes the \textit{Kinematic Solution} that allows the end-point $S$ to achieve a target transform $T$, containing an orientational component $T_{ori}$ and/or a world-space position component $T_{pos}$. 

A comprehensive summary of the most popular IK techniques has already been gathered by Aristidou et al. \cite{Aristidou2018} and is recommended to the interested reader. 
Given that the latter one is recent and already describes nearly every option of inverse kinematics up to date, we will refrain from extending this literature section beyond the bare minimum.
As such we describe here only the techniques that are central to our contribution in addition to the basic Jacobian methods, while solely providing a mention to various other relevant techniques such as \cite{Boulic1996,Kulpa2005,CourtyArnaud2008,Hecker2008,Unzueta2008,Harish2016,Starke2016}.

\label{sec:related_ik_jacobian}
While in general, the IK problem is highly non-linear, the Jacobian methods provide linear approximations to it. 
They are based on the computation and inversion of the Jacobian matrix which contains the partial derivatives of the entire chain system, relative to the end-effectors. 

\begin{figure}
	\centering
	\includegraphics[width=0.7\columnwidth]{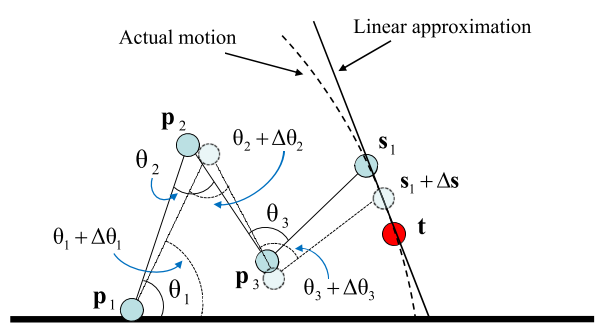}
	\caption[The Jacobian solution as a linear approximation of the actual motion of the kinematic chain.]{The Jacobian solution as a linear approximation of the actual motion of the kinematic chain. Description and image cited verbatim from \protect\cite{Aristidou2018}. }
	\label{fig:ik_jacobian}
\end{figure}

An extensive explanation of these methods is provided by Buss \cite{Buss2009} and should be consulted for more details.
The problem is illustrated by Figure \ref{fig:ik_jacobian}.
In simple terms, given the current position and/or orientation $\vec{s}$ (i.e. \textit{transform}) of an end-effector, and a target position and/or orientation $\vec{t}$ that it should achieve, let $\vec{e} = \vec{t} - \vec{s}$ represent the error vector (or \textit{task}) between the end-effector and the desired target values, 
and $\theta = (\theta_1,...,\theta_n)^T$, the current joint angles of the system, having $n$ as the number of joints.
The value $m$ will be the dimension of $\vec{e}$ (and consequently, of both $\vec{s}$ and $\vec{t}$) and will depend on the target IK task.
If the task is e.g. the 3D position constraint or 3D orientation constraint of a single end-effector then $m=3$.
If it is to control both the 3D position and 3D orientation, then $m=6$.
However one might choose a task that controls the orientation of only two of the rotation axes, in which case $m=2$.
Alternatively, one might also require to set one end-point to a given XY position, regardless of its position in the Z axis; in that case $m$ would also be $2$.

Note that for position control, the task is directly calculated as $\vec{e} = \vec{t} - \vec{s}$, while for orientation control many parametrizations exist.
When using Euler angles, one option is to calculate it the same way, i.e., solving at the differential level, by using the desired angular velocity vector.
When using quaternions, we may take the vector part of the target quaternion orientation $q_{\vec{v}}$, and use this vector as the task.

The Jacobian matrix $J$ of size $m\times n$ (rows$\times$ columns) is a function of the current $\theta$ values defined by 
\begin{equation}
\label{eq:jacobian_def_partial}
J(\theta) = \bigg(\frac{\partial{s_i}}{\partial{\theta_j}}\bigg)_{i,j}
\end{equation} 

It will result in a matrix such as
\[
J = \begin{bmatrix} 
\psi_{1,1} & \psi_{1,2} & \dots & \psi_{1,n} \\
\psi_{2,1} & \psi_{2,2} & \dots & \psi_{2,n} \\
\vdots & \vdots  & \ddots & \vdots \\
\psi_{m,1} & \psi_{m,2} & \dots & \psi_{m,n} \\
\end{bmatrix}
\]
where each column represents the influence of joint $j$ over each task $i$.
A simple rule for calculating each element $\psi_{i,j}$ is presented in Table \ref{tab:jacobian}.

{
\renewcommand{\arraystretch}{1.3}
\begin{table}[]

	\centering
	\begin{tabular}{|l|l|l|l|}
		\hline
		\multicolumn{2}{|c|}{\multirow{2}{*}{$\psi_{i,j} =$}} & \multicolumn{2}{c|}{Joint $j$ is} \\ \cline{3-4} 
		\multicolumn{2}{|c|}{} & Prismatic & Revolute \\ 
		\hline
		& & & \\
		\multirow{2}{*}{Task $i$ is} 
		& Translation 
		&  $\vec{\text{rot}}^{i}_z$
		&  $\vec{\text{rot}}^{i}_z\times (\vec{\text{pos}}^{i} - \vec{\text{pos}}^{i-1})$
		\\ 
		& & & \\
		\cline{2-4} 
		& & & \\
		& Rotation or Posture
		& 0 
		&  $\vec{\text{rot}}^{i}_z$
		\\ 
		& & & \\
		\hline
	\end{tabular}
	\caption[Calculation of the Jacobian terms]{Calculation of the Jacobian terms $\psi_{i,j}$.}
	\label{tab:jacobian}
\end{table}
}

Let $T_i$ be the transform matrix for the frame of joint $i$:
{
\def\VR{\kern-\arraycolsep\strut\vrule &\kern-\arraycolsep}
\def\vr{\kern-\arraycolsep & \kern-\arraycolsep}
\[
\mathbf{T_i} = 
\bordermatrix{ & \vec{\text{rot}}^{i}_x & \vec{\text{rot}}^{i}_y & \vec{\text{rot}}^{i}_z & \vr \vec{\text{pos}}^{i} \cr
& 	\begin{bmatrix}	r_{x_1} \\ r_{x_2} \\ r_{x_3}	\end{bmatrix} & 
	\begin{bmatrix}	r_{y_1} \\ r_{y_2} \\ r_{y_3}	\end{bmatrix} & 
	\begin{bmatrix}	r_{z_1} \\ r_{z_2} \\ r_{z_1}	\end{bmatrix} & \VR
	\begin{bmatrix}	x \\ y \\ z	\end{bmatrix}\cr
& & & & \cr
& 0 & 0 & 0 & & 1 \cr
}\qquad
\]
}

The term $\vec{\text{pos}}^{0,i}$ is the translation between the root frame and joint $i$'s frame, while $\vec{\text{rot}}^{0,i}_z$ is the $z$-vector of the rotation between the root frame and joint $i$'s frame.
Assuming that the root frame is located at $[0, 0, 0]$ and that its rotation is equal to $I_3$ (i.e., $T_{\text{root}}=I_4$), we can take the values of both $\vec{\text{rot}}^{0,i}_z$ and $\vec{\text{pos}}^{0,i}_{i}$ directly from matrix $T_i$.
If that is not the case, then either $T_i$ must be transformed by $T_{\text{root}}^{-1}$, or both vectors $\vec{\text{rot}}$ and $\vec{\text{pos}}$ must be transformed by that inverse.

Please refer to \cite{Buss2009} or \cite{Baerlocher2001} for more information on how to calculate the Jacobian matrix, or alternatively to \cite{Orin1984} for a fully detailed description.
This matrix allows to approximate the change in the end-effector's transform given an increment in the system's joint angles of $\Delta\theta$: 
\begin{equation}\label{eq:ikForwardJacobian}
\Delta\vec{s} \approx J\Delta\theta
\end{equation} 
The problem will be solved by seeking a value for $\Delta\theta$ such that $\Delta\vec{s}$ becomes approximately equal to $\vec{e}$, by making:
\begin{equation}\label{eq:ikForwardJacobian2}
\vec{e} = J\Delta\theta
\end{equation} 
This equals the question \textit{how much must I increment each joint angle $\theta$ in order for the end-effector to move by the amount $\vec{e}$?}

A solution to the IK problem is therefore given by equation (\ref{eq:ikForwardJacobian}) for $\Delta\theta$, using the inverse of the Jacobian: 
$$\Delta\theta = J^{-1}\vec{e}$$
The implementation of any variation of the Jacobian methods typically follow a similar approach, 
which is as an optimization problem that minimizes the residual error $e^{\text{total}}=\|\vec{e}\|$.

In most cases however, this equation cannot be solved uniquely, as Jacobian $J$ may be non-square, non-invertible, or nearly singular (which would provide poor and unstable results).
Several alternatives have been found to calculate the Jacobian's inverse.
One of them is to use the Jacobian's transpose $J^T$ instead of its inverse, and multiplying it by an appropriate scalar $\alpha$ (Equation \ref{eq:jacobian_transpose}).
\begin{equation}
\label{eq:jacobian_transpose}
\Delta\theta = \alpha J^{T}\vec{e}
\end{equation} 
Another possibility is to use its pseudoinverse $J^\dag$ (also called the \textit{Moore-Penrose inverse of J}) as shown in Equation \ref{eq:jacobian_pseudoinverse}.
\begin{equation}
\label{eq:jacobian_pseudoinverse}
\Delta\theta = J^{\dagger}\vec{e}
\end{equation} 
Using the pseudoinverse method also allows to perform a projection into the nullspace of the Jacobian, meaning that we may further optimize the solution towards a secondary task as shown in Equation \ref{eq:pinv_secondary_task}.
An example of that would be to use the end-effector's orientation as the main task $\vec{e}$, for which the solved $\Delta\theta$ would minimize the error $J\Delta\theta-\vec{e}$, while choosing a $\vec{z}$ vector of the same dimension as $\theta$, that would attempt to keep the resulting angles as close as possible to zero (secondary task), without disrupting the main task.
\begin{equation}
\label{eq:pinv_secondary_task}
\begin{split}
P_{N(J)} &= I-J^{\dagger}J\\
\Delta\theta &= J^{\dagger}\vec{e}+P_{N(J)}\vec{z}
\end{split}
\end{equation} 
The $\vec{z}$ vector can be calculated by minimizing a criterion $h(\theta)$, using $\vec{z} = \xi\nabla h(\theta)$, where $\xi$ is a gain factor.
Baerlocher shows an example of the typical application of keeping the joint angles as close as possible to some desired values (e.g. to zero) \cite{Baerlocher2001}, by using $h(\theta, \theta_{desired}) = \|\theta - \theta_{desired}\|^2$. 
The example of keeping the joint angles close to zero would therefore be to have just $h(\theta) = \|\theta\|^2$. 
Alternatively, if the secondary task $e_2$ is clearly represented as a Jacobian matrix $J_2$, then we might also use Equation \ref{eq:pinv_secondary_task_jacob_z}, as explained by \cite{Baerlocher2001}.
\begin{equation}
\label{eq:pinv_secondary_task_jacob_z}
z = (J_2P_{N(J_1)})^\dagger(\vec{e_2}-J_2J_1^\dagger\vec{e})
\end{equation}

Both the transpose and the pseudoinverse methods however, suffer from either approximation errors, or from instability near singularities. 
Such methods also suffer from poor results when the target is too distant from the current position or orientations.
One method to mitigate that problem is also presented by Buss \cite{Buss2009}, and consists in clamping the $\vec{e}$ vector so that its norm is never greater than a constant value $D_{max}$, as shown in Equation \ref{eq:buss_clampmag}.
\begin{equation}
\label{eq:buss_clampmag}
\vec{e} = \texttt{ClampMag}(\vec{t} - \vec{s}, D_{max})\\
\end{equation} 
\[
\text{ClampMag}(w, d) = \left\{
\begin{array}{ll}
w & \text{if}\ \|w\| \leq d\\
d\frac{w}{\|w\|} & otherwise
\end{array}
\right.
\]

The damped least squares method (DLS), also called the Levenberg-Marquardt method further attempts to address these issues, by including a non-zero damping constant.
This constant however, must be chosen carefully depending on the kinematic configuration of the system and on its purpose, in order to remain numerically stable near singularities, without keeping the convergence rate too slow.
Equation \ref{eq:dls} shows how to calculate $\Delta\theta$ using the DLS method, where $\lambda$ is the damping constant, which must be carefully selected based on the details of the multibody and expected target positions, in order to ensure stability. 
A larger damping value allows the solutions to become more stable near singularities, however if the constant is too large then the convergence rate will be lower (as it will require more iterations).
\begin{equation}
\label{eq:dls}
\begin{split}
J^{\dagger^{\lambda}} &= J^T(JJ^T+\lambda^2I)^{-1}\\
\Delta\theta &= J^{\dagger^{\lambda}}\vec{e}
\end{split}
\end{equation} 

Alternatively, the DLS method may also be implemented through the Singular Value Decomposition method (SVD), which decomposes a matrix $J$ of $m\times n$ into three matrices $U$ ($m\times m$), $D$ ($m\times n$) and $V$ ($n\times n$), such that $J=UDV^T$. $D$ is the singular value matrix of $J$, with its only non-zero values being along its diagonal $d_{i,i}=\sigma_i$, being $\sigma_i$ the $i^{th}$ singular value of $J$. Also, because $\sigma_i$ may be zero, let $r$ be the largest value such that $\sigma_r\neq 0$, with $\sigma$ being sorted such that $\sigma_i\geq\sigma_{i+1}$.
Based on the SVD of $J$ and following the elaboration by \cite{Buss2009}, the DLS method can also be expressed as in Equation \ref{eq:dls_svd}:
\begin{equation}
\label{eq:dls_svd}
\begin{split}
J^{\dagger^{\lambda}} &= (\sum_{i=1}^{r} \frac{\sigma_i}{\sigma_i^2+\lambda^2})\textbf{v}_i\textbf{u}_i^T\\
\Delta\theta &= J^{\dagger^{\lambda}}\vec{e}
\end{split}
\end{equation}

As mentioned before, the major issue with the DLS technique is the selection of an appropriate damping factor.
Buss and Kim \cite{BussKim2005} address this issues with the Selectively Damped Least Squares (SDLS) method that adjusts the damping factor for each singular vector of the Jacobian's singular value decomposition (SVD). This method converges faster than DLS and does not require ad hoc damping constants.
First a global $\gamma_\text{max}$ is chosen, for which they recommend a typical value to be $\pi/4$ (45 degrees). This will be the maximum permissible change in any joint angle in a single iteration.
Then we take the SVD of $J = UDV^T$ and express the desired change in end-effector position as $\vec{e}=\sum_{i}\alpha_i u_i$ where $u_i$ is the $i^{th}$ column of $U$ and $\alpha_i=\langle\vec{e},u_i\rangle = u_i^T\vec{e}$.
Let also $\rho_{\ell,j} = \|\partial s_\ell/\partial\theta_j\|$ be the relative magnitude of the change of the $\ell\ th$ task variable in response to a small change in the $j$th joint angle (from Equation \ref{eq:jacobian_def_partial}).
We further define the auxiliary $N$ and $M$ vectors along with the selective damping factor $\gamma$:
\begin{equation}
\label{eq:sdls_NM}
\begin{split}
N_i &= \sum_{j=1}^m\|u_{j,i}\|, \forall i \in [1,n]\\
M'_{i,\ell} &= \sigma_i^{-1}\sum_{j=1}^n|v_{j,i}|\rho_{\ell,j}, \forall i \in [1,m], \forall \ell \in [1,m] \\
M_i &= \sigma_{\ell=1}^m M{i,\ell}, \forall i \in [1,n] \\
\gamma_i &= \text{min}(1, \frac{N_i}{M_i})\cdot\gamma_\text{max}\\
\end{split}
\end{equation} 
Finally, the SDLS solution is expressed as $\Delta\theta$:
\begin{equation}
\begin{split}
\label{eq:sdls_clampmaxabs_thetas}
\varphi_i &= \texttt{ClampMaxAbs}(\sigma_i^{-1}\alpha_i v_i, \gamma_i)\\
\Delta\theta &= \texttt{ClampMaxAbs}(\sum_{i=1}^{r}\varphi_i,\gamma_{\text{max}})\\
\end{split}
\end{equation} 

Baerlocher introduced techniques that allow to solve the IK problem for multiple tasks with priorities, i.e., by specifying the priority in which each task should be achieved \cite{Baerlocher2001}.
In particular he aimed at solving the problem of postural control for virtual humans, by allowing to specify e.g. a task for one hand to reach a certain goal position, plus another task for the head to face a certain direction, while keeping the whole body balanced.
His technique is actually a rewritten version of the solution initially proposed by Maciejewski \cite{Maciejewski85}, upon also being modified to account for algorithmic singularities.
We found his approach to be the most significant one to compare to given our goals.
Equation \ref{eq:baerlocher_eq1} presents Baerlocher's formulation of the DLS applied to two tasks $\vec{e_1}$ and $\vec{e_2}$, whose corresponding Jacobian matrices are $J_i$ and damping constants $\lambda_i,\ i\in[1,2]$, with the first task having a higher priority than the second.
\begin{equation}
\label{eq:baerlocher_eq1}
\begin{split}
\Delta\theta &= J_1^{\dagger^{\lambda_1}}\vec{e_1}+(J_2 P_{N(J_1)})^{\dagger^{\lambda_2}}(\vec{e_2}-J_2J_1^{\dagger^{\lambda_1}}\vec{e_1})
\end{split}
\end{equation} 
He finally elaborates towards a formulation that supports more than two levels of priority, by following the same approach.
In that case, given a set of tasks $[\vec{e_1}, \vec{e_2}, ..., \vec{e_p}]$, for which $J_i$ and $\lambda_i,\ i\in[1,p]$ are the corresponding Jacobian and damping constants, with $i=1$ corresponding to the highest priority, and $i=p$ to the lowest, Equation \ref{eq:baerlocher_multiple_tasks} presents the general formulation for the multiple-task-priority method:
\begin{equation}
\label{eq:baerlocher_multiple_tasks}
\begin{split}
\Delta\theta_i &= \Delta\theta_{i-1}+(J_i P_{N(J^A_{i-1})})^{\dagger^{\lambda_i}}(\vec{e_i}-J_i\Delta\theta_{i-1})\\
\Delta\theta_1 &= J_1^{\dagger^{\lambda_1}}\vec{e_1}\\
J^A_{i} &= 
\begin{bmatrix}
J_{1} \\
J_{2} \\
\vdots \\
J_{i}
\end{bmatrix}
\end{split}
\end{equation} 

The major difference between his problem statement and ours is that his problem is especially directed at virtual humans (VH) with many DoFs while ours is directed at robots with much fewer DoFs than the VH, therefore his problem is more under-constrained (or redundant) than ours.
One of the consequences of that is that the null-space projection operator in his situation will allow for the secondary task to perform much better than in our case.

Finally, within his techniques, Baerlocher also suggests the use of Maciejewski's method for computing an appropriate damping factor based on the minimum singular value of the Jacobian \cite{Maciejewski88}.
Let $b_{\text{max}}$ be a bound on the norm of the solution such that $\norm{J^{\dagger^\lambda}\Delta x}\leq b_{\text{max}}$, then Maciejewski's damping factor can be calculated through Equation \ref{eq:maciejewski_damping_factor}.
\begin{equation}
\begin{split}
\lambda &= \left\{
\begin{array}{ll}
\frac{d}{2} & \text{if}\ \sigma_{\text{min}}\leq\frac{d}{2}\\
\sqrt{\sigma_{\text{min}}(d-\sigma_{\text{min}})} & \text{if}\ \frac{d}{2}\leq\sigma_{\text{min}}\leq d\\
0 & \text{if}\ \sigma_{\text{min}}\geq d\\
\end{array}
\right.\\
d &= \frac{\norm{\vec{e}}}{b_{\text{max}}}
\end{split}
\label{eq:maciejewski_damping_factor}
\end{equation} 

Conclusions drawn from the comparison of several Jacobian techniques (e.g., Jacobian Transpose, Damped Least Squares (DLS), Selectively Damped Least Squares (SDLS)), both by Buss \cite{Buss2009} and by Aristidou \cite{Aristidou2018} are that the Jacobian methods are mostly appropriate for single end-effector situations, not always suitable for time-critical situations (e.g. real-time computation) and the incorporation of constraints using this family of methods is neither straightforward nor controllable towards an optimal solution.
Furthermore, while the SDLS seems to be the most promising method, it is not clear how to use it along with a secondary task.

To conclude this section we share the base pseudocode for such methods in Algorithm \ref{alg:jacobian}.

\begin{algorithm}[!tp]
	\SetKwInOut{Input}{input}
	\Input{$\theta, \vec{t}$\tcp*[r]{initial joint angles, \\target task variables (position and/or orientation)}}
	$\theta' \leftarrow (\theta_1,...,\theta_N)^T$\\
	$\dot{\theta} \leftarrow \vec{0}$\\
	$\text{best}_{\dot{\theta}} \leftarrow \dot{\theta}$\\
	$\text{best}_{\text{error}} \leftarrow \text{MAX\_FLOAT}$\\
	$\vec{s} \leftarrow \text{ForwardKinematics}(\theta')$\tcp*[r]{calculate EE position and/or orientation from $\theta'$}
	\For{$N\leftarrow 1$ \KwTo $\text{MAX\_ITERATIONS}$}{
		$\vec{e} \leftarrow \text{ClampMag}(\vec{t}-\vec{s}, D_{max})$\tcp*[r]{where $\vec{t}$ is the target position and/or orientation}
		\uIf{$\|\vec{e}\| \leq \text{best}_{\text{error}}$}{
			$\text{best}_{\text{error}} \leftarrow \|\vec{e}\|$\\
			$\text{best}_{\dot{\theta}} \leftarrow \dot{\theta}$\\
		}
		\uIf{$\|\vec{e}\| \leq \text{ERROR\_TOLERANCE}$}{
			break\\
		}
		$J \leftarrow \text{Jacobian}(\theta')$\tcp*[h]{calculate Jacobian of $\theta'$}\\
		$J^{-1} \leftarrow \text{CalculateInverse}(J)$\tcp*[h]{using one of the possible methods}\\
		$\dot{\theta} \leftarrow J^{-1}\cdot\vec{e}$\\
		$\theta' \leftarrow \theta' + \dot{\theta}$\\
		$\vec{s} \leftarrow \text{ForwardKinematics}(\theta')$\tcp*[r]{calculate EE position and/or orientation from $\theta'$}
	}
	\Return{$\theta + \text{best}_{\dot{\theta}}$}\\
	\caption{Pseudocode for a typical Jacobian method's iterative solver.}
	\label{alg:jacobian}
\end{algorithm}

Analytical solutions may seem like another candidate option, as they are good for time-critical situations, however their closed-form nature makes them unsuitable for scalability.
For scalable and extensible time-critical situations, which are the ones we are interested in, heuristic approaches seem to be the best choice, both because of efficiency (faster), scalability (no virtual limit on the DoFs) and extensibility (multiple end-effectors, multiple goals/tasks, etc.).

One such heuristic approach is the Cyclic Coordinate Descent (CCD), which is very popular both in computer graphics animation and in robotics \cite{WangChen1991, Aristidou2018}. 
It is an iterative, univariate type of algorithm, as it solves each variable (DoF) one by one, through a series of iterations that attempt to minimize the error between the current KS and the given target.
Some of its main advantages are that it is very easy to implement, fast to compute, and has linear-time complexity regarding the number of DoFs. However it typically returns un-natural poses and consecutive executions of it with similar parameters frequently result in large discontinuities.


FABRIK is another heuristic iterative method that takes on a geometric approach to the IK problem \cite{Aristidou2011a,Aristidou2016}. 
It was inspired by the knot-tying problem \cite{Brown2004} and borrows the idea of iterating through each joint individually as in CCD, but instead works in the joint-position space (instead of angles), and each iteration includes a forward step (traversing from the end-point to the base) followed by a backward step (that traverses from the base back to the end-point).
This technique was created for, and works in virtual space, as it intentionally breaks the kinematic configuration of the system by stretching each segment during the Forward phase, which most likely ends up bringing the base joint to a position that is not the origin of the space as it was initially.
However the Backward phase solves this, while bringing the whole KS closer to a solution.
By working directly in the joint-position space, FABRIK avoids calculation of angles, which is one of its main advantages, making it even faster to compute than CCD. Other of its main features are that it does not suffer from singularity problems, produces naturally smooth and continuous motion, and emphasizes movement in the joints closer to the base. Following an approach similar to \cite{MerrickDwyer2004}, it also supports multiple end-effectors, and as such, full-body IK solving. Regarding the application of constraints, the authors present successful results in a system where each link is modelled as a generic 3-DoF, by decomposing the induced quaternions into swing and twist components, and enforcing limits on them separately following on the method described in \cite{BaerlocherBoulic2001}.
However FABRIK cannot be trivially used with robots, as it is suited for 3-DoF links.
In particular, by operating only in the Cartesian space, it does not properly represent twist motions, i.e., a joint whose rotation axis is aligned with its segment.
While virtually one could see it performing twist, enforcing angular limits to such joints becomes impractical.

Regarding expressive posture control, Neff \& Fiume have presented the Body Shape Solver \cite{MicahelNeff2004} which addresses the problems of pose modelling, balance, and world-space and body-space constraints into a single integrated solver for humanoid skeletons. 
The technique can be used by animators to solve for character poses either based on a given set of parameters, or by selecting a shape set.
However their algorithm is specific to the human body, as it is a hybrid technique that uses both analytical and optimization methods

Johnson has proposed an Expressive IK solution that also uses expert body knowledge (example poses given by animators) to augment the quality of the results given by a CCD algorithm \cite{Johnson2003}. 
The examples are both used to estimate joint constraints, and also to perform multi-target pose blending which would then be used as an initial solution before the IK algorithm is ran (this step was not developed, however). The algorithm, QuCCD, is a Quaternion-based version of the popular CCD algorithm. QuCCD includes a fast joint-limit constraint approach similar to \cite{BaerlocherBoulic2001}, that takes on a geometrical approach instead of clamping angles as usual (which would require converting the quaternion to Euler angles, clamp, and then back to a quaternion).
Some of Johnson's proposed techniques were used to animated Anemone, an expressive IK robot \cite{Breazeal2002}. This robot used a hybrid between pose-blending, for the DoFs near its base, and QuCCD, to animate the upper half, so that it could both maintain an expressive posture, while still facing its "head" towards things in its environment. 
The whole computation was performed through quaternions, holding off the conversion until "just-in-time", before converting and sending the actual Euler angles to the motors.
Despite presenting promising results for 3D animated characters, the author does end up announcing that \enquote{this method tends to produce very slow convergence for 1 DOF joints which are constantly bumping into a boundary}.

Grochow et al. propose an IK system that is trained through a set of human poses \cite{GrochowEtal2004}. 
The selected poses will therefore define the style of the resulting motion. By training with different poses, one can drive the solver to produce different styles of animation. A key feature is that it can both extrapolate a new pose from a style training set, while also allowing to interpolate between different styles. However, despite addressing the problem of style and expressivity of IK, the system was especially developed for motion capture, and requires off-line training, which confines the results to be highly dependant on the quality of the training data.



The Particle IK Solver, featured in the video-game \textit{Spore} and mentioned in the previous subsection, was developed to allow characters with various custom morphologies to walk naturally and to perform actions in their surrounding environment such as looking towards a direction, or grasping an object \cite{Hecker2008}.
Particle IK can solve for various goals by using embodiments that result in an \textit{underdetermined system}, i.e., ones that will result in more DoFs than IK goals.
Therefore the remaining DoFs can be used to achieve secondary objectives.
The solver runs in two phases. 
First it solves for the spine of the character and then for the limb poses, while treating the spine as fixed.
Their argument was that a single-phase solver based on existing techniques did not allow them to make specific ad hoc tuning adjustments or treat special cases, without compromising the quality of the solution in other areas of the pose.
By elaborating a new solver, they managed to achieve \textit{local control} over the solution, which was not possible using conventional IK techniques.

\section{\erik\ Pipeline and Model}
\erik\ is an iterative algorithm for expressive kinematics that was developed with articulated structures of 1-DoF joints in mind, such as real robots, and in particular, robotic manipulators.
It provides a joint model that allows to use techniques initially developed for CGI and not for robotics, such as FABRIK or other IK techniques, which solve for Cartesian (position-based) solutions, instead of angle-based solutions as is commonly used in robotics.

Our algorithm was initially developed towards the problem of expressive gazing, in which a given embodiment, composed of an articulated kinematics chain, is required to orient its end-point towards a target, while also providing expressive control over its posture using expert body knowledge provided by character animators.

Although technically an iterative algorithm, we may also describe \erik\ as a multi-phase super-iterative algorithm given that for each set of goals, it solves them iteratively, while using other iterative techniques within each of its iterations.
In particular within each iteration it may solve small steps using the popular CCD technique, and will use the custom BWCD technique, which is an adaptation of the CCD algorithm, tailored to simplify some of the steps within \erik.
\subsection{From FABRIK to Expressive Robots}
The major portion of the algorithm was inspired by the FABRIK technique \cite{Aristidou2016}.
While CCD is commonly used in isolation to solve the IK problem required for a given end-point to face a given direction, its solutions suffer from discontinuities and un-natural poses.
In this aspect, FABRIK performs significantly better, which makes it more appropriate to be used for expressive motion.
However, by operating on the Cartesian level, it cannot ensure reliable orientation constraints.
Given a set of parallel, 1-DoF joints as we commonly find in robots, it frequently runs into indeterminations, given that a Cartesian representation of a skeleton can not properly represent induced parallel rotations (i.e., twist).
As the authors point out, that results in deadlock situations \cite{Aristidou2016}.
They propose that deadlocks can be detected by checking if the distance between the target and the end-point is becoming smaller on each iteration.
If not, a deadlock situation is detected.
We have imported this concept into \erik, although we have called these the Nonconvergence cases, for which we provide additional \textbf{Nonconvergence Tricks}.
Our dealing if the Nonconvergence cases is expressly different, given that under constraints, we must allow the end-point orientation to temporarily move away from the target in some situations, while it is e.g. twisting its root joint to readjust the whole chain to allow reaching the goal, which makes the Nonconvergence detection less trivial.
Furthermore, the \textbf{Tricks} we apply must consider the fact that we expect to hold the given expressive target posture as best as possible, while in FABRIK, one of the proposed solutions when the target is detected to be out of reach, is solely to place the whole chain in a straight line (which is OK if we do not care for the resulting posture).
These limitations have restricted FABRIK's use for robotics, as it was especially formulated for motion-capture of virtual humans, and on IK problems for position-based targets.
Still, FABRIK provides various benefits, such as supporting full-body IK i.e., multiple end-points, non-leaf end-points, closed loops, and prismatic (i.e., sliding) joints.
Therefore, we chose FABRIK as the starting point for \erik\ so that in the future we may have the chance to replicate and adopt those same features.

\subsection{BWCD: Backward Coordinate Descent}
\label{sec:BWCD}
The BWCD is an IK technique that was specifically created to solve some of the intermediate steps within \erik.
Its execution is similar to CCD's except that execution starts at the root of the chain instead of at the end-point.
Therefore the bulk of the warping introduced by BWCD will be concentrated at the bottom of the chain, while CCD tens to introduce it at the top of the chain.
The formulation of BWCD was necessary to allow warping postures towards an orientation goal, with preference for having such warping at the root of the chain.
That is because by concentrating most of the warping at the root, we expect to maintain more of the shape of the posture through the rest of the chain, up to the tip.
Because the warping occurs at the root, which is typically less constrained (such as in a \textit{turret}, or a \textit{pan-tilt} mount), BWCD can return an acceptable solution in a small number of iterations (e.g. <5).
Therefore, while being an iterative algorithm, it is fast enough to be used as an internal step within \erik.

Within \erik, BWCD is used to operate both on Postures and on Solutions.
The \textbf{Posture} version solves it in Cartesian space and does not enforce joint rotation limits.
The \textbf{Solution} version runs in angular space and enforces joint rotation limits.

\subsection{The \erik\ Pipeline}

\begin{figure}[!bp]
	\centering
	\includegraphics[width=0.6\columnwidth]{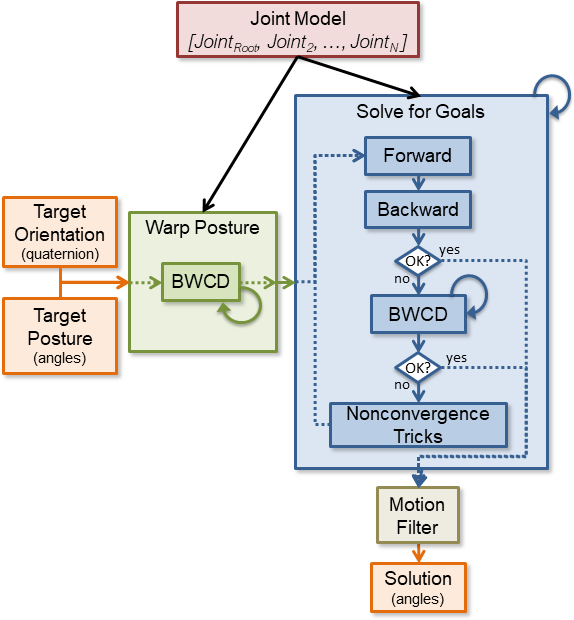}
	\caption[The \erik\ Pipeline]{\erik\ Pipeline. The given Target Posture and Orientation are first warped using BWCD, so that the posture's end-point is aiming towards the Target Orientation, without enforcing joint limits. The result feeds the first iteration of the Iterative portion, which, through various phases on each iteration, returns the final solution. The Joint Model containing the skeletal information and auxiliary operations. The final solution runs through a motion filter to ensure smooth, continuous output.}
	\label{fig:erik_overview}
\end{figure}

Figure \ref{fig:erik_overview} shows the main components of \erik: the inputs \textbf{Target Orientation} and \textbf{Posture}, the \textbf{Joint Model}, the \textbf{Warp Posture} phase, the \textbf{Solve for Goals} phase and the \textbf{Motion Filter}.

\erik\ takes in a \textbf{Target Orientation}, along with a \textbf{Target Posture}, that are to be achieved by the given skeleton, which is the representation of the embodiment's structure, depicted as the \textbf{Joint Model}.
The \textbf{Target Posture} is first naively warped using BWCD, so that its end-point is pointing towards the target orientation. 
This step, however, breaks the kinematic constraints.
Therefore \erik\ moves on to the FABRIK-inspired iterative portion that starts by running a \textbf{Forward Phase}, and a \textbf{Backward Phase} (inspired by FABRIK's own Forward and Backward phases).
After the \textbf{Backward} phase, the candidate solution exhibits a shape as close as possible to the given \textbf{Target Posture}, and respects all kinematic constraints, but its end-point orientation may not match the given \textbf{Target Orientation}.
Upon testing the candidate solution, if it is within the acceptable parameters, then the solution is returned.
Otherwise, the BWCD algorithm is used to orient the solution's end-point towards the given \textbf{Target Orientation}.
This step will likely cause a slight deformation to the intended posture.
If after this step, the new candidate solution is still not acceptable, then \erik\ will proceed with a new iteration, starting from the current candidate solution.
Before doing so however, it may perform some \textbf{Noncovergence Tricks}, in case the algorithm detects that the candidate solution errors are not properly minimizing.

\subsubsection{Nonconvergence Tricks}
Upon detection of a non-converging execution, we attempt two approaches, which we call \textit{tricks}, to attempt to get the solution to converge.
The first attempt is to add a small offset to the target orientation. 
It may be the case that the specific target orientation may not be mechanically achievable, and that the algorithm will deadlock trying to achieve it.
In that case we attempt to perform a random disturbance of a pre-specified magnitude $\Lambda_{\text{Disturbance}\theta}$ on the Target Orientation, and proceed to the next iteration using the new target.

If the execution comes again to a non-convergence detection, then we attempt to run the CCD technique, using the current intermediate solution as the initial state.
This CCD step will likely disturb the expected resulting posture, but will ensure that the end-point is pointing towards the target as best as possible.
Given that we take the current solution as the initial state, it is, however expected that the introduced posture disturbance is minimal.

If still this CCD step was unable to provide an acceptable solution, then it is likely that the intermediate solution has become locked due to joint constraints, and that CCD will not be able to solve it.
In that case, and only in that final case, will we disregard the target posture, and therefore run the CCD technique again, but starting from the zero-pose.

\subsection{The \erik\ Joint Model and LALUT}
In order to allow the use of a FABRIK-based approach with robot-oriented calculus, we started by developing the \erik\ Joint Model (\ejm) that contains all the required information and operations.

\begin{figure}[!htbp]
	\centering
	\includegraphics[width=0.6\columnwidth]{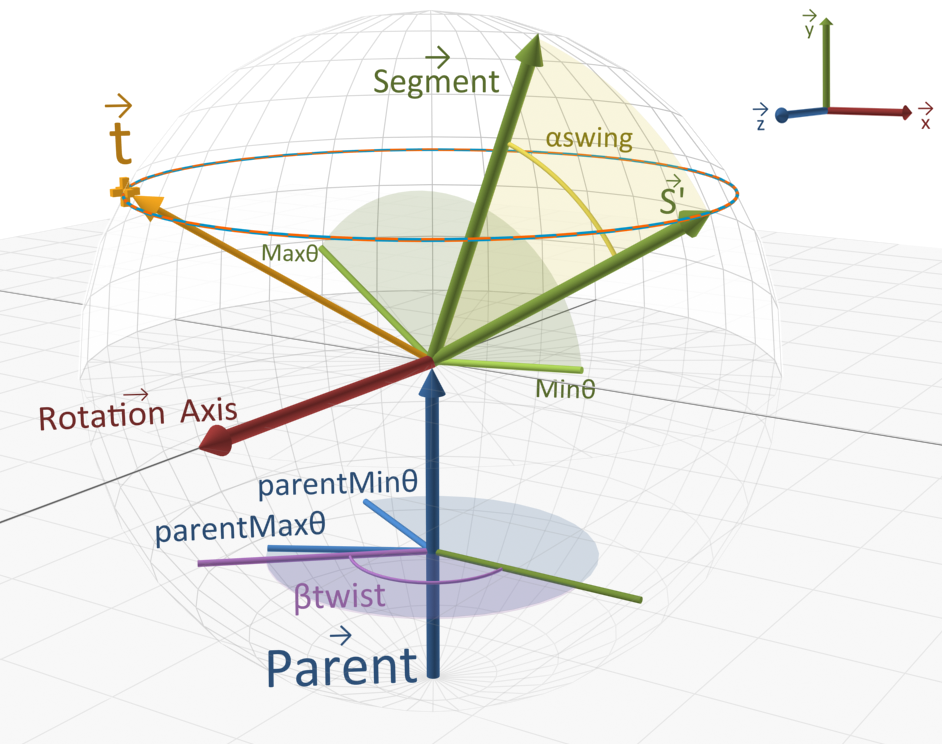}
	\caption[The \erik\ Joint Model]{The \erik\ Joint Model. 
	A joint is defined as having its origin at the tip of its $\vec{\textit{Parent}}$ segment, and in this coordinate frame, to contain its own $\vec{\textit{Segment}}$, which can rotate about a given $\vec{\textit{RotationAxis}}$, within an angle that lies in the range $\{\textit{Min}_\theta, \textit{Max}_\theta\}$. 
	In order to achieve a given target $t$, which is defined in its own local space, it can perform a local rotation of $\alpha_{\text{swing}}$, bringing its segment to $S'$, and then have its parent joint perform a twist of $\beta_{\text{twist}}$ in case the parent is a twister joint.}
	\label{fig:erik_ejm}
\end{figure}

Figure \ref{fig:erik_ejm} shows the unit-sphere \ejm\ space of a joint, where the $\vec{\textit{Parent}}$ segment is connected to the link's $\vec{\textit{Segment}}$, which can rotate about a $\vec{\textit{RotationAxis}}$, within the angular limits of $[\textit{Min}_\theta, \textit{Max}_\theta]$. 

Vector $\vec{t}$ is a target vector, which specifies the direction where we wish to compute a solution for the joint. 
Note that the $\vec{\textit{Parent}}$ was purposely misplaced so that it ends at the origin, to help to visualize this representation as a segment hierarchy, and that all the vectors used are normalized to unit length. 
Note also that we suggest always considering that the $\vec{\textit{Parent}}$ is aligned with the $\vec{y}$ of the child's local space, although other conventions can be used.
The coordinate axes on the top-right corner of Figure \ref{fig:erik_ejm} should help to clarify the convention in case of any doubt.

The goal of the \ejm\ is to provide answers to the following question: \textit{What angular rotation do I need to apply to the local joint, if I know the joint's rotation limits, and if I know that the Parent joint is a Twister, along with how much it can twist?}

Taking figure \ref{fig:erik_ejm} as example, and note the segment $\vec{S'}$.
In order to point the $\vec{\textit{Segment}}$ to $\vec{t}$, the \ejm\ provides the rotation of $\alpha_{\text{swing}}$ on $\vec{\textit{Segment}}$, resulting in $\vec{S'}$, followed by $\beta_{\text{twist}}$ on $\vec{\textit{Parent}}$. 
This would be because the segment wound not achieve $\vec{t}$ through a positive rotation due to its rotational limit $\textit{Max}_\theta$. 
Therefore it needs to locally rotate away from the target, and then rely on its parent's Twist capabilities to finally turn to the right direction.

The $\alpha_{\text{swing}}$ value is calculated using a pre-computed look-up table which we call the \textbf{LALUT}.
First, the target direction $\vec{t}$ is turned into a single decimal number we call a latitude $\lambda$, which is calculated from Equation \ref{eq:erik_latitude}.

\begin{equation}\label{eq:erik_latitude}
\begin{split}
\lambda(\vec{t})&=\sigma(\vec{t})\cdot\frac{\vec{t}\cdot\vec{P}+1}{2}.\\
\sigma(\vec{t}) &= sign(\vec{t}\cdot\vec{POA})\\
\vec{POA} &= \vec{R}\times\vec{P}\text{  \textit{(only computed once)}}
\end{split}
\end{equation}
This $\lambda$ is then used to query the LALUT, which therefore stands for \textit{LAtitude Look-Up Table}.
Additionally some auxiliary vectors are computed only once on joint initialization following Equations \ref{eq:joint_init}

\begin{equation}\label{eq:joint_init}
\begin{split}
\vec{OA} &= \left\{
\begin{array}{ll}
\vec{R}\times\vec{S} &,\text{if}\ \neg(\vec{R}\parallelVec\vec{S})\\
\vec{R}\times\uvecY &,\text{else if}\ \lvert\vec{S}\cdot\uvecX\rvert=1\\
\vec{R}\times\uvecZ &, \text{else}
\end{array}
\right.\\
\vec{POA} &= \left\{
\begin{array}{ll}
\vec{R}\times\vec{P} &,\text{if}\ \neg(\vec{R}\parallelVec\vec{P})\\
\uvecX &,\text{else if}\ \vec{R}\cdot\uvecX=0\\
\uvecZ &, \text{else}
\end{array}
\right.\\
\end{split}
\end{equation}

\begin{figure}[!htbp]
	\centering
	\includegraphics[width=0.6\columnwidth]{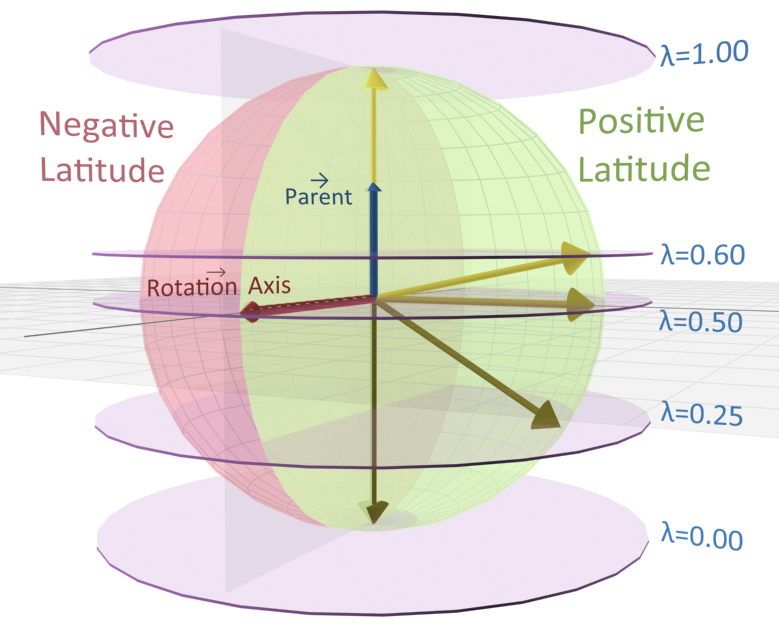}
	\caption[The latitude coordinate system]{The latitude coordinate system. Given a target, represented by the yellow arrows, the corresponding latitude $\lambda$ is calculated following Equation \ref{eq:erik_latitude}. A target pointing at the south pole has a $\lambda$ of zero (0.00), while one pointing at the north pole has a $\lambda$ of one (1.00). The $\lambda$ will be positive or negative depending on if it lies in the right or left hemisphere, which is defined by the plane $\vec{\textit{RotationAxis}}\times\vec{\textit{Parent}}$.}
	\label{fig:erik_latitude}
\end{figure}

Figure \ref{fig:erik_latitude} illustrates the concept of latitude. 
Given one of the target vectors shown, the latitude will be a number between zero and one, which is inspired on the concept of geographic latitude.
The \textit{south pole} corresponds to zero (0.00), while the \textit{north pole} corresponds to one (1.00).
The unit sphere is also split in two vertical hemispheres using the plane defined by the vectors $\vec{\textit{RotationAxis}}$ and $\vec{\textit{Parent}}$.
Given a target vector $\vec{t}$, the hemisphere it lays in is is used to define the sign of the latitude (positive or negative).

\subsubsection{The Latitude Look-Up Table}
The LALUT serves as a look-up table (LUT), 
which consists of an indexed array, stored in memory, for which we can associate a value $y$ to an index $x$.
For intermediate values of $x$ that are not present in the table, it should be able to compute the corresponding values of $y$ by interpolation.
It is computed only once, on initialization, given that the kinematics of a joint are not expected to change in run-time (i.e., the specific vectors ${\vec{\textit{Parent}},\ \vec{\textit{Segment}},\ \vec{\textit{RotationAxis}}}$ of the system remain the same throughout execution). 

This table is computed by iterating a variable $a$ from $\alpha_{\text{min}}$ to $\alpha_{\text{max}}$, in small steps (e.g. $\frac{\pi}{180}rad$). 
The size of the step can be adjusted depending on the needs, with a smaller step requiring linearly more initialization time, but providing higher accuracy. 
In any case the total execution time should be less than a few seconds on a typical computer.

On each iteration, we rotate $\vec{\textit{Segment}}$ using a quaternion $Q_{\text{lutStep}} = \text{AxisAngle}(\vec{\textit{RotationAxis}},a)$, which produces a new vector $\vec{u}$. 
We then store the value of $a$ in the LALUT, indexed by its latitude $\lambda(\vec{u})$. 
Conceptually, this means that the LALUT stores, for a given latitude, the local angle that resulted in it.

Because the LALUT is stored for two hemispheres, it actually contains a positive LUT and a negative LUT.
An entry is placed in either the positive one or the negative one depending on the sign of $\sigma(\vec{u})$.
Later, for retrieval, the same procedure is followed:
Given a target $t$, a latitude $\lambda$ and a sign $\sigma$ are calculated from Equation \ref{eq:erik_latitude}.
If $\sigma$ is negative then the negative LUT is queried for $\lambda$, otherwise the positive one is queried.

\subsection{\erik\ Parameters and Model Specification}
\label{sec:erik_model_specification}
The algorithm relies on a set of Parameters ($\Pi$) used to define the execution goals, which are expected to change frequently (even between each solution), along with a set of Hyperparameters ($\Lambda$) which should remain unchanged throughout the execution, and are used to configure the algorithm execution. 
Table \ref{tab:erik_params} outlines the main Parameters and Hyperparameters required for \erik, along with the symbol by which they shall be represented throughout the document, and especially within Appendix \ref{sec:Algorithm} (Algorithmic Specification).

\begin{table}[!htbp]
\centering
  \caption[Description of Parameters and Hyperparameters of \erik]{Description of Parameters and Hyperparameters of \erik.}
  \label{tab:erik_params}
  \begin{tabular}{cl}
    \toprule
    Symbol&Meaning\\
    \midrule
    $\Pi$ & \erik\ Parameters\\
    $\Lambda$ & \erik\ Hyperparameters\\
    $\Pi_\tau, \Pi_{\vec{\tau}}$ & Target Orientation, Target Direction\\
    $\Pi_\Psi$ & Target Posture\\
    $\Pi_{\Theta_{t-1}}$ &  Previous Solution\\
    $\Lambda_{\text{Sk}}$ & Skeleton Information (\ejm) \\
    $\Lambda_{\text{Sk}_i}$ & $i^{th}$ joint counting from the root\footnotemark\\
    $N_{\text{DoFs}}$ & Number of DoFs of the Skeleton\\
    $\Lambda_\phi$ & Error Function\\
    $\Lambda_{\text{MaxERIKIterations}}$ & Maximum iteration count\\
    $\Lambda_{\text{MaxCCDIterations}}$ & Maximum iteration count for CCD (and BWCD)\\
    $\Lambda_{\text{foo}}$ & Value of Hyperparameter $foo$ \\
    $\Lambda_{\Xi_{\text{bar}}}$ & Extension $bar$ is active \\
    $\Theta_\varepsilon$ & Solution's error value\\
  \bottomrule
\end{tabular}
\end{table}
\begin{table}[!htbp]
	\centering
	\caption[List of symbols and notation for ERIK joint information]{List of joint information, given a joint $k$ of a Solution ($\Theta_k$), a Posture ($\Psi_k$), or a Skeleton ($Sk_k$). Let $\Phi$ represent either a Solution or a Posture.}
	\label{tab:alg_symbols_solution_posture}
	\begin{tabular}{cl}
		\toprule
		Symbol&Meaning\\
		\midrule
		$\Psi_{\text{EE}}, \Theta_{\text{EE}}$ & A Posture or Solution's End-Effector joint.\\
		$\vec{k_{\sigma}}$ & Joint's (child) Segment.\\
		$\vec{k_{RA}}$ & Joint's Rotation Axis.\\
		$\vec{k_{OA}}$ & Joint's Orthogonal Rotation Axis.\\
		$\vec{k_{POA}}$ & Joint's Parent-Orthogonal Rotation Axis.\\
		$\Phi_{k_\rho}$ & World-Position of joint.\\
		$\Phi_{k_\theta}$ & Local angle of joint.\\
		$\Phi_{k_Q}$ & World-Frame (basis) of joint (Quaternion).\\
		$\Phi_{k_L}$ & Local-Frame orientation transform (Quaternion).\\
		$\Phi_{k_\Omega}$ & World-frame orientation transform (Quaternion).\\
		$\vec{\Phi_{k_d}}$ & Direction where the joint segment is pointing at (unit vector).\\
		\bottomrule
	\end{tabular}
\end{table}

Besides the Parameters and Hyperparameters, \erik\ requires the concept and model of Solutions ($\Theta$), Postures ($\Psi$) and Links ($K$).
The Solution object is used both for intermediate and candidate solutions, used internally during the execution of \erik, and also to represent initial and final solutions provided to and by the algorithm.
The Posture object is similar to the Solution one, except that it is used to represent a target pose, which may be represented either based on a set of angles, or a set of positions for each joint, and which may or may not comply with the mechanical limits of the Skeleton.
In case of Solutions, they contain kinematic information that adheres to the joints' kinematic limits.

Additionally, candidate and final solutions contain an error value $\Theta_\varepsilon$ which represents the result of the error function $\Lambda_\phi(\Theta)$.
The Skeleton information object contains the set of Links, along with information such as which is the root or end-point joints of the chain.

Both Solutions and Postures contain joint information represented in a similar way.
The joint fields used by both are listed in Table \ref{tab:alg_symbols_solution_posture}, while some additional algebraic definitions are listed in Table \ref{tab:alg_symbols}.
The computation of some of the fields from Table \ref{tab:alg_symbols_solution_posture} is explained in Equations \ref{eq:joints_1}--\ref{eq:joints_3}.
Let us clarify that the \textit{orientation transforms} $\Phi_{k_L}$ and $\Phi_{k_\Omega}$ represent the orientation to which the joint's segment is facing, after applying its own local rotation.
Also note that an alternative to Equation \ref{eq:joints_3} would be to take the $\vec{y}$ axis of $\Phi_{k_\Omega}$'s corresponding matrix.

\footnotetext{Thus the root joint is $\Lambda_{\text{Sk}_1}$, the end-effector is $\Lambda_{\text{Sk}_N}$, and the Superpoint (Section \ref{sec:erik_model_superpoint}) will be $\Lambda_{\text{Sk}_{N+1}}$.}

\begin{subequations}
\begin{align}
    \Phi_{k_L} &= Q_{\text{AxisAngle}}(\vec{k_{RA}}, \Phi_{k_\theta})\label{eq:joints_1}\\
    \Phi_{k_\Omega} &= \Phi_{k_Q}\cdot \Phi_{k_{Q_L}}\label{eq:joints_2}\\
    \vec{\Phi_{k_d}} &= Q_{\text{AxisAngle}}(\hat{Y}, \Phi_{k_\theta})\label{eq:joints_3}
\end{align}
\end{subequations}

\begin{table}[!htbp]
\centering
  \caption{Definition of mathematical symbols used in the algorithms.}
  \label{tab:alg_symbols}
  \begin{tabular}{cl}
    \toprule
    Symbol&Meaning\\
    \midrule
    $Q_\theta, Q_v$ & Scalar and Vector parts of quaternion $Q$\\
    $Q_M$ & Rotation Matrix that corresponds to quaternion $Q$.\\
    $Q_{k}$ & Axis $k$ of $Q$'s rotation matrix $Q_M$, $k\in\{x,y,z\}$ (simplification of $Q_{M_k}$).\\
    $\uvecX, \uvecY, \uvecZ$ & Unit-vectors in the X, Y or Z directions.\\
  \bottomrule
\end{tabular}
\end{table}

\subsection{The Error Function}
To measure the quality of the solutions produced by \erik, we established two concurrent error measures, $\epsilon_{\text{Orientation}}$ and $\epsilon_{\text{Posture}}$.
These are concurrent measures because in most cases, minimizing one results in not minimizing the other.
Through successive iterations, the algorithm attempts to minimize the error function $\Lambda_\phi$ (Equation \ref{eq:error_function}), which calculates a weighted sum of the two measures.
The error threshold $\Lambda_{\text{Threshold}\varepsilon}$ specifies when the result of the error function is small enough to be acceptable (for which it can successfully terminate and return the computed solution).
In all cases, any value that measures error lies within the interval $[0.0, 1.0]$.
The orientation error function $\phi_{\text{Orientation}}$ calculates the $\epsilon_{\text{Orientation}}$ for a given solution, while similarly, $\phi_{\text{Posture}}$ calculates its $\epsilon_{\text{Posture}}$. 

\begin{equation}
	\Lambda_\phi(\Theta, \tau, \Psi, \Lambda) = \Lambda_{\text{OrientationErrorWeight}}\cdot\phi_{\text{Orientation}}(\Theta_{{\text{EE}}_\Omega}, \tau, \Lambda) + \Lambda_{\text{PostureErrorWeight}}\cdot\phi_{\text{Posture}}(\Theta, \Psi, \Lambda)
	\label{eq:error_function}
\end{equation}

These two error functions are defined in equations \ref{eq:ef_orientation} and \ref{eq:ef_posture}, and further specified in the appendix in Algorithms \ref{alg:ef_orientation} and \ref{alg:ef_posture}.
The posture error function $\phi_{\text{Posture}}$ measures how different the posture of a solution is, in shape, from the target one.
It does so by measuring the local angular deviation between each non-twister solved joint, and target joint, and is designed to punish more for deviations closer to the end-point than closer to the root, which supports our preference. \\ In equations \ref{eq:ef_posture} and \ref{eq:posture_normalizer}, $\alpha$ is a shortcut for the aggravation factor $\Lambda_{\text{ErrorAggravation}}$.

\begin{equation}
\begin{split}
\phi_{\text{Orientation}}(\omega, \tau, \Lambda) &= \left\{
\begin{array}{ll}
\text{min}(Z(\tau, \omega), Z(\tau, \QAA{\RotVQ{\uvecY}{\omega}}{\pi}\cdot\omega)) & , \mathit{if}\ \ \Lambda_{\Xi_{\text{SymmetricEndpoint}}}\\
Z(\tau, \omega) & , \mathit{otherwise}\\
\end{array}
\right.\\
Z(\tau, \omega) &= \frac{\text{min}(\lvert\tau - \omega\rvert, \lvert\tau + \omega\rvert)}{\sqrt{2}}\cdot\\
\end{split}
\label{eq:ef_orientation}
\end{equation}

\begin{equation}
\begin{split}
\phi_{\text{Posture}}(\Theta, \Psi, \Lambda) &= \frac{1}{\Lambda_{\text{PostureNorm}}}\sum_{i=1}^{N_{\text{DoFs}}} \left\{
\begin{array}{ll}
0 & , \mathit{if}\ \ \text{IsTwister}(\Lambda_{\text{Sk}_i})\\
\alpha^i\cdot\lvert (1- \frac{1+\Upsilon(\Psi, i)}{2} ) - (1- \frac{1+\Upsilon(\Theta, i)}{2} ) \rvert & , \mathit{otherwise}\\
\end{array}
\right.\\
\Upsilon(P, i) &= \left\{
\begin{array}{ll}
\lVert\Lambda_{\text{Sk}_{\text{Root}_\sigma}}\rVert\cdot\lVert P_{(i+1)_\rho}-P_{i_\rho}\rVert &, i=1\\
&\\
\lVert P_{i_\rho}-P_{(i-1)_\rho}\rVert\cdot\lVert P_{(i+1)_\rho}-P_{i_\rho}\rVert & otherwise\\
\end{array}
\right.\\
\end{split}
\label{eq:ef_posture}
\end{equation}

The Hyperparameter $\Lambda_{\text{ErrorAggravation}}$ (used in the Equations \ref{eq:ef_posture}-\ref{eq:posture_normalizer} as $\alpha$) defines how worse the punishment becomes, as the function calculates deviations closer to the end-point. A value of 1.0 would mean that the punishment is the same across the links. A value of 2.0 means that a given deviation amount at one link would result in twice the error value, one level up the kinematic chain.
We can see that the resulting value of $\phi_{\text{Posture}}$ is divided by the $\Lambda_{\text{PostureNorm}}$, which reduces the final sum to a value in the interval $[0.0, 1.0]$.
This hyperparameter is calculated once on the skeleton's initialization and given by Equation \ref{eq:posture_normalizer}.

\begin{equation}
	\Lambda_{\text{PostureNorm}} = \sum_{i=1}^{N_{\text{DoFs}}} \left\{
	\begin{array}{ll}
	0 & , \textit{if}\ \ \text{IsTwister}(\Lambda_{\text{Sk}_i})\\
	\alpha^i & \text{otherwise} \\
	\end{array}
	\right.\\
	\label{eq:posture_normalizer}
\end{equation}

Depending on the target application, and the embodiment used, one can use different values for the error measure weights, and for the error threshold.
We share, as an example, that for a 5-link robotic manipulator aimed at entertainment applications, where expressivity and responsiveness is more important than precision, we achieved good results using an error threshold of 0.04, with a weight of 1.0 for $\Lambda_{\text{OrientationErrWeight}}$ and 0.2 for $\Lambda_{\text{PostureErrWeight}}$.
As such, we took these values as a reference when evaluating the algorithm as we will report further in the appropriate section of the document (Section \ref{sec:erik_evaluation}).

\subsection{The Nutty Motion Filter}
\ifboolexpr{togl {Thesis}}{
	The final component of the pipeline is the Nutty Motion Filter, which we refer to as the NMF and has been extensively described in Section \ref{sec:filter}.
}{
	The final component of the pipeline is the Nutty Motion Filter, which we refer to as the NMF \cite{ribeiro2019nutty}.
}
This piece's function is to interpolate successive \erik\ solutions, to ensure that the final produced movement is smooth and continuous.
Furthermore, it can shape the motion to make it appropriate for use with robots.

The NMF allows to define limits for the velocity, acceleration and jerk\footnote{Jerk is commonly used in robotics. It is the derivative of the acceleration. 
Think of it as the speed at which the acceleration changes.} of the signal.
Additionally it includes a set of tweaking parameters that can be creatively explored to provide different characteristics to the motion, such as allowing it to respond fast, as in a light character, or respond very slowly and with a lot of inertia, as in a heavy character.

The motion filter is calculated individually for each joint, at the end of each frame in the animation engine's animation cycle, which is not necessarily synchronized (and should not be) with the \erik\ solver engine.
We recommend not attaching these given that the \erik\ cycle may have inconsistent frame times and drop to a lower rate than is expected in the animation cycle.

The output of the NMF on each frame is given by the function $X(x(t), t(i), s)$, where $x(t):\mathbb{R}^{+}_0\to[P_{\text{min}}, P_{\text{max}}]$ is the motion signal history, i.e., the previous positions that were output from the filter. 
The parameters $P_{\text{min}}$ and $P_{\text{max}}$ represent the minimum and maximum values respectively (e.g. angular limits).
Note that each joint may define its own limits and motion parameters for the NMF.
$x(0)$ corresponds to the initial position of the joint and must be initially specified.
The function $t(i):\mathbb{N}_0\to\mathbb{R}^{+}_0$ (shortened to $t_i$) represents the time at each sample $i$, such that $0 \leq t_{i-1} < t_i$, and $t_i-t_{i-1} = \Delta t$, where $\Delta t$ is a fixed time-step, calculated from the animation output rate $R$, such that $\Delta t=\frac{1}{R}$. 
Note that from this definition, $i$ refers to the current sample, and therefore the current time is always represented by $t_i$, while the time of the last sample is $t_{i-1}$ and so on.

Finally, the set-point $s$ is the new target position, and is used to calculate the \textit{induced velocity} $\dot{x}(t_i)$.
With this consideration, $x(t_i)$ is used to represent the output that will be computed of the filter at the current time (not in the history yet), while $s$ therefore represents the input.
As such, $\dot{x}(t_i)$ must be calculated from $s$ instead of $x(t_i)$.

Equation \ref{eq:filter} contains the explicit definition of the NMF equations.
Within them we can find the various motion parameters, which we follow to explain. 

The $\beta$ parameter controls the exponent of the position-limiter de-acceleration, allowing to control how close to the angular limit of the joint the output is allowed to get before being saturated.
As $\beta$ increases, the saturation becomes more similar to a hard clamping function.
The use of a soft limiter allows the output filter to avoid overshooting any joint beyond its physical limits, given that in most cases, overshooting at the software's output level would result in a hard break at the \textit{hardware} level.
The default value for $\beta$ is 1.

The $\{\sigma,\rho\}$ parameters both represent \textit{smoothness} and \textit{responsiveness} respectively, and allow to tweak the filter, changing how quickly it responds and how much it is allowed to oscillate.
We call these the \textit{character parameters}, as different configurations for them will shape the motion differently.
As such we argue that they can be used to model different character traits, even when the same physical limits are enforced.
The \textit{smoothness} parameter $\sigma$ will ease out the oscillations. 
However, depending on other filter parameters such as the physical limits, fully easing out might become too slow and make the motion seem too muddy and flat.
That is where the \textit{responsiveness} parameter $\rho$ comes in, which allows to precipitate the easing out, so that it may still be smooth, but faster, and thus, more responsive.

\ifboolexpr{togl {Thesis}}{
Please refer to Section \ref{sec:filter} for more details and examples on the Nutty Motion Filter and the use of its parameters.
}{
Please refer to the Nutty Animation paper \cite{ribeiro2019nutty} for more details and examples on the Nutty Motion Filter and the use of its parameters.
}

\begin{equation}
\label{eq:filter}
\begin{split}		
\chi(x, t_i) & = x(t_{i-1}) + \lambda(\psi(x, t_i), \mathit{velocity\_limit}) \\
\psi(x, t_i) & = \dot{x}(t_{i-1}) + \lambda(\frac{\xi(x, t_i)-\dot{x}(t_{i-1})}{\Delta t}, \mathit{acceleration\_limit}) \\
\xi(x, t_i) & = \ddot{x}(t_{i-1}) + \lambda(\frac{\frac{v\cdot\Eta(v)-\dot{x}(t_{i-1})}{\Delta t}-\ddot{x}(t_{i-1})}{\Delta t}, \mathit{jerk\_limit}) \\
v &= \Omega(\dot{x}(t_i), x(t_{i-1}), P_{\text{max}}, P_{\text{min}}, \beta) \\
\dot{x}(t_k) &= \left\{
\begin{array}{ll}
\frac{s-x(t_{k-1})}{\Delta t} & , \mathit{if}\ k = i \\
& \\
\frac{x(t_{k})-x(t_{k-1})}{\Delta t} & \mathit{otherwise} \\
\end{array}
\right.\\
\Eta(v) &= \frac{v}{2} \cdot \Bigg(\tanh\bigg(\Big(\frac{\abs{v}}{1-\rho}\Big)^{1-\sigma}-\pi\bigg)+1\Bigg), 0 \leq\sigma\leq 1, 0\leq\rho<1\\
\lambda(x, k) &= \frac{k}{2}\cdot \tanh(x/\frac{k}{2})\\
\Omega(\dot{x}, x, P_{\text{max}}, P_{\text{min}}, \beta) &= \left\{
\begin{array}{ll}
\dot{x}\cdot\Bigg(1-\bigg(\frac{x-P_{\text{min}}-\alpha}{\alpha}\bigg)^{2\beta}\Bigg), & \mathit{if} (x>\alpha\ \&\ \dot{x}>0)\ |\ (x<\alpha\ \&\ \dot{x}<0) \\
& \\
\dot{x}, & \mathit{otherwise} \\
\end{array}
\right.\\
\alpha &= \frac{P_{\text{max}} - P_{\text{min}}}{2}\\
\end{split}
\end{equation}


\subsection{The Superpoint}
\label{sec:erik_model_superpoint}
In order for some of the calculations to work on the end-point link, we created the concept of the \textit{Superpoint}.
This is a \textit{fake}, 0-DoF joint, used within Postures, that extends the end-point's segment.
It allows the End-point to be treated as if it had a child link with 0-DoF.
Whenever the Posture's data for the End-point is changed, the data for the Superpoint is also updated, using the rules in Equation \ref{eq:superpoint}.
Also note, by the definitions in Table \ref{tab:erik_params} that the Superpoint may be referred to either as $\Psi_{EE_{\text{Child}}}$ or $\Lambda_{\text{Sk}_{N+1}}$.
\begin{equation}
\label{eq:superpoint}
\begin{aligned}
    \Psi_{EE_{\text{Child}_\theta}} &= 0 &\text{\textit{Let\ $\Psi$\ be\ the\ posture\ and\ EE\ the\ Endpoint}}\\
    \Psi_{EE_{\text{Child}_Q}} &= \Psi_{EE_\Omega}&\\
    \Psi_{EE_{\text{Child}_\rho}} &= \Psi_{EE_\rho}+Rotate(EE_{\sigma}, \Psi_{EE_\Omega}) &
\end{aligned}
\end{equation}

\subsection{\erik\ Extensions}
Not all embodiments and application pose the same requirements.
As such, \erik\ was designed with the idea of extensions ($\Xi$) in mind.
Think of extensions as options that you may want to have activated or not, which may change the way the algorithm runs, and thus can result on better outcomes for a given situation (while possibly providing worse outcomes, for a different situation, with different criteria).
In that sense, Extensions fall in the category of Hyperparameters, and are therefore contained within those.
The extensions we have designed and included in the algorithm on this paper were all found to yield better results given the purpose we define (i.e., entertainment). 
If your purpose or criteria is different, there is an option to disable such extensions, to modify them, or even to create new ones.
The currently included extensions are:
	
\begin{description}
\item[$\Xi_{\text{SymmetricEndpoint}}$] Allows the algorithm to flip the end-point upside down. 
This is useful if the end-point is symmetric, and can be used both ways. 
By using such a design, and activating this extensions, the possible solution space doubles, and therefore allows the algorithm to properly solve in many more cases.
\item[$\Xi_{\text{AvoidEdges}}$] Instructs the algorithm to avoid positioning joints exactly on its angular limits.
In cases where some minor deviation from the goals is accepted, this extensions helps to avoid dead-lock situation when the joint limits are equivalent to singularity-prone angles (such as $\pm\frac{\pi}{2}$).
\item[$\Xi_{\text{NonConvOffsetTrick}}$] Allows \erik\ to attempt the \textit{Non-converging Offset Trick} when a non-converging execution is detected. 
This trick applies a small, random orientational offset to the target orientation in cases where the execution has become non-converging.
It results in an increase on the amount of cases where the algorithm is able to converge, as long as a minor deviation from the goals is accepted.
The deviation applied is defined by hyperparameter $\Lambda_{\text{Disturbance}\theta}$.
\item[$\Xi_{\text{NonConvCCDTrick}}$] Allows \erik\ to run the CCD algorithm on a non-converging solution, after the \textit{Non-converging Offset Trick} failed to bring the execution into a converging state. 
It typically results in achieving the orientation goal esier, while allowing the posture goal to become more disrupted (as expected through the direct use of CCD).
\end{description}

\section{Evaluation}
\label{sec:erik_evaluation}
Before claiming on the quality and success of \erik, we are required to run extensive evaluation procedures.
Given that the algorithm aims at being used with any embodiment and expressive pose created by animators, we did not want to access if the resulting solutions were able to solve particular use cases, as those should be tailored creatively by such animators in the future.
Instead, we realized that we wanted to assure that the algorithm would be able to fulfill an animator's intentions while authoring expressive postures for use with \erik.
Therefore, given an expressive posture, we wanted to test how well the algorithm was able to hold its shape, while orienting its endpoint towards various different target orientations.
At the same time, we were concerned with how well the resulting solution effectively aimed at the given target orientation, regardless of the resulting expression.
This is because, for interactive, real-work situations, we consider it particularly important to get the aiming right, so that the character is believable, and is able to provide an immersive experience for the user.
The expressivity of any particular posture is not, in fact, evaluated. 
Instead, the evaluation focused on what can be regarded as a meta-expressivity, i.e., given any posture, which an animator would have thought to be appropriately expressive for some purpose, we measure how well the algorithm is able to reach a shape that is similar to the one given by that posture, and that capability is what is evaluated as the expressive goal.

With the purpose of evaluating how well \erik\ solves both the orientational goals and the expressive goals, we performed what can be dubbed as a \textit{brute-force} evaluation procedure.
This procedure consisted of generating many different expressive postures, and testing how well \erik\ is able to solve them for a large set of different orientation targets. 
All this was done for several different embodiments.
It is impossible to cover every possible case through such approach.
However we consider that the tested cases are a sufficient reflection of how the algorithm performs in general, and are representative of both 1) the space of different expressive postures that any animator would possibly produce; and 2) the space of different orientations to which the character might possibly have to face.

Additionally we compared ERIK against an existing technique. 
In this case we followed the description  by Baerlocher on how to solved an IK problem for multiple tasks \cite{Baerlocher2001} based on the DLS method.
Taking the example of a two-priority problem, the first task, with higher priority, would be the orientation constraint, while the secondary task, of lower priority, would be the postural constraint.
The technique was evaluated in the same way we tested ERIK with multiple embodiments, and the results were further included in the same analysis.

\subsubsection{Error Measures}
Through preliminary experimentations, we decided to established a weight of 1.0 for $\epsilon_{\text{Orientation}}$ and 0.2 for $\epsilon_{\text{Posture}}$, along with an error threshold $\Lambda_{\text{Threshold}\varepsilon}$ of 0.04.
The use of these weights states that it is more important to get the orientation goal solved than the expressive posture one.
This is because we prefer that the resulting solution is properly aiming at the target orientation, and, because we are aiming at expressive applications, we tolerate that the posture may fall slightly out of shape, as long as it is still within an acceptable amount of disfigurement.

As to the error threshold, while it should be adapted to each embodiment, we found 0.04 to be a decent tolerance to demonstrate and compare the results among different embodiments.
In a real-world application, we would have tweaked a different error tolerance for each of the different skeletons.

\subsubsection{Evaluation Embodiments}
In order to see how the algorithm performed for embodiments with various amounts of DoFs, we established 7 different test skeletons, which are presented in Table \ref{tab:test_skeletons}. 

It is important to note that we have included skeletons with a low number of DoFs in order to validate that the algorithm behaves as expected, even in such highly constrained situations.
Our hypothesis here is that these low-DoF skeletons will yield very poor results, and that by adding more DoFs, or configuring them in different ways and with different angular limits, we can augment the expressive capabilities of the expressive character, which should be proven by yielding better results in the same type of evaluation.
\newpage

\newcolumntype{R}[1]{>{\raggedright\arraybackslash}m{#1}}
\newcolumntype{C}[1]{>{\centering\arraybackslash}m{#1}}
\newcolumntype{L}[1]{>{\raggedleft\arraybackslash}m{#1}}

\begin{minipage}{\linewidth}
\begin{spacing}{1.0}
	\centering
	\captionof{table}[Definition of test-skeletons used in the ERIK evaluation procedure.]{Definition of test-skeletons used in the evaluation procedure. In the figures, green nodes represent a $\vec{Y}$-oriented rotation axis, while a red one is oriented with $\vec{X}$, and blue with $\vec{Z}$.}
	\label{tab:test_skeletons}
	\begin{tabular}{cm{3cm}m{3cm}m{3cm}}
		\toprule
		\multirow{4}{*}{Skeleton} & 
		\multirow{4}{*}{\shortstack{\\\# DoFs and\\rotation axis\\sequence\\\small(root to endpoint)}} & 
		\multirow{4}{*}{\shortstack{Angular Range}} & 
		\multirow{4}{*}{Illustration} \\
		&&&\\
		&&&\\
		&&&\\
		\hline	
		\multirow{4}{*}{A} &
			\multirow{4}{*}{\shortstack{3 links\\Y-X-Y}} & 
			\multirow{4}{*}{\shortstack{$[-\frac{\pi}{2}, \frac{\pi}{2}]$\\(all links)}}  &
			\multirow{4}{*}{
				\begin{minipage}{\linewidth} \includegraphics[width=\linewidth]{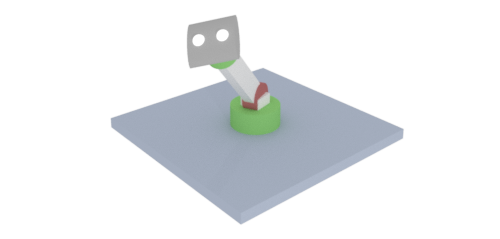} \end{minipage}
			}\\
			&&&\\&&&\\&&&\\
		\hline	
		\multirow{5}{*}{B} &
		\multirow{5}{*}{\shortstack{4 links\\Y-X-Z-Y}} & 
		\multirow{5}{*}{\shortstack{$[-\pi, \pi]$\\(all links)}}  &
		\multirow{5}{*}{
			\begin{minipage}{\linewidth} \includegraphics[width=\linewidth]{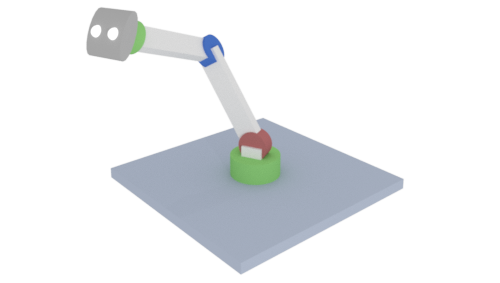} \end{minipage}
		}\\
		&&&\\&&&\\&&&\\&&&\\
			
		\hline	
		\multirow{6}{*}{C} &
		\multirow{6}{*}{\shortstack{5 links\\Y-X-X-Z-Y}} & 
		\multirow{6}{*}{\shortstack{$[-\frac{\pi}{2}, \frac{\pi}{2}]$\\(all links)}}  &
		\multirow{6}{*}{
			\begin{minipage}{\linewidth} \includegraphics[width=\linewidth]{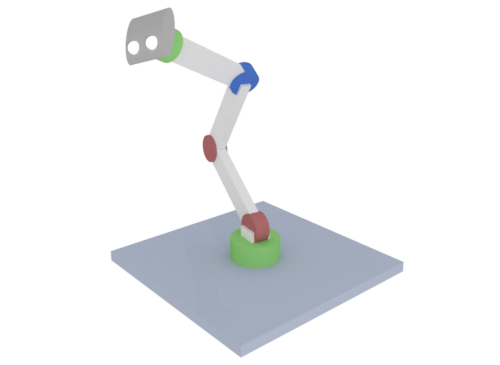} \end{minipage}
		}\\
		&&&\\
		&&&\\
		&&&\\
		&&&\\
		&&&\\
		\hline	
		\multirow{3}{*}{D} &
		\multirow{3}{*}{\shortstack{5 links\\Y-X-Z-X-Y}} & 
		\multirow{3}{*}{\shortstack{$[-\pi, \pi]$\\(all links)}}  &
		\multirow{6}{*}{
			\begin{minipage}{\linewidth} \includegraphics[width=\linewidth]{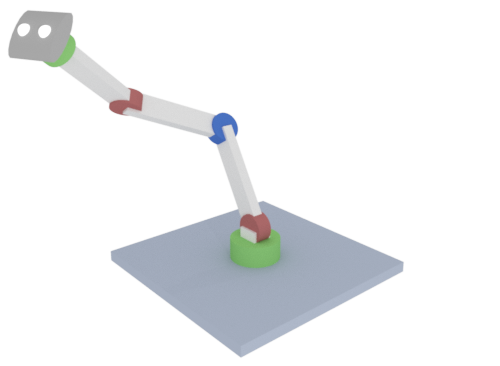} \end{minipage}
		}\\
		&&&\\
		&&&\\
		\cline{1-3}	
		\multirow{3}{*}{E} &
		\multirow{3}{*}{\shortstack{5 links\\Y-X-Z-X-Y}} & 
		\multirow{3}{*}{\shortstack{$[-\frac{\pi}{2}, \frac{\pi}{2}]$\\(all links)}}  &
		\\
		&&&\\
		&&&\\					
		\hline	
		\multirow{6}{*}{F} &
		\multirow{6}{*}{\shortstack{6 links\\Y-X-X-Z-X-Y}} & 
		\multirow{6}{*}{\shortstack{$[-\frac{\pi}{2}, \frac{\pi}{2}]$\\(all links)}}  &
		\multirow{6}{*}{
			\begin{minipage}{\linewidth} \includegraphics[width=\linewidth]{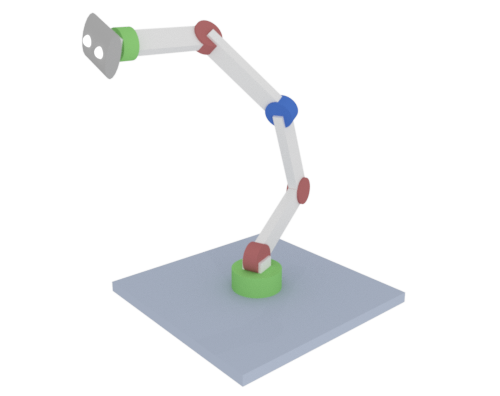} \end{minipage}
		}\\
		&&&\\&&&\\&&&\\&&&\\&&&\\		
		\hline	
		\multirow{7}{*}{G} &
		\multirow{7}{*}{\shortstack{8 links\\Y-X-Z-X-Y-X-Z-Y}} & 
		\multirow{7}{*}{\shortstack{$[-\frac{\pi}{2}, \frac{\pi}{2}]$\\(all links)}}  &
		\multirow{7}{*}{
			\begin{minipage}{\linewidth} \includegraphics[width=\linewidth]{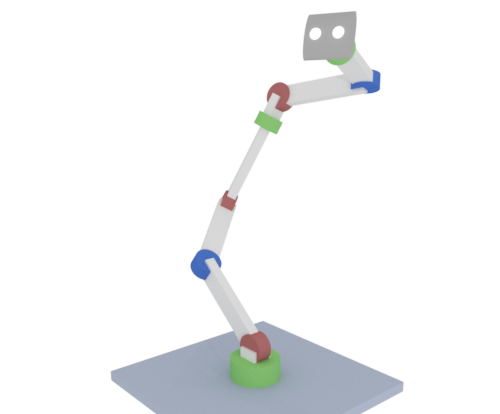} \end{minipage}
		}\\
		&&&\\&&&\\&&&\\&&&\\&&&\\&&&\\
		\bottomrule
	\end{tabular}
\end{spacing}
\end{minipage}

\subsubsection{Procedure}
Each skeleton was used to test \erik\ in various different target postures and target orientations.
The target postures were generated by sweeping the angular range of each joint as long as it is not a root twist-joint, or an endpoint twist-joint, with a given resolution, from its $\textit{min}_\theta$ to its $\textit{max}_\theta$, and combining them to create a large set of postures.
Based on our convention, the twist-joints are the ones whose rotation axis is aligned with $\vec{Y}$.
In fact, the full set of skeletons has twist-joints both as root and as endpoints, which means that for each skeleton, all joints except these two were swept to generate the target postures.
The reason why we exclude these two are that they do not change the actual shape of the posture, and including them would dramatically increase the simulation space.

\begin{minipage}{\linewidth}
	\centering
	\includegraphics[width=0.9\linewidth]{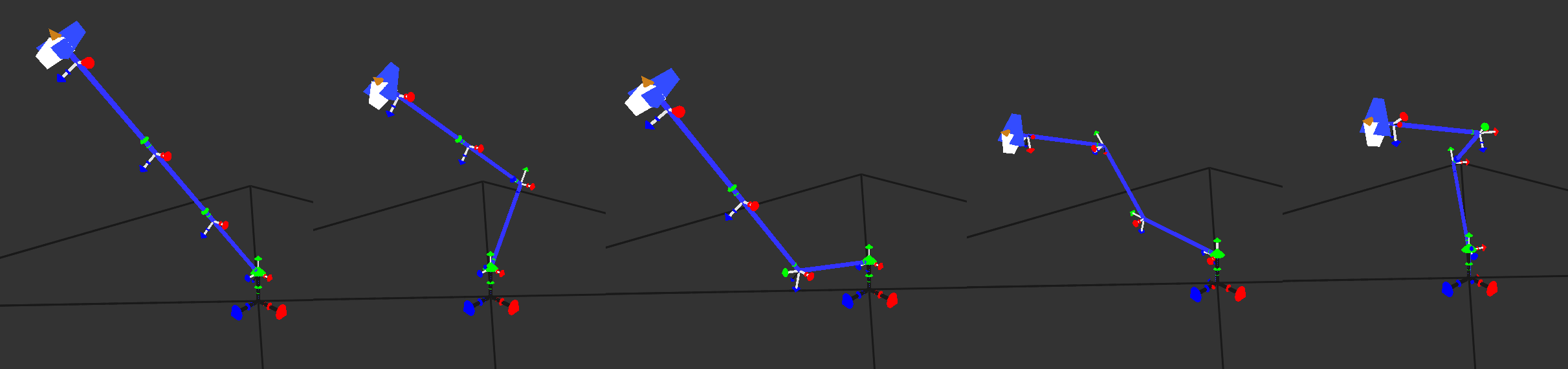}
	\captionof{figure}[Five example test postures for skeleton C]{Five example test postures for skeleton C, representing the type of generated postures tested.}
	\label{fig:procedure_postures}
\end{minipage}\\\\

\begin{minipage}{\linewidth}
	\centering
	\includegraphics[width=0.7\linewidth]{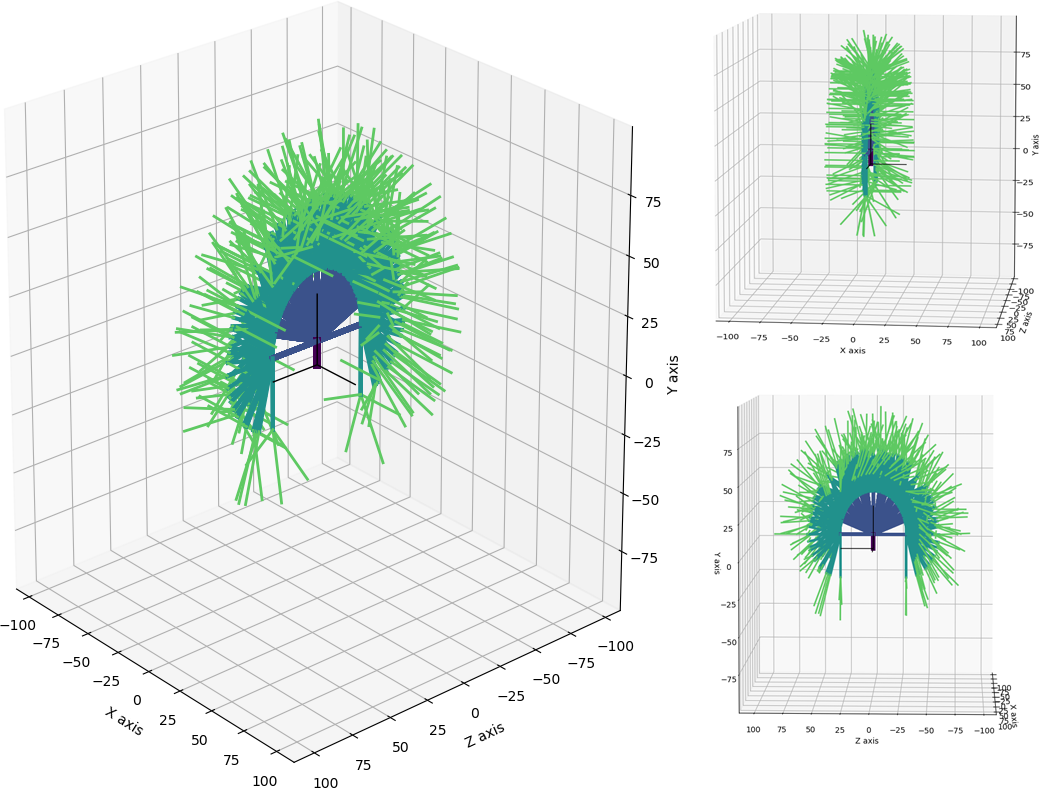}
	\captionof{figure}[The postural simulation space for skeleton C, illustrating 3789 target postures]{The postural simulation space for skeleton C, illustrating 3789 target postures in three different views.}
	\label{fig:skeleton_E_posture_space}
\end{minipage}\\\\

In Figure \ref{fig:procedure_postures} we can see examples of different target postures generated for Skeleton C, with 5 links.
The whole postural simulation space for the same skeleton is illustrated in Figure \ref{fig:skeleton_E_posture_space}, where we see each of the 3789 generated postures overlapped.
Note that the simulation space contains no rotation on the root joint, as it would merely revolve the posture around the vertical $\vec{Y}$ axis, and thus would not change the posture's actual shape.

Similarly, for the target orientations, we wanted to test the most various orientations in all different directions around the character.
For that we swept a horizontal angle $\alpha_h$, a vertical angle $\alpha_v$, and a twist angle $\alpha_t$, all in the range $\{-\pi$, $\pi\}$. 
The sets of three angles were then used to generate a large number of target orientations (as quaternions) through the Yaw-Pitch-Roll composition method.
It may seem that for $\alpha_v$, sweeping in the range $\{-\frac{\pi}{2}, \frac{\pi}{2}\}$ would have been enough; however extending the range to $\{-\pi, \pi\}$ introduces additional target orientations in which the target orientation is defined \textit{upside-down}.
We wanted to include such cases in the evaluation, to ensure that the algorithm was also numerically capable of dealing with them.
As a result, for each of the generated postures of each skeleton, we took a \textit{point-cloud} centered on the robot, each point representing a target orientation (including the roll component).
This method allowed us to run the algorithm on a large amount of different parameters, while also taking extra care to ensure that potential failure points, such as angles set to $\pm\pi$, and orientations aligned with any of the coordinate axes, were guaranteed to be included.

\begin{figure}[!htbp]
	\centering
	\includegraphics[width=0.5\columnwidth]{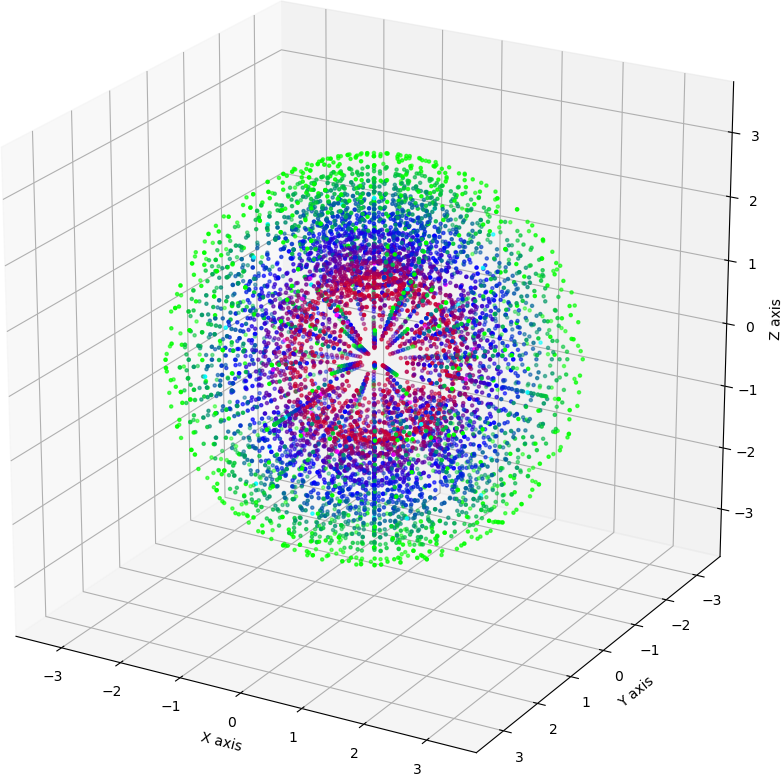}
	\caption[Illustration of a point cloud corresponding to 7609 test samples]{Illustration of a point cloud corresponding to 7609 test samples, each representing a different quaternion to be used as the target orientation. Each point represents a polar and azimuthal orientation angle, along with a twist that is given by its radial distance.
	The colors of the points are modulated from the twist angle (red are negative, blue is zero, green are positive).
	Please note that the apparent existence of blue or even green dots at the center is an illusion - they are in fact part of target directions that are roughly aligned with the viewing direction.}
	\label{fig:point_cloud}
\end{figure}

Figure \ref{fig:point_cloud} shows the orientational simulation space as a point-cloud with 7609 points, which was used to run simulations for each posture or each skeleton.
Each point illustrates a polar and azimuthal orientation angle, along with a twist that is given by the radial distance from the center. 
Therefore variously twisted orientation quaternions are tested in the same direction. 
Both positive and negative twist angles are tested - in this representation, the zero-twist orientations are represented by the points that lie at the center of the point-cloud radius, while positive ones increase towards the exterior, and negative ones towards the interior.

\subsubsection{Comparison with the two-priority DLS}

In order to compare ERIK against another existing technique, we chose to use the two-task-priority DLS as described by Baerlocher \cite{Baerlocher2001}, using Maciejewski's damping factor \cite{Maciejewski88} and the SVD method.
These techniques have already been reviewed in Section \ref{sec:related_ik_jacobian}, and are reiterated in Equation \ref{eq:dls_baerlocher_reit}.

This technique posed as the most appropriate to provide a comparison to ERIK, as it allows us to define two tasks: 
the orientation task characterized by  $J_1\Delta\theta = \vec{e_1}$, with a high priority, and the postural task $J_2\Delta\theta = \vec{e_2}$ with a lower priority.
Both $J_1$ and $\vec{e_1}$ are calculated as they would usually be for an orientation-constraint task.
The secondary task is meant to keep the joint angles as close as possible to a given target posture $\Psi$.
Therefore we calculate $\vec{e_{2,}} = \Psi - (\vec{s} + \Delta x)$, having $\Delta x = J_1^{\dagger^{\lambda_1}}\vec{e_1}$, i.e., the current solution to the primary task. We recall also that $\vec{s}$ is the initial joint configuration.
Therefore the error vector for the secondary task represents the error between the target posture and the posture that results from solving the primary task.
As the secondary task aims at solving towards a given posture, its Jacobian matrix $J_2$ should correspond to the Identity matrix $I^n$, where $n$ is the number of joints.
We add just one correction to it, by setting the value for the $1^{st}$ and $n^{th}$ joint to zero in case that joint is a twist joint, given that as in ERIK, those do not change the resulting posture's overall shape.

\begin{equation}
\label{eq:dls_baerlocher_reit}
\begin{split}
\Delta\theta &= J_1^{\dagger^{\lambda_1}}\vec{e_1}+(J_2 P_{N(J_1)})^{\dagger^{\lambda_2}}(\vec{e_2}-J_2J_1^{\dagger^{\lambda_1}}\vec{e_1})\\
J^{\dagger^{\lambda}} &= (\sum_{i=1}^{r} \frac{\sigma_i}{\sigma_i^2+\lambda_\sigma^2})\textbf{v}_i\textbf{u}_i^T\ \ \ \  //\text{\textit{SVD}}\\
P_{N(J)} &= I-J^{\dagger}J\\
\lambda_\sigma &= \left\{
\begin{array}{ll}
\frac{d}{2} & \text{if}\ \sigma_{\text{min}}\leq\frac{d}{2}\\
\sqrt{\sigma_{\text{min}}(d-\sigma_{\text{min}})} & \text{if}\ \frac{d}{2}\leq\sigma_{\text{min}}\leq d\\
0 & \text{if}\ \sigma_{\text{min}}\geq d\\
\end{array}
\right.\\
d &= \frac{\norm{\vec{e}}}{b_{\text{max}}}
\end{split}
\end{equation} 

The simulations using DLS were ran using Skeleton C, for which each joint can rotate only 90$\degree$ to each side, therefore making it much more difficult to face orientations that are \textit{behind} the robot.
Using ERIK, that was not a problem given that we have the extension $\Xi_{\text{SymmetricEndpoint}}$, which allows the end-effector to be used upside-down.
While this feature is still used within DLS at the error function level, it is not properly considered by the actual algorithm.
We therefore also apply a correction to the orientation target in order to keep its up-side oriented in a way that it is reachable by the test skeleton given its joint limits.
This correction was the only one that we ever added to enhance the results for a particular skeleton or technique, and in fact, is used only to enhance the results of the technique to which we are comparing ERIK, for more realistic results.
Initially we considered the results of DLS too bad, and therefore the comparison (while optimistic for ERIK) was considered inappropriate.

The correction is made by flipping the target orientation's quaternion upside-down (i.e., performing a rotation of $\pi$ about the unit $\vec{Z}$ vector) in specific regions of the target space, so that we guarantee that the target's up-side is always directed to facilitate the result of DLS using Skeleton C, i.e., when the target is facing forward then its Y-axis will always be facing down; when the target is facing backward, then its Y-axis will always be facing up. 
Therefore the target is never an orientation that is mechanically unachievable a priori.

The classic Denavit-Hartenberg parameters used to model Skeleton C are presented in Table \ref{tab:DH_SkeletonC}.
\begin{table}[]
	\centering
	\begin{tabular}{l|cccc}
		Link & $\theta$ & $\alpha$ & a & d \\
		& (Joint Angle) & (Twist Angle) & (Link Length) & (Joint Offset) \\
		\hline\hline
		1 & 0 & $\frac{\pi}{2}$ & 0 & 10 \\
		\hline
		2 & $\frac{\pi}{2}$ & 0 & 30 & 0 \\
		\hline
		3 & 0 & $\frac{\pi}{2}$ & 30 & 0 \\
		\hline
		4 & $\frac{\pi}{2}$ & $\frac{\pi}{2}$ & 0 & 0 \\
		\hline
		5 & $\frac{\pi}{2}$ & 0 & 0 & 40
	\end{tabular}
	\caption{Denavit-Hartenberg parameters (classic) used to run the simulations of DLS on Skeleton C.}
	\label{tab:DH_SkeletonC}
\end{table}

\leavevmode \\
Finally, because the DLS technique outcome is very dependent on the maximum number of iterations execution, we also ran several trials with the technique using 100, 200, 400 maximum iterations.
Each will be referred to as e.g. DLS100 or DLS400. 
Whenever we refer solely to DLS, we will be referring to the best version of it (DLS400).

We started by running a set of simulations using DLS100\_nopost, which is DLS100 without the homogeneous solution, i.e., without the second part of the equation which attempts to solve for the target posture using the primary Jacobian's null space.
The goal with these simulations was to assess how well the DLS implementation was able to solve solely for orientation targets using Skeleton C, which is highly prone to singularities.
Although the technique we follow is stated to be free of \textit{algorithmic} singularities, the parametrization of the skeleton may also introduce \textit{kinematic} singularities (e.g. gimbal lock).
In fact, upon running the simulation using DLS100\_nopost, we found that there were many target orientations for which the algorithm became stuck yielding a very high orientation error, which we attribute to such type of singularities.
In order not to impair the results of the DLS simulations when compared to ERIK (which does not suffer from such singularities), we further used this simulation to filter the DLS results in order to remove all the samples for which the DLS100\_nopost version yielded an \textbf{orientation error} above 3x the specified threshold, thus excluding from the comparison exceptionally bad results that were not due to the posture constraint, but to inappropriate handling of kinematic singularities.
As such, in the comparison of ERIK and the DLS variants (Section \ref{sec:erik_dls}) the results from the DLS simulations presented are the results of applying such filter.

\begin{figure}[!htbp]
	\centering
	\begin{subfigure}{\linewidth}
		\includegraphics[width=\textwidth]{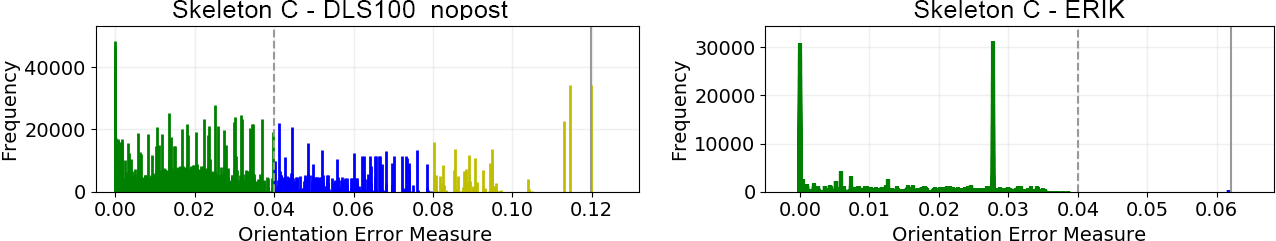}
		\caption{Orientation error histogram for skeleton C running ERIK compared to DLS100\_nopost.}
	\end{subfigure}
	\par\bigskip
	\begin{subfigure}{0.8\linewidth}
		\includegraphics[width=\textwidth]{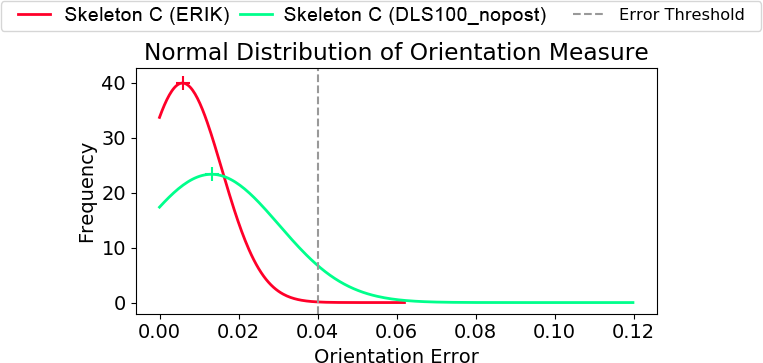}
		\caption{Normal distribution plots of the orientation error of ERIK compared to DLS100\_nopost.}
	\end{subfigure}
	\caption[Comparison of orientation errors of ERIK on Skeleton C to DLS100\_nopost.]{Comparison of orientation errors of ERIK on Skeleton C compared to DLS100\_nopost after filtering out samples with excessive orientation error in the latter.}
	\label{fig:erik_dls}
\end{figure}

Figure \ref{fig:erik_dls} compares the resulting orientation error histograms and normal distribution plots for ERIK and the filtered DLS100\_nopost  on Skeleton C.
Note that the data from ERIK actually resulted of the full simulation of ERIK with both orientation and posture targets.
Therefore we see that ERIK has performed better in solving the orientation constraint even through it also had the posture goal.
The DLS performed slightly below expectations, but we must consider that the simulation attempted to orient the end-effector to a very large big range of orientations in full 3D, i.e., pan, pitch and roll of the end-effector, towards the full vertical and horizontal 360\textdegree range.
The DLS results here have been corrected to match the number of samples of the ERIK one (which was also solved for each posture), therefore presenting a comparable scale; otherwise considering only the range of orientation targets, the DLS result would present far less samples and thus make these difficult to compare.

\subsubsection{Results}

Using ERIK, a total of 239 245 243 samples were simulated from 39 739 postures across all 7 skeletons.
The DLS was simulated for a total of 86 491 503 samples from 3789 postures using Skeleton C, in this case using three different maximum iteration counts.
However as explained in the previous section, the DLS results were further filtered resulting in a total of 61 684 497 selected samples.

The simulations were ran on a high-performance computer cluster (HPCC) containing a mix of nodes with AMD Opteron 6180 SE and 6344, and AMD EPYC 7401 CPUs, organized into nodes of either 48 or 96 CPUs.
In order to normalize and interpret the performance results from these simulations in comparison with a typical laptop CPU, we have searched for single-core benchmarks of these CPUs on community-sourced benchmark websites.
We took as an example the Intel i7-7700HQ, which is a popular CPU, featured in many mid and high-end personal laptops, and that is also at least 2 years old (launched Q1'2017) to represent an average laptop CPU.
Despite the multiprocessing capabilities of any of them, we were interested in the single-core performance, as each simulated sample ran as a single-core process, and are also expected to run as such in a real-world application (even if it is used within a multi-threaded/multi-core application, the IK engine per se should run sequentially in a single thread).

Table \ref{tab:cpus} shows the highest benchmark of each of these CPUs, using a Linux 64-bit system, as found on the community-sourced Geekbench website\footnote{\protect\url{https://browser.geekbench.com/} \urlDate}.
We consider these values to stand as an acceptable comparison of how the performance statistics collected through the HPCC compare to those of an average computer.
By considering this score instead of theoretical values such as MIPS or GFLOPS, we are also considering more of a general performance capability without considering particular architecture-wise optimizations.
For ERIK, all the simulations except the largest one were arbitrarily assigned to an Opteron node, which leads us to consider an average of both those CPUs for those simulations (these CPUs were distributed 50/50 among the total).
The largest simulation, for Skeleton G, was specifically assigned to an EPYC 7401 node.
For DLS, the simulations were arbitrarily assigned to any of the available nodes, being mostly attributed to an Opteron one. 
However in various cases the simulations were ran on an EPYC node. 
As such, for the DLS simulations we consider the weighted average benchmark score for all nodes, given that from a total of 672 CPUs, there were 192 EPYCs, and 240 of each of the Opteron types.

\begin{table}[!htbp]
	\centering
	\caption[Comparison of single-core performance of the CPUs used in the HPCC]{A comparison of the single-core performance of the CPUs used in the HPCC for the simulations, and how they related with the performance of a typical laptop CPU (ratio).}
	\label{tab:cpus}
	\begin{tabular}{lrr}
		\toprule
		CPU & Max Benchmark Score & Ratio \\
		\midrule
		Intel i7-7700HQ & 5341 & 1.0000 \\
		AMD Opteron 6180 SE & 1615 & 0.3024 \\
		AMD Opteron 6344 & 2233 & 0.4181 \\
		AMD EPYC 7401 & 3853 & 0.7214 \\
		AMD 6180 SE \& 6344 average & 1924 & 0.3602 \\
		ALL AMD - weighted average (4:5:5) & 2475 & 0.4634 \\
		\bottomrule
	\end{tabular}
\end{table}

The statistics regarding the whole procedure are summarized in Table \ref{tab:procedure_statistics}.
This table contains the number of postures and total samples ran for each skeleton (recall that each posture was simulated on 7609 target orientations). 
It additionally contains various run-time statistics regarding the execution time to process a single sample (posture-orientation pair), and on the number of iterations that were ran.
The execution time presented was corrected based on the ratios from Table \ref{tab:cpus} and therefore represent $\textit{measured time}\cdot\textit{ratio}$ in order to present all the statistics corrected as if they had all been ran on an average computer (taking an Intel i7-7700HQ as example).

\begin{table}[!htbp]
\centering
  \caption{Statistics regarding the evaluation experiments with a total of $\sim$~239M samples. Note that for the DLS cases, we present the total number of postures simulated, but the number of samples corresponds to the result of applying the filter explained in the previous section.}
  \label{tab:procedure_statistics}
  \setlength\tabcolsep{5pt}
  \begin{tabular}{crrr|rrrr|rrrr}
    \toprule
    \multirow{2}{*}{Skeleton}&\multicolumn{3}{c}{Number of}&\multicolumn{4}{c}{Iteration Count}&\multicolumn{4}{c}{Time per Sample (ms)}\\
    & DoFs & Postures & Samples & Min & Max & Mean & S.D. & Min & Max & Mean & S.D. \\
    \midrule
    A & 3 & 33 & 251 097 &       1 &  7 & 4.14 & 2.17 & 6	& 126	& 45 & 28\\
    B & 4 & 377 & 2 868 593 &    1 & 10 & 3.42 & 2.19 & 8	& 189	& 45 & 30 \\
    C & 5 & 3 789 & 28 830 501 & 1 & 12 & 2.31 & 2.08 & 6	& 1165  & 37 & 36 \\
    C-DLS100 & 5 & 3 789 & 20 561 499 & 1 & 100 & 63  &  43.6 & 1.60  & 2112  & 167 & 121 \\
    C-DLS200 & 5 & 3 789 & 20 561 499 & 1 & 200 & 120 &  92.8 & 1.57  & 3822  & 298 & 240 \\
    C-DLS400 & 5 & 3 789 & 20 561 499 & 1 & 400 & 234 & 191.6 & 1.55  & 7520  & 678 & 570 \\
    D & 5 & 3 789 & 28 830 501 & 1 & 13 & 2.02 & 1.67 & 10	& 458	& 36 & 30 \\
    E & 5 & 3 789 & 28 830 501 & 1 & 12 & 1.93 & 1.78 & 6	& 225	& 28 & 27 \\
    F & 6 & 8 305 & 63 192 745 & 1 & 13 & 1.61 & 1.46 & 13	& 368	& 35 & 35 \\
    G & 8 & 19 657 & 149 570 113 & 1 & 11 & 1.35 & 1.02 & 22	& 759	& 74 & 63 \\
  \bottomrule
\end{tabular}
\end{table}

\subsubsection{Analysis of Results: ERIK}

\begin{figure}[!htbp]
    \centering
    \begin{subfigure}{0.35\linewidth}
        \includegraphics[width=\textwidth]{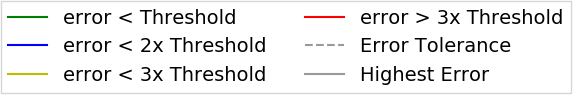}
    \end{subfigure}
\par\bigskip
    \centering
    \begin{subfigure}{\linewidth}
        \includegraphics[width=\textwidth]{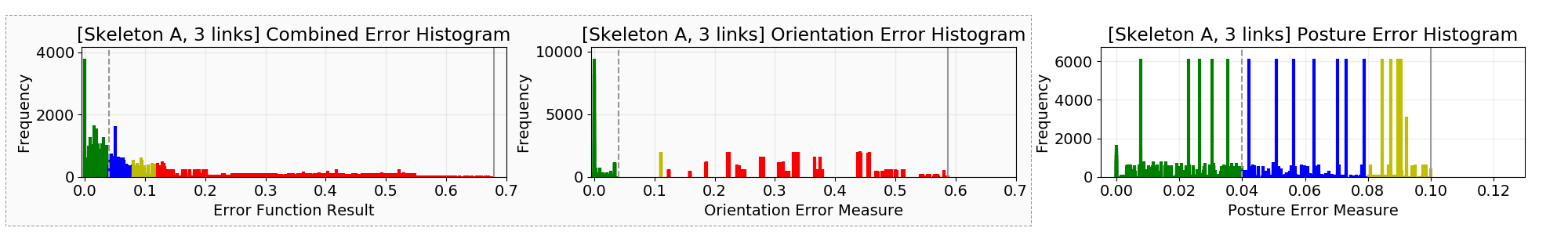}
        \par
        \includegraphics[width=\textwidth]{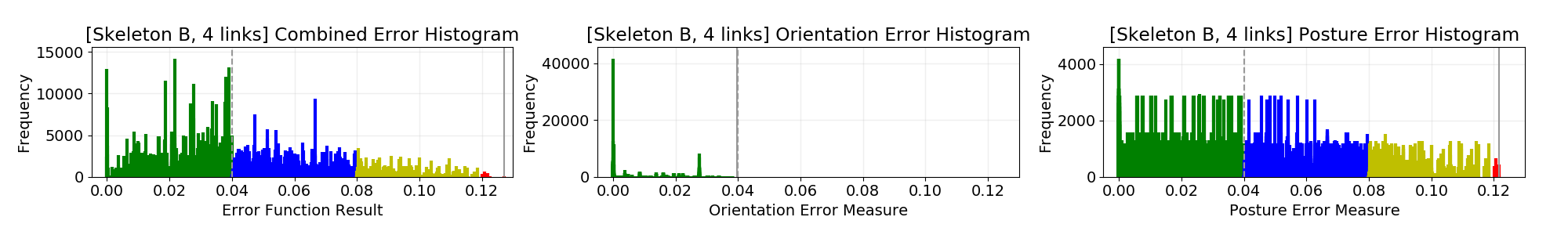}
        \caption{Results for skeletons A and B, the 3- and 4-link skeletons, which perform below expectations, but contribute to verify that the algorithm results the expected results for those cases.}
    \end{subfigure}
\par\bigskip
    \begin{subfigure}{\linewidth}
        \includegraphics[width=\textwidth]{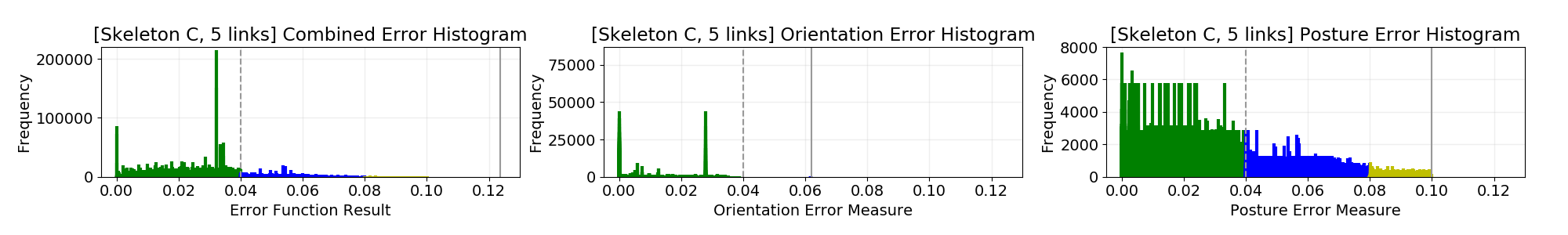}
		\par
        \includegraphics[width=\textwidth]{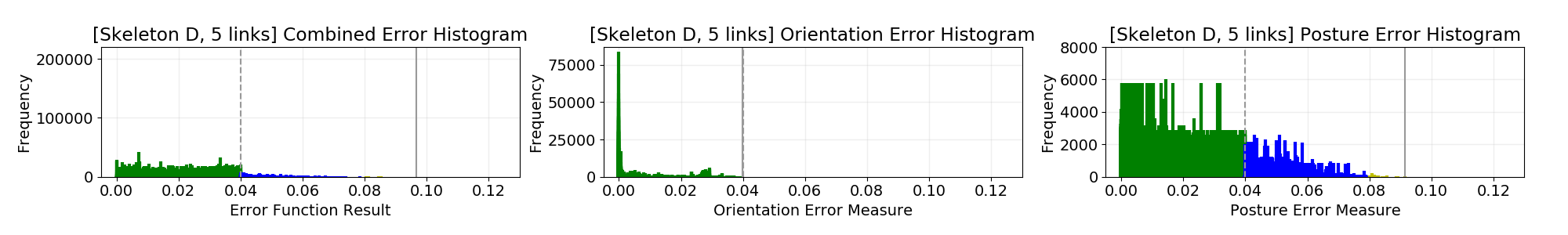}
		\par
		\includegraphics[width=\textwidth]{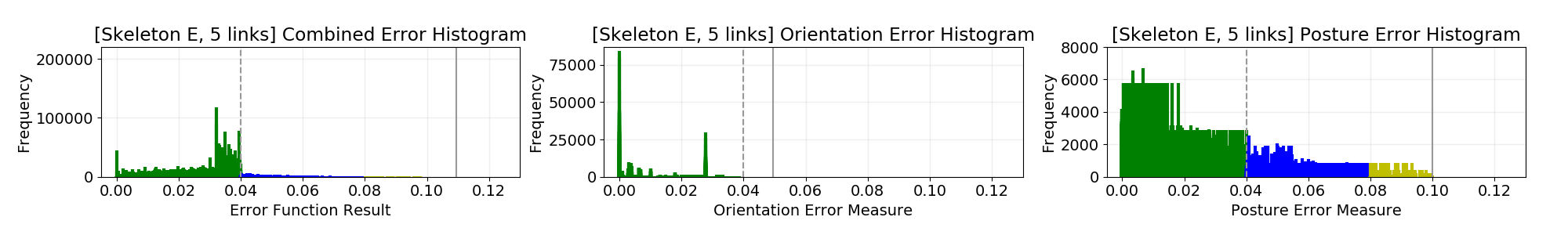}
		\caption{Results for skeletons C, D and E, the three different 5-link skeletons, which start to yield satisfactory results.}
	\end{subfigure}
\par\bigskip
	\begin{subfigure}{\linewidth}
		\includegraphics[width=\textwidth]{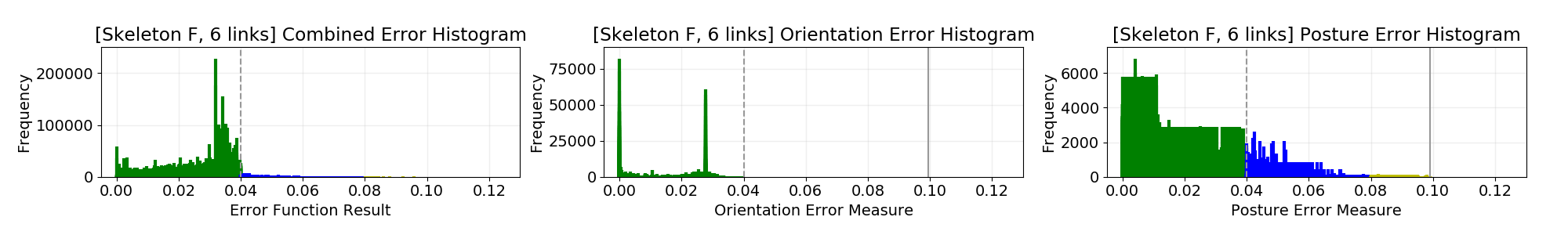}
		\par
		\includegraphics[width=\textwidth]{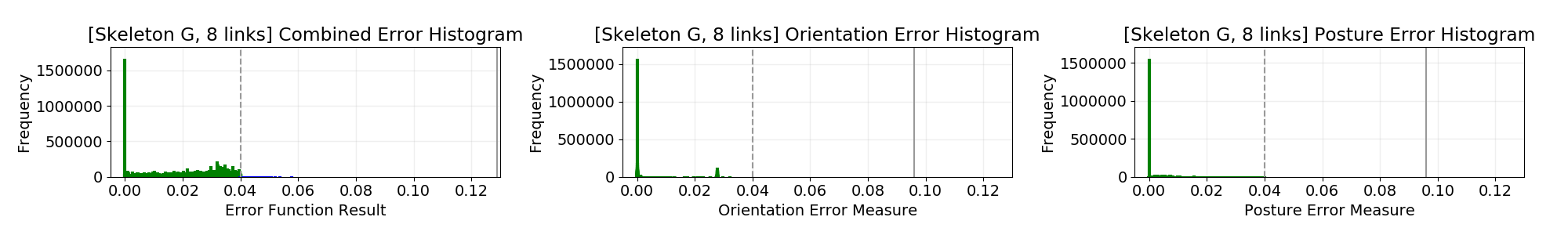}
		\caption{Results for skeletons F and G, the 6- and 8-link skeletons, which return the most satisfactory results.}
	\end{subfigure}
    \caption[Results of ERIK's evaluation process.]{Results of ERIK's evaluation process. Each line corresponds to one of the seven skeletons used. 
    	The columns correspond to each one's Error Function result, Orientation Error and Posture Error. 
    	Please note that the legend above the graphs applies to all of them.
        }\label{fig:error_histograms}
\end{figure}

After running the simulations on the different skeletons, we collected all the data and plotted the histogram for the error function and measures, as presented in Figure \ref{fig:error_histograms}.
Each line of the histogram figure represents an embodiment, from skeleton A to G, as indicated in the titles of the individual graphs.
The first column of graphs contains the results for the value of the (combined) error function $\Lambda_\phi$ for each final solution.
The second and third columns of graphs contain the final error for the individual measures $\epsilon_{\text{Orientation}}$ and $\epsilon_{\text{Posture}}$.
At the top of the figures matrix we have placed the Legend, which applies to all the graphs.

Each graph shows the distribution of the error for all the solutions.
The vertical axis represents the total count (frequency) of solutions that yielded a final error, given by the horizontal axis.
Note also the dashed vertical lines, which represent the intended maximum error ($\Lambda_{\text{Threshold}\varepsilon}$), and also the solid vertical line, which aids in the visualization of the data, by representing the maximum error produced within the graph's samples.
Note also that the range of the horizontal axis (error range) is the same in all rows except for the shaded ones in the first row, and that in the 5-link skeleton rows, each column presents the same Y value across all the three rows in order to help comparing between these cases.

\begin{figure}[!htbp]
	\centering
	\includegraphics[width=\textwidth]{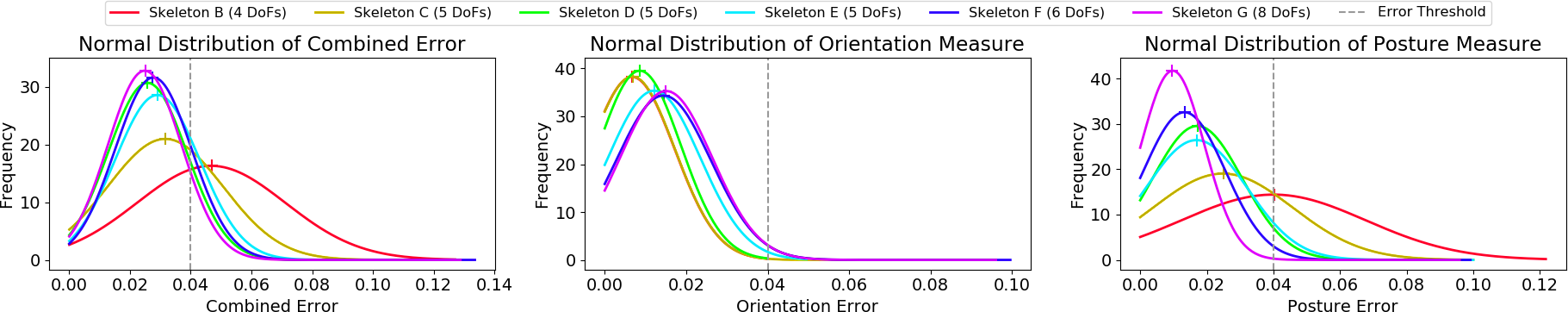}
	\caption[Normal distribution plots of the final combined error and error-measures]{Normal distribution plots of the final combined error and error-measures for each skeleton except skeleton A.}
	\label{fig:normal_plots}
\end{figure}

Plotting the normal distribution of the error function results for each skeleton, provides further support on the interpretation of the results beyond the individual histograms, as illustrated in Figure \ref{fig:normal_plots}.
This figure shows the normal distribution for the combined error and for each of the error measures, for each of the skeletons except for A, which, due to its large error, disrupts the presentation of the others (and does not provide a significant interpretational value).
The general interpretation taken from the normal plots is that all skeletons performed well regarding the Orientation Measure, and that the performance on the Posture Measure increased with the number of DoFs in the skeleton. Detailed interpretations will follow below.

\subsubsection{Results for \textbf{Skeleton A}} 
These show that the target orientation goal failed immensely. 
This was highly expected given that the mechanical limitations of its joints, with only one \textit{pitch} joint, limited to $[-\frac{\pi}{2}, \frac{\pi}{2}]$, would not allow it to aim at any orientations below the horizon. 
We also see that the posture errors do not seem so bad - that is because being a single-segment embodiment, the single and only posture it can perform is a straight line. 
Given that the whole corpus of experiment data would also generate only straight postures to be tested, it ended up not performing \textit{so bad} there.
Despite that, this case was meant to test if the algorithm reflected the expected results on such a constrained embodiment, with nearly no possibility of performing expressive postures while aiming at a given direction.
The results confirm our hypothesis.

\subsubsection{Results for \textbf{Skeleton B}} 
These show a substantial decrease in error compared to Skeleton A. 
By adding one more DoF, and allowing each DoF to have a higher range of motion, the skeleton was able to aim even at orientations below the horizon, as can be seen in its Orientation Error histogram, which always produced an error below the threshold.
However, in order to achieve all the target orientations, the posture goal was largely missed, as seen in its Posture Error histogram.
The normal plot shows that the error for Skeleton B (in red) was largely distributed beyond the specified threshold on the Posture Measure, and consequently on the Combined Error. However the graph for the Orientation Measure shows a good performance (its curve overlaps with the one of Skeleton C, in yellow). 
Still this skeleton does not represent a useful use-case for ERIK - instead it provides further support over the validity of the algorithm and its evaluation, as its \textit{bad} results go in line with our expectation.

\subsubsection{Results for \textbf{Skeletons C, D, E}} By adding another DoF, these results show lower error values compared to those of Skeleton B.
We can note that in particular, the maximum Posture Error has decreased, meaning that the extra DoF provided the character with the ability to perform more expressive postures towards any direction.
In C and E, some Orientation Error outliers have however produced an error above the intended threshold.
However, it seems that there were very few of these situations, which makes them nearly imperceptible in the graph, if it wasn't for the \textit{Highest Error} line.

These skeletons start to yield results as we expect: to successfully orient to any given direction, while holding an arbitrary expressive posture that is allowed to slightly distort in order to ensure the prioritized orientation constraint. 
We take these conclusions from the Orientation Error histogram, which contains only some outliers beyond our given error threshold, and by the fact that the majority of the Posture Error is within the threshold, and that the ones that were distorted beyond it are contained within at most 3x that threshold, with frequency decreasing as the error increases.

Is is however interesting to perform a comparison between these three 5-link skeletons.
Although they all have the same number of DoFs, they are configured in different ways.
As seen in Table \ref{tab:test_skeletons}, skeleton C has an YXXZY configuration, while skeletons D and E use an YXZXY configuration.
Furthermore, the angular limits of skeleton D are $[-\pi, \pi]$ while skeletons C and E have limits $[-\frac{\pi}{2}, \frac{\pi}{2}]$.
The normal plots make this comparison more explicit. It becomes clear that from these three (C in yellow, D in green, E in cyan), all performed approximately well in the Orientation Measure, with Skeleton D performing best in the Posture Measure and the Combined Error, where Skeleton C performed worst.
This draws the conclusion and illustrates that 1) a different joint configuration such as between C and E affects the performance, with, in this case, the layout of E providing better expressive capabilities than the one of C, while also showing, as expected, that by providing a wider range of motion, as in D versus E, that D, the one with the wider motion, can also perform better.

\subsubsection{Results for \textbf{Skeleton F}} 
By introducing just one additional DoF as compared to C, D and E, the algorithm increases its performance.
It is interesting here to compare in particular skeleton F to skeleton D, being that F has a lower angular range than D, but an additional DoF.
While it may seem unclear from the histograms which of the two performed best, the normal plots does elucidate that Skeleton F performs better as seen in the normal distribution of the Combined Error, and of the Posture Measure. Interestingly skeleton D performed better in the Orientation Measure, however both performed within the threshold.

\subsubsection{Results for \textbf{Skeleton G}} 
Finally, Skeleton G, with 8 links shows the best results as can be clearly seen in both the histograms and the normal plots.
Again, through the normal plots it is seen to be not the best performer on the Orientation Measure, however, its ability to perform well on that measure, and perform exceptionally on the Posture Measure make it the best from this case set.

\subsubsection{Results Comparison} 
The results presented here confirm our initial hypothesis that, as long as an embodiment has enough DoFs, it is able to use ERIK orient its endpoint towards any given target orientation, while successfully portraying a given expressive posture with minimal disruption.
We group the results in three groups. Skeletons A and B can be seen as proofs of concepts, that serve to show that the algorithm fails when and how we expect it to fail (in highly constrained skeletons, with very few DoFs). Skeletons C, D, E and F are representative of cases where the algorithm starts to show positive results - with 5 or 6 DoFs it is mostly able to comply with all the constraints we have imposed, such as the joint limits and the error threshold, when solving for an integrated posture-orientation goal.
Finally, skeleton F, with 8 DoFs already represents a case where the problem is solved in the most acceptable way, with both error measures performing below the threshold for nearly all the tested samples.

\subsubsection{Analysis of Results: ERIK vs DLS}
\label{sec:erik_dls}
\begin{figure}[!htbp]
	\includegraphics[width=\textwidth]{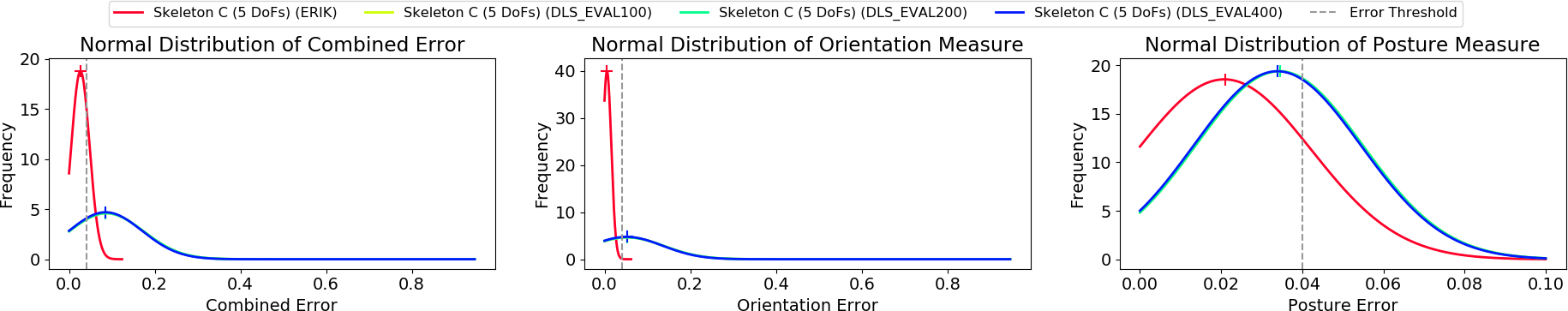}
	\caption{Comparison of the normal distribution plots of the errors for each DLS version and for ERIK Skeleton-C.}
	\label{fig:dls_comparison}
\end{figure}

The same procedure was followed for analysing the results of the DLS technique.
Figure \ref{fig:dls_comparison} shows the normal distribution plots of the errors compared to the ones of ERIK with the same skeleton.
Here we find that despite the improvements, there was nearly no difference in the general distribution of the errors across the different variations of DLS.
In fact the three curves nearly overlap and become indistinguishable.
We additionally detail the mean value and standard deviation for each of the cases in Table \ref{tab:erik_dls_mean_sd}, which shows that there was a very slight improvement in the errors as the maximum number of iterations was increased.

In particular, and as we had already foreseen, ERIK performed better in achieving the correct target orientation.
What we were most interested in finding out was how the posture error of the DLS would perform.
Here we find similarly shaped curves for both ERIK and DLS, although ERIK's curve is centred around a lower mean error, thus revealing that it did in fact also perform better than DLS on solving the the posture target.

\begin{table}[hbp]
	\centering
	\begin{tabular}{lrr|rr|rr}
		\multirow{2}{*}{} & \multicolumn{2}{c}{Combined Error} & \multicolumn{2}{c}{Orientation Error} & \multicolumn{2}{c}{Posture Error} \\
		& \multicolumn{1}{c}{Mean} & \multicolumn{1}{c}{S.D.} & \multicolumn{1}{c}{Mean} & \multicolumn{1}{c}{S.D.} & \multicolumn{1}{c}{Mean} & \multicolumn{1}{c}{S.D.} \\ 
		\cline{2-7}
		\multicolumn{1}{l|}{ERIK} & 0.026614 & 0.021258 & 0.005819 & 0.009977 & 0.020795 & 0.021520 \\
		\multicolumn{1}{l|}{DLS100} & 0.087067 & 0.086930 & 0.052744 & 0.085997 & 0.034323 & 0.020593 \\
		\multicolumn{1}{l|}{DLS200} & 0.086973 & 0.086897 & 0.052652 & 0.085980 & 0.034321 & 0.020591 \\
		\multicolumn{1}{l|}{DLS400} & 0.086831 & 0.086581 & 0.052523 & 0.085669 & 0.034308 & 0.020579
	\end{tabular}
	\caption{Mean value and Standard Deviation for the ERIK and DLS variants comparison.}
	\label{tab:erik_dls_mean_sd}
\end{table}

\label{sec:evaluation}
\section{Discussion}
The \erik\ technique is a promising new step in the field of character animation, especially for robots and other interactive and immersive characters that are driven by AI.
When driven by such AIs, and/or subject to stimuli such as user perception, it is important that the character animation engines for real-time, interactive characters, are able to process the flow of information that arrives through its sensors, and use it to influence and drive the character's behavior and animation.
Our work takes an important step in that direction as the results support our initial claim that \erik\ is able to provide expressive inverse kinematics solutions in real-time which simultaneously solve for an expressive posture goal, and for a target orientational goal.

Aiming at characters that are driven in real-time, and need to be expressive while also using their body to interact, such as gaze-tracking a person or object,
\erik\ succeeds in tackling both goals simultaneously for the majority of the situations.
It was expected that by having more DoFs in the embodiment, both goals could be solved with lower error measures.
In average, for a 5-link skeleton, the algorithm took 34ms to calculate a solution, and 74ms for a more complex 8-link skeleton, yielding a solution rate of 30 and 14 Hz respectively.
While it would be desirable to have higher performance rates, our own implementation has show it to be adequate for real-time applications, as long as the IK solver is not synchronously running with the output module.
By using the Nutty Motion Filter on the output module to smoothly interpolate the IK solutions in real-time, we are able to achieve smooth, sustained motion that can be used in such applications.

We have tested various skeleton configurations and ran extensive simulations in order to validate our claims.
It is arguable how such an evaluation should be performed, however, in order to provide a general view, we opted out of evaluating the use of a robot using ERIK in a particular application with a smaller set of expressive postures, as that would also confine the validity of any conclusions to that single embodiment and set of postures.
We therefore outlined the requirements that should be met by the algorithm to allow it to be used in any application, with any posture and with an arbitrary skeleton layout.
Instead of using a small set of animator-designed postures, we took a sample of all the possible postures that each skeleton would allow to design.
Instead of measuring how well a result met an animator's expectations (which is a subjective evaluation), we measured how close the resulting postures were to the original posture in terms of shape, using a heuristic method (the Posture Measure).
By ensuring that the resulting posture is similar to the original, which in a real-world application, would be given by an animator, we expect and claim that the animator would also find the resulting posture satisfactory.

We additionally compared the results of the ERIK simulations to the same simulations using the DLS technique with postural control as the secondary task.
Results showed that ERIK performed better in both the individual orientation task and the posture task.
We argue that the DLS simulations using the tested Skeleton C are prone to kinematic singularities, which may have not been properly addressed.
Therefore the result comparison was made with a filtered version of the DLS results, in order to excluded samples that seemed excessively bad due to that issue.
The filtered results still show a worse performance compared to ERIK.
Furthermore, even if all the kinematic singularities were properly dealt with in every case, the DLS simulations performed significantly slower than ERIK.
Therefore, for the problem that we specify, ERIK represents a substantial improvement against using the DLS technique with the secondary task for posture-control, given that:
\begin{itemize}
	\item Using ERIK one can use any embodiment even if highly redundant, which would be prone to kinematic singularities using Jacobian methods;
	\item ERIK performs on average about 4.5x faster than DLS on a similar task.
\end{itemize}
\section{Future Directions}
At the moment we have only implemented support for embodiments that are structured as a single chain of DoFs with arbitrary rotations axes and limits.
In the future we will bring in more features that are part of the FABRIK technique, such as the support for multiple end-points, which allows for full-body control of e.g humanoids.
We will also develop support for prismatic (sliding) joints, which are more commonly found in robots.
One of the reasons why FABRIK was chosen as the building block for ERIK was because of all those features that it supports by extension, which could later be introduced into ERIK as well.
Currently we identify no theoretical obstacle on the implementation of such features.

One additional potential direction for ERIK is the support of full \textit{out-of-the-box} animation warping in real-time, i.e., given not just a posture, but an animated motion, to have that motion warped so that the end-effector aims at a given direction, again, in real-time, for an arbitrary embodiment, and without any training or machine learning technique required.
At the moment there is theoretically no obstacle to such a feature, hindered only by the performance of the algorithm.
Because ERIK operates at an unstable rate, with an average of e.g. 14Hz and peaks of 1.3Hz on an 8-link skeleton, we would be unable to guarantee the proper execution of an animation that runs faster than 1.3fps across its whole timeline.
Further optimization and development is required to achieve performance rates that allow ERIK to be used seamlessly on animations as well as on postures.

\section*{ACKNOWLEDGEMENTS}
This work was supported by national funds through FCT - Funda\c{c}\~{a}o para a Ci\^{e}ncia e a Tecnologia with references UID/CEC/50021/2019 and SFRH/BD/97150/2013. A special thank you goes to Sergio Almeida from CENTRA - Center for Astrophysics and Gravitation of Instituto Superior T\'{e}cnico, who provided access and support to the High-Performance Computing Cluster used on the algorithm's evaluation.
\bibliographystyle{unsrt}  
\bibliography{main}
\clearpage
\SetKwProg{LoopI}{LoopI(}{) do}{end}
\SetKwProg{ForJoint}{ForJoint}{ as $k$ do}{end}
\SetKwFor{uWhile}{while}{do}{}

\ifboolexpr{togl {Thesis}}{
\chapter{ERIK Algorithm}
	\section{Algorithmic Specification}
	\label{sec:Algorithm}

One of this paper's major contribution is to share the full algorithmic specification that allows to implement \erik.
However it would be impractical and even unintelligible to present the whole algorithm here in detail.
Instead we have done our best to describe in detail only the major parts of it, while presenting either a textual description or a mathematical formulation for the parts that are less particular to \erik, and which may be understood and implemented by someone with appropriate CGI animation knowledge.

Aiming at a more comprehensible reading experience, we have shifted all the detailed algorithms to the next section of the appendix, \ref{sec:Algorithm2}.
The entry point to the algorithm is the \textbf{Calculate\erik} function, outlined in Algorithm~\ref{alg:CalculateERIK}.
This function takes as input the \erik\ Parameters ($\Pi$), and Hyperparameters ($\Lambda$), which have been described in detail in Section \ref{sec:erik_model_specification}.


Some of the macros or functions used in the algorithms are briefly described in the next section. 
For simplicity, all the quaternions used are rotation quaternions, i.e, quaternions of unit length.
As an additional reminder, please note that the child link of the end-point link refers to the posture's SuperPoint (Section \ref{sec:erik_model_superpoint}).

\subsection{Description of functions used throughout the algorithms}
\label{sec:alg_functions}
This section outlines a short description and/or mathematical formulation for some of the auxiliary functions and operations used within \erik.
\begin{description}
	\item[EmptySolution($Sk$)] Return an empty solution for skeleton $Sk$.
	\item[SafeAngle($k, \theta, bCycle=\mathit{False}$)] Returns $\theta'$ as an angle that is safe for joint $k$ given its maximum and minimum angle limits, while allowing the angle to cycle instead of purely clamping:
	\[
	\theta' = \left\{
	\begin{array}{ll}
	min(k_{\text{Max}\theta}, max(k_{\text{Min}\theta}, 2\pi\cdot(\frac{\theta}{2\pi}-\round{\frac{\theta}{2\pi}}))) & \mathit{if bCycle} \\
	min(k_{\text{Max}\theta}, max(k_{\text{Min}\theta}, \theta)) & \mathit{otherwise}
	\end{array}
	\right.
	\]
	\item[SetOriFromParent($k, \Theta$)] Sets $k$'s basis orientation from its parent:
	\[
	\solGO{k} = \left\{
	\begin{array}{ll}
	\solGO{\parent{k}}\cdot\solLO{\parent{k}} & \mathit{if not}\ \isRoot{k}\\
	\textit{I}  & \mathit{otherwise}\\
	\end{array}
	\right.
	\]
	\item[SetPosFromParent($k, \Theta$)] Sets $k$'s basis position from its parent:
	\[
	\solPos{k} = \left\{
	\begin{array}{ll}
	\solPos{\parent{k}}+\RotVQ{\segment{\parent{k}}}{\solGO{k}} & \mathit{if not}\ \isRoot{k}\\
	\vec{0}  & \mathit{otherwise}\\
	\end{array}
	\right.
	\]
	\item[SetFrameFromParent($k, \Theta$)] Call $SetOriFromParent(k, \Theta)$ and $SetPosFromParent(k, \Theta)$ and returns the new $\Theta$.
	\item[ApplyFK($\Theta, k=\Root{\Theta}$)] Performs Forward Kinematics calculus on solution $\Theta$ starting from node $k$ (optional).
	\item[RotVQ($\vec{v}, Q$)] Returns vector $V$ rotated by quaternion $Q$.
	\item[QAA($\vec{v}, \alpha$)] Returns a normalized quaternion that represents a rotation of $\alpha$ about the axis $\vec{v}$ (axis-angle).
	\item[VDiffAsQ($\vec{v_1}, \vec{v_2}$)] Returns the orientation difference between $\vec{v_1}$ and $\vec{v_2}$ as a Quaternion.
	\item[QDiff($Q_1, Q_2$)] Returns rotation difference between $Q_1$ and $Q_2$ as a Quaternion.
	\item[TBasis($k,Q')$] Transforms the world-space basis of link $k$ globally by $Q'$ ($k_Q = Q'\cdot k_Q$).
	\item[TBasisRoll($k,Q')$] Transforms the world-space basis of link $k$ locally by $Q'$ ($k_Q = k_Q\cdot Q'$).
	\item[VecAngle($\vec{v_1}, \vec{v_2}$)] Angle $\theta$ between $\vec{v_1}$ and $\vec{v_2}$ using atan2.
	\item[VecAngle($\vec{v_1}, \vec{v_2}, \vec{r}$)] Angle $\theta$ between $\vec{v_1}$ and $\vec{v_2}$ using atan2 with $sign(\theta) = -1\ \textbf{if}\ r\cdot (\vec{v_1} \times \vec{v_2})<0\ \textbf{else}\ 1$.
	\item[LALUT($k, \lambda$)] Queries joint $k$'s LALUT for latitude $\lambda$.
	\item[TargetLatitude($k, \vec{\tau}$)] Calculates the latitude $\lambda$ for target $\vec{\tau}$ on $k$'s joint model:
	\[
	\varsigma = -1\ \textbf{if}\ \hat{\tau}\cdot\POA{k}<0\ \textbf{else}\ 1
	\]
	\[
	\lambda=max(k_{\text{Bottom}\lambda}, min(k_{\text{Top}\lambda}, \frac{\hat{\tau}\cdot\uvecY+1}{2}))
	\]
	\item[PitchRA($k$)] Returns $\vec{r}$ such that:
	\[
	\vec{r} = \left\{
	\begin{array}{ll}
	\child{k}_{RA} & \textbf{if}\ \IsTwister{k}\ \textbf{and}\ \isRoot{k}\\
	\parent{k}_{RA} & \textbf{if}\ \IsTwister{k}\ \textbf{and}\ \textbf{not}\ \isRoot{k}\\
	k_{RA} & \mathit{otherwise}
	\end{array}
	\right.
	\]
	\item[EPA($Q, \vec{u}$)] Ensures Positive Axis on quaternion $Q$ based on a rotation axis $\vec{u}$:
	\[ 
	EPA(Q) = -Q\ \textbf{if}\ \vec{u}\cdot \vec{Q_v} < 0\ \textbf{else}\ Q 
	\]
	\item[IsTwister($L$)] Returns $\mathit{True}$ if link $L$'s rotation axis is aligned with its own segment.
	\item[AvoidJointEdge($k, \delta$)] Performs an angular offset of $\pm\delta$ on joint $k$ if it is currently at its minimum ($+\delta$) or maximum ($-\delta$) value.
	\item[AvoidPostureJointEdges($\Psi, \delta$)] Runs $AvoidJointEdge(k, \delta)$ on each joint $k$ in posture $\Psi$.
	\item[BWCD($\Psi|\Theta, \vec{\tau}, \Lambda$)] Performs BWCD (Section \ref{sec:BWCD}) on posture $\Psi$ or solution $\Theta$, towards the target direction $\vec{\tau}$.
	\item[CCD($\Theta, \vec{\tau}, \Lambda$)] Finds a new solution that turns the current solution $\Theta$'s end-point towards direction $\vec{\tau}$ using CCD.
	\item[NonConversionDetected($aux, \Lambda$)] Checks whether or not the current $aux_{\Theta_{\varepsilon}}$ is converging towards a fixed value or behaving as a cyclic function, and therefore not converging.
	\item[NonConvOffsetTrick($aux, k$)] Shifts the current Target Orientation to a slightly different direction choosing link $k$ and child as the expected offset solvers, by applying:
	\[
	\Omega(\Theta, k, \delta) = \left\{
	\begin{array}{ll}
	QAA(RotVQ(k_{RA}, \Theta_{k_Q}), \delta) & \textbf{if} \\ & |\Theta_{k_\theta}-k_{\text{Min}\theta}|>\\&|\Theta_{k_\theta}-k_{\text{Max}\theta}| \\
	QAA(RotVQ(k_{RA}, \Theta_{k_Q}), -\delta) & \mathit{otherwise}
	\end{array}
	\right.
	\]
	\begin{flalign*}
	\delta&= \Lambda_{\text{Disturbance}\theta}\\
	k&= \Lambda_{\Root{\text{Sk}}}\\
	aux_\tau&= \Omega(aux_\Theta, k_{\text{Child}}, \delta) \cdot \Omega(aux_\Theta, k, \delta) \cdot aux_\tau\\
	aux_{\text{TriedNonconvOffset}}&)= True
	\end{flalign*}

	\item[SelectBestSolution($aux$)] Returns $\Theta$ such that:
	\[
	\Theta = \left\{
	\begin{array}{ll}
	aux_{\Theta} & \textbf{if}\ aux_{\Theta_{\varepsilon}} \leq aux_{\text{best}\Theta_{\varepsilon}} \\
	aux_{\text{best}\Theta} & \mathit{otherwise}
	\end{array}
	\right.
	\]
	
\end{description}

}{
	\section{Algorithmic Specification}
	\label{sec:Algorithm}
	
}
\pagebreak
\section{Detailed Algorithms}
\label{sec:Algorithm2}
\begin{figure}[H]
	\centering
	\includegraphics[width=0.8\columnwidth]{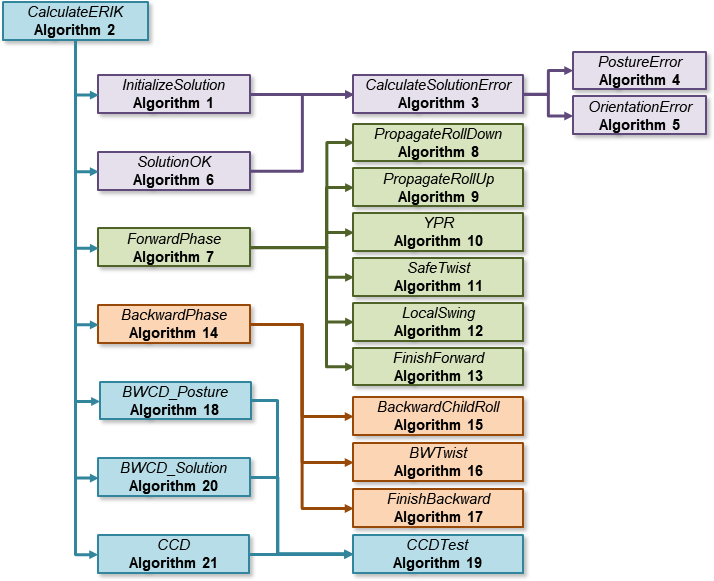}
	\caption[A map of the algorithmic description of ERIK]{A map of the algorithmic description of ERIK, to be used as a visual index throughout this section. 
		Each arrow means that the algorithm from where it departs invokes the algorithm at which it arrives.}
	\label{fig:alg_map}
\end{figure}

We start by presenting a map of the algorithms in Figure \ref{fig:alg_map}, which serves as a visual index to know where to find each of the pieces, and where they are invoked.
Please refer back to Tables \ref{tab:erik_params}--\ref{tab:alg_symbols} in Section \ref{sec:erik_model_specification} and to the description of functions in Section \ref{sec:alg_functions} while following or implementing these algorithmic descriptions.

\SetKwFunction{BackwardChildRoll}{BackwardChildRoll}%
\SetKwFunction{BackwardLocalTwist}{BWTwist}%
\SetKwFunction{BackwardPhase}{BackwardPhase}%
\SetKwFunction{BWCDPosture}{BWCD\_Posture}%
\SetKwFunction{BWCDSolution}{BWCD\_Solution}%
\SetKwFunction{CalculateSolutionError}{CalculateSolutionError}%
\SetKwFunction{CCD}{CCD}%
\SetKwFunction{CCDTest}{CCDTest}%
\SetKwFunction{FinishBackward}{FinishBackward}%
\SetKwFunction{FinishForward}{FinishForward}%
\SetKwFunction{ForwardPhase}{ForwardPhase}%
\SetKwFunction{InitializeSolution}{InitializeSolution}%
\SetKwFunction{LocalSwing}{LocalSwing}%
\SetKwFunction{OrientationError}{OrientationError}%
\SetKwFunction{PostureError}{PostureError}%
\SetKwFunction{PropagateRollDown}{PropagateRollDown}%
\SetKwFunction{PropagateRollUp}{PropagateRollUp}%
\SetKwFunction{SafeTwist}{SafeTwist}%
\SetKwFunction{SolutionOK}{SolutionOK}%
\SetKwFunction{YPR}{YPR}%


\begin{algorithm}
	\caption{InitializeSolution\label{alg:InitializeSolution}}
	\SetKwInOut{Input}{input}\SetKwInOut{Output}{output}
	
	\DontPrintSemicolon
	\Input{$\tau, \Psi, \Lambda$\tcp*[f]{Orientation, Posture, Hyperparams}}
	\Output{aux\tcp*[f]{container of execution variables}}
	\Begin{
		$aux \leftarrow \varnothing$\;
		$aux_\tau\leftarrow \tau$\tcp*[f]{save a working copy of orientation}\;
		\uIf{$\IsTwister{\Lambda_{\EE{\Sk}}}$}{
			$aux_\tau\leftarrow \tau\cdot\RotVQ{\uvecY}{\Psi_{EndPoint_\theta}}$\;
		}
		$aux_\Psi\leftarrow\Psi$\tcp*[f]{save a working copy of posture}\;
		$aux_{previous\Theta}\leftarrow \EmptySolution{\Lambda_{\Sk}}$\;
		\CalculateSolutionError{$aux_{\text{previous}\Theta}, \tau, \Psi, \Lambda_\phi$}\;
		$aux_{best\Theta}\leftarrow aux_{\text{previous}\Theta}$\;
		$aux_{\Theta}\leftarrow aux_{\text{previous}\Theta}$\;
		\Return{$aux$}
	}
\end{algorithm}

\begin{algorithm}
	\caption{Calculate\erik\label{alg:CalculateERIK}}
	\SetKwInOut{Input}{input}\SetKwInOut{Output}{output}

	\SetKw{inif}{if}
	\SetKw{inelse}{else}
	
	\DontPrintSemicolon
	\Input{$\Pi, \Lambda$\tcp*[f]{Parameters, Hyperparameters}}
	\Output{$\Theta$\tcp*[f]{Solution}}
	\Begin{
		$aux \leftarrow$ \InitializeSolution{$\Pi_\tau, \Pi_\Psi, \Lambda$}\;
		$aux_\Psi \leftarrow$ \BWCDPosture{$aux_\Psi, aux_{\vec{\tau}}, \Lambda$}\;
		\uIf{$\Lambda_{\Xi_{\text{AvoidJointEdges}}}$}{
			$\AvoidPostureJointEdges{aux_\Psi}{\Lambda_{\text{Disturbance}\theta}}$\;
		}
		\For{$i\leftarrow 1$ \KwTo $\Lambda_{\text{MaxERIKIterations}}$}{
			\For{$k\leftarrow \Lambda_{\EE{\Sk}}$ \KwTo $\Lambda_{\Root{\Sk}}$}{
				$\tau \leftarrow aux_\tau$ \inif{$\isEE{k}$} \inelse{$aux_{\Theta_{k_{\child{Q}}}}$}\;
				\ForwardPhase{$k, \tau, aux_\Psi, aux_\Theta, \Lambda$}
			}
			\For{$k\leftarrow \Lambda_{\Root{\Sk}}$ \KwTo $\Lambda_{\EE{\Sk}}$}{
				\BackwardPhase{aux, $k, \Pi, \Lambda$}
			}
			\uIf{\SolutionOK{$\Pi, \Lambda$, aux}}{
				\Return{$aux_\Theta$}
			}
			$aux_\Theta \leftarrow$ \BWCDSolution{$aux_\Theta, aux_{\vec{\tau}}, \Lambda$}\;
			\uIf{\SolutionOK{$\Pi, \Lambda$, aux}}{
				\Return{$aux_\Theta$}
			}
			\uIf{$\NonConversionDetected{aux}{\Lambda}$}{
				\uIf{$\Lambda_{\Xi_{\text{NonConvOffsetTrick}}}$ \textbf{and} $\lnot aux_{\text{TriedNonconvOffset}}$}{
					$aux \leftarrow \mathit{\NonConvOffsetTrick{aux}{\Lambda}}$\;
					\textbf{continue}
				}
				\uElseIf{$\Lambda_{\Xi_{\text{NonConvCCDTrick}}}$}{
					
					$aux_\Theta \leftarrow$\CCD{$aux_\Theta, aux_\tau, \Lambda$}\;
					\uIf{\SolutionOK{$\Pi, \Lambda$, aux}}{
						\Return{$aux_\Theta$}
					}
					\uElse{
						$aux \leftarrow$ \CCD{$\EmptySolution{\Lambda_{\Sk}}, aux_\tau, \Lambda$}\tcp*[r]{Try from a new empty solution.}
						\uIf{\SolutionOK{$\Pi, \Lambda$, aux}}{
							\Return{$aux_\Theta$}
						}
					}
				}
				\Return{$\SelectBestSolution{aux}$}
			}
		}
		\Return{$ \SelectBestSolution{aux} $}
	}
\end{algorithm}

\begin{algorithm}[hbtp]
	\caption{CalculateSolutionError\label{alg:CalculateSolutionError}}
	\SetKwInOut{Input}{input}\SetKwInOut{Output}{output}

	\DontPrintSemicolon
	\Input{$\Theta, \tau, \Psi, \Lambda$\tcp*[]{\small Solution, Target Orientation, Target Posture, Hyperparameters}}
	\Output{The solution's combined error $\Theta_\epsilon$}
	\Begin{
		$\epsilon \leftarrow \Lambda_{\text{OrientationErrorWeight}}\cdot$ \OrientationError{$\EE{\Theta}, \tau, \Lambda$} + $\Lambda_{\text{PostureErrorWeight}}\cdot$ \PostureError{$\Theta, \Psi, \Lambda$}\;
		\Return{$\epsilon$}
	}
\end{algorithm}

\begin{algorithm}[hbtp]
	\caption{OrientationError\label{alg:ef_orientation}}
	\SetKwInOut{Input}{input}\SetKwInOut{Output}{output}
	\DontPrintSemicolon
	\Input{$\Phi, \tau, \Lambda$\tcp*[]{\small End-Effector Solution, Target Orientation, Hyperparameters}}
	\Output{The solution's orientation error $\epsilon_{\text{Orientation}}$}
	\Begin{
		$\omega_\epsilon \leftarrow \frac{\text{min}(\lvert\tau - \Phi_{{\text{EE}}_\Omega}\rvert, \lvert\tau + \Phi_{{\text{EE}}_\Omega}\rvert)}{\sqrt{2}}$\tcp*[]{\small Absolute distance between quaternions} 
		\uIf{$\Lambda_{\Xi_{\text{SymmetricEndpoint}}}$}{
			$\xi\leftarrow\QAA{\RotVQ{\uvecY}{\Phi_{{\text{EE}}_\Omega}}}{\pi}\cdot\Phi_{{\text{EE}}_\Omega}$\tcp*[]{\small Rotate symmetrically about its rotation axis} 
			$\xi_\epsilon \leftarrow \frac{\text{min}(\lvert\tau - \xi\rvert, \lvert\tau + \xi\rvert)}{\sqrt{2}}$\tcp*[]{\small Absolute distance between quaternions} 
			$\omega_\epsilon\leftarrow \text{min}(\omega_\epsilon, \xi_\epsilon)$
		}
		\Return{$\omega_\epsilon$}
	}
\end{algorithm}

\begin{algorithm}[hbtp]
	\caption{PostureError\label{alg:ef_posture}}
	\SetKwInOut{Input}{input}\SetKwInOut{Output}{output}
	\DontPrintSemicolon
	\Input{$\Theta, \Psi, \Lambda$\tcp*[]{\small Solution, Target Posture, Hyperparameters}}
	\Output{The solution's posture error measure $\epsilon_{\text{Posture}}$}
	\Begin{
		$\epsilon\leftarrow0,\ \ \ \ i\leftarrow0$\;
		$\vec{s}\leftarrow\vec{t}\leftarrow\Lambda_{\Root{\Sk}_\sigma}$\;
		$a\leftarrow\Lambda_{\text{ErrorAggravation}}$\;
		\For{$k\leftarrow \Lambda_{\Root{\Sk}}$ \KwTo $\Lambda_{\EE{\Sk}}$}{
			\uIf{$\lnot\IsTwister{k}$}{
				$\vec{u}\leftarrow\solPos{\child{k}}-\solPos{k}$\;
				$\vec{v}\leftarrow\postPos{\child{k}}-\postPos{k}$\;
				$d_{su}\leftarrow1-\frac{(1+\vec{s}\cdot\vec{u})}{2}$\;
				$d_{tv}\leftarrow1-\frac{(1+\vec{t}\cdot\vec{v})}{2}$\;
				$\epsilon\leftarrow\epsilon+a^i*\lvert d_{tv}-d_{su}\rvert$\;
				$i\leftarrow i+1,\ \ \ \ \vec{s}\leftarrow\vec{u},\ \ \ \ \vec{t}\leftarrow\vec{v}$\;
			}
		}
		\Return{$\frac{\epsilon}{\Lambda_{\text{PostureNorm}}}$}
	}
\end{algorithm}

\begin{algorithm}
	\caption{SolutionOK\label{alg:SolutionOK}}
	\SetKwInOut{Input}{input}\SetKwInOut{Output}{output}
	\DontPrintSemicolon
	\Input{$\Pi, \Lambda, aux$\tcp*[f]{Parameters, Hyperparameters}}
	\Output{Is $aux_\Theta$ acceptable (Boolean)}
	\Begin{
		\CalculateSolutionError{$aux_\Theta, \Pi_{\tau}, \Pi_\Psi, \Lambda_\phi$}\;
		\uIf{$aux_{\Theta_{\varepsilon}} \leq aux_{\text{best}\Theta_{\varepsilon}}$}{
			$aux_{\text{best}\Theta} \leftarrow aux_{\Theta}$\;
		}
		\Return{$aux_{\Theta_{\varepsilon}} \leq \Lambda_{\text{Threshold}\varepsilon}$}
	}
\end{algorithm}

\begin{algorithm}
	\caption{ForwardPhase\label{alg:ForwardPhase}}
	\SetKwInOut{Input}{input}\SetKwInOut{Output}{output}

	\DontPrintSemicolon
	\Input{$k, \tau, \Psi, \Theta, \Lambda$\tcc*[l]{{\small Joint, Target Orientation, Target Posture, (aux) Solution, HypParams}}}
	\Output{Intermediate solution $\Theta$ after forward phase}
	\Begin{
		\lIf{$\isEE{k}$}{$\solPos{k} \leftarrow \postPos{k}$}
		\uIf{\textbf{not} $\isRoot{k}$}{
			$\vec{s}\leftarrow\RotVQ{\segment{k}}{\solGO{k}}$\\
			$\TBasis{k}{\QVRot{\vec{s}}{\postPos{k}-\postPos{\parent{k}}}}$\\
			$\vec{r_p}\leftarrow\RotVQ{\PitchRA{k}}{\postBasis{k}}$\\
			$\vec{r_s}\leftarrow\RotVQ{\PitchRA{k}}{\solGO{k}}$\\
			$\theta\leftarrow\VAngleRA{\proj{\vec{r_s},\vec{\JointDir{k}}}}{\proj{\vec{r_p},\vec{\JointDir{k}}}}{\vec{\JointDir{k}}}$\\
			$\TBasisRoll{k}{\QAA{\uvecY}{\theta}}$\\
		}
		\uIf{\textbf{not}\ $\IsTwister{k}$}{
			$\vec{s}\leftarrow\postPos{\child{k}}-\postPos{k}$\\
			$\vec{r_p}\leftarrow\vec{0}$,\ \ \ $n\leftarrow k$,\ \ \ $flipped\leftarrow False$\\
			\uWhile{$\|\vec{r_p}\|\approx 0$ \textbf{and} $n\neq\varnothing$}{
				$m\leftarrow n$\\
				$\vec{p}\leftarrow\segment{n}\ \textbf{if}\ \isRoot{n}\ \textbf{else}\ (\postPos{n}-\postPos{\parent{n}})$\\
				$\vec{r_p}\leftarrow\hat{p}\times \hat{s}$\\
				\uIf{$\isRoot{n}$\ \textbf{and}\ \textbf{not}$flipped$}{
					$\vec{s}\leftarrow\postPos{\child{\child{k}}}-\postPos{\child{k}}$\\
					$flipped\leftarrow True,$\ \ \ $n\leftarrow\child{n}$\\
				}\lElse{
					$\vec{s}\leftarrow\vec{p},$\ \ \ $n\leftarrow\parent{n}$
				}			
			}
			$\vec{s}\small{\leftarrow}\RotVQ{\PitchRA{m}}{I\ \textbf{if}\ \isRoot{m} \textbf{else}\ \solGO{\parent{m}}}$\\
			$\vec{p}\leftarrow\RotVQ{\PitchRA{m}}{\solGO{k}}$\\
			\lIf{$(\vec{p}\cdot \vec{r_p} <0)\neq (\vec{r_p}\cdot \vec{s} <0) $}{
				$\vec{r_p}\leftarrow-\vec{r_p}$
			}
			$\theta\leftarrow\VAngleRA{\proj{\vec{s},\vec{\JointDir{m}}}}{\proj{\vec{r_p},\vec{\JointDir{m}}}}{\vec{\JointDir{m}}}$\\
			\PropagateRollDown{$\QAA{\uvecY}{\theta}, k, \Theta$}
		}
		\uIf{\textbf{not}\ $\isEE{k}$}{
			$\vec{a}\leftarrow\vec{\RA{k}}$\ \textbf{if} $\IsTwister{\child{k}}$\ \textbf{else} $\vec{\RA{\child{k}}}$\\
			$\vec{r}\leftarrow\RotVQ{\vec{a}}{\solGO{k}}$\\
			$\vec{r_c}\leftarrow\RotVQ{\vec{a}}{\solGO{\child{k}}}$\\
			\uIf{$\IsTwister{\child{k}}$\ \textbf{and}\ $\vec{r_c}\cdot \vec{r} < 0$}{
				\PropagateRollUp{$\QAA{\uvecY}{\pi}, k, \Theta, True$}
			}\textbf{else} \uIf{$\|\proj{\vec{r_c}}{\vec{\JointDir{k}}}\|\approx 0$\ \textbf{or}\ $\|\proj{\vec{r}}{\vec{\JointDir{k}}}\|\approx 0$}{
				$\theta\leftarrow\VAngleRA{\proj{\vec{r_c}}{\vec{\JointDir{k}}}}{\proj{\vec{r}}{\vec{\JointDir{k}}}}{\vec{\JointDir{k}}}$\\
				\PropagateRollUp{$\QAA{\uvecY}{\theta}, k, \Theta, \theta\approx\pi$}
			}
		}
		$Q_y, Q_p, Q_r \leftarrow YPR(\QDiff{\solGO{k}}{\tau}, \RA{k})$\\
		\uIf{$\isEE{k}$\ \textbf{and}\ $Q_{r_\theta}>\frac{\pi}{2}$}{
			$Q_r \leftarrow \QAA{\vec{Q_{r_v}}}{\pi-Q_{r_\theta}}$
		}\uElseIf{$\isEE{k}$\ \textbf{and}\ $Q_{r_\theta}<-\frac{\pi}{2}$}{
			$Q_r \leftarrow \QAA{\vec{Q_{r_v}}}{-\pi-Q_{r_\theta}}$
		}
		\uIf{$\IsTwister{k}$}{
			$\solAng{k} \leftarrow\ $\SafeTwist{$k, Q_{y_\theta}+Q_{r_\theta}$}
		}
		\uElse{
			$\solAng{k} \leftarrow\ $\LocalSwing{$k, \RotVQ{\vec{\tau}}{\Theta_{k_{Q^{*}}}}$}\\
			\lIf{$Q_r\neq \textbf{I}$}{\PropagateRollDown{$Q_r, k, \Theta$}}
		}
		\Return{\FinishForward{$k, \Theta, \Lambda$}}
	}
\end{algorithm}

\begin{algorithm}
	\caption{PropagateRollDown\label{alg:PropagateRollDown}}
	\SetKwInOut{Input}{input}\SetKwInOut{Output}{output}
	\DontPrintSemicolon
	\Input{$Q, k, \Theta$\tcp*[f]{Roll, Start link, Solution}}
	\Output{Solution $\Theta$ with Roll propagated down the kinematic chain.}
	\Begin{
		\Repeat{$\isRoot{k}$}{
			$k\leftarrow\parent{k}$\\
			$\solGO{k}\leftarrow\solGO{k}\cdot Q$\\
		}
		\Return{$\Theta$}
	}
\end{algorithm}

\begin{algorithm}
	\caption{PropagateRollUp\label{alg:PropagateRollUp}}
	\SetKwInOut{Input}{input}\SetKwInOut{Output}{output}
	\DontPrintSemicolon
	\Input{$Q, k, \Theta, bFlip=False$}
	\Output{Solution $\Theta$ with Roll propagated up the kinematic chain, with pitch angles flipped if $bFlip=True$.}
	\Begin{
		\While{$\lnot\isEE{k}$}{
			$\solGO{\child{k}}\leftarrow\solGO{\child{k}}\cdot Q$\\
			\uIf{$bFlipPitch$\ \textbf{and not} $\IsTwister{\child{k}}$}{
				$\solAng{\child{k}}\leftarrow-\solAng{\child{k}}$
			}
			$k\leftarrow\child{k}$\\
		}
		\Return{$\Theta$}
	}
\end{algorithm}

\begin{algorithm}
	\caption{YPR\label{alg:YPR}}
	\SetKwInOut{Input}{input}\SetKwInOut{Output}{output}
	\DontPrintSemicolon
	\Input{$Q, \vec{R}$\tcp*[f]{Quat orientation, Rotation axis}}
	\Output{$Q_y, Q_p, Q_r$\ such that $Q=Q_y\cdot Q_p\cdot Q_r$}
	\Begin{
		$\vec{u}\leftarrow\uvecY,$\ \ \ $\vec{x}\leftarrow\vec{R}$\\
		$\vec{y_Q}\leftarrow\vec{u}\cdot Q_M$\\
		$\vec{x_Q}\leftarrow\vec{x}\cdot Q_M$\\
		$\vec{N}\leftarrow\vec{u}\times \vec{y_Q}$\\
		\uIf{$\|\vec{N}\|=0$}{
			\uIf{$\vec{u}\cdot \vec{y_q}\approx 1$\ \textbf{or}\ $\|\vec{y_Q}\|=0$}{
				$Q_y \leftarrow Q_p \leftarrow\textbf{\textit{I}}$\\
				$Q_r \leftarrow \QAA{\vec{u}}{Q_\theta\ \textbf{if}\ \vec{u}\cdot \vec{Q_v}\approx 1\ \textbf{else}\ -Q_\theta}$\\
			}\uElse{
				$Q_y \leftarrow \textbf{\textit{I}}$\\
				$Q_p \leftarrow \QAA{\vec{x}}{-\pi}$\\
				$Q_r \leftarrow \QAA{\vec{u}}{\VAngleRA{\vec{x}}{\vec{x_Q}}{\vec{y_Q}}}$\\
			}	
		}\uElse{
			\lIf{$\vec{N}\cdot \vec{x}<0$}{$\vec{N} \leftarrow -vec{N}$}
			$Q_y \leftarrow \QAA{\vec{u}}{\VAngleRA{\vec{x}}{\hat{N}}{\vec{u}}}$\\
			$Q_p \leftarrow \QAA{\vec{x}}{\VAngleRA{\vec{u}}{\vec{y_Q}}{\hat{N}}}$\\
			$Q_r \leftarrow \QAA{\vec{u}}{\VAngleRA{\hat{N}}{\vec{x_Q}}{\vec{y_Q}}}$\\
		}
		\Return{$\EPA{Q_y}{\vec{u}}, \EPA{Q_p}{\vec{x}}, \EPA{Q_r}{\vec{u}}$}
	}
\end{algorithm}

\begin{algorithm}
	\caption{SafeTwist\label{alg:SafeTwist}}
	\SetKwInOut{Input}{input}\SetKwInOut{Output}{output}
	\DontPrintSemicolon
	\Input{$k, \theta$\tcp*[f]{Joint, Angle}}
	\Output{Safe $\theta$ as cyclic local twist for \erik's Joint Model}
	\Begin{
		$\theta'\leftarrow\SafeAngle{k, \theta, bCycle\leftarrow True}$\;
		$\beta \leftarrow (\theta-\theta') \bmod \pi$\;
		\uIf{$\lvert\beta\rvert>0$ \textbf{and}\ $\lnot\isEE{k}$}{
			\lIf{$\theta\leq k_{\text{Min}\theta}$}{
				$\theta'\leftarrow-k_{\textit{Min}\theta}+\beta$
			}\lElseIf{$\theta\geq k_{\text{Max}\theta}$}{
				$\theta'\leftarrow-k_{\textit{Max}\theta}+\beta$
			}
		}
		\Return{$\theta'$}
	}
\end{algorithm}

\begin{algorithm}
	\caption{LocalSwing\label{alg:LocalSwing}}
	\SetKwInOut{Input}{input}\SetKwInOut{Output}{output}
	\DontPrintSemicolon
	\Input{$k, \vec{\tau}$\tcp*[f]{Joint, Target direction (local)}}
	\Output{$\theta$ local swing angle using \erik's LALUT}
	\Begin{
		$\lambda\leftarrow\TargetLatitude{k}{\vec{\tau}}$\;
		\Return{$\SafeAngle{k}{\LALUT{k}{\lambda}}$}\;
	}
\end{algorithm}

\begin{algorithm}
	\caption{FinishForward\label{alg:FinishForward}}
	\SetKwInOut{Input}{input}\SetKwInOut{Output}{output}
	\DontPrintSemicolon
	\Input{$k, \Theta, \Lambda$\small\tcp*[f]{Link, Partial Solution, Hyperparams}}
	\Output{Solution $\Theta$ after Forward Phase}
	\Begin{
		\uIf{$\Lambda_{\Xi_{\text{AvoidEdges}}}$}{
			$\solAng{k}\leftarrow\AvoidJointEdge{k}{\Lambda_{\text{Disturbance}\theta}}$\;
		}
		\uIf{$\lnot\isEE{k}$}{
			$\solGO{k}\leftarrow\EPA{\QDiff{\solJO{k}}{\solGO{\child{k}}\cdot\solGO{k}}, \RA{k}}$\;
			$\solPos{k}\leftarrow\solPos{\child{k}}-\RotVQ{\segment{k}}{\solGO{\child{k}}}$\;
			$\Theta\leftarrow\ApplyFK{\Theta, k}$\;
		}
	}
\end{algorithm}

\begin{algorithm}
	\caption{BackwardPhase\label{alg:BackwardPhase}}
	\SetKwInOut{Input}{input}\SetKwInOut{Output}{output}
	\DontPrintSemicolon
	\Input{$k, \tau, \Theta, \Lambda$\tcc*[f]{Joint, Target Orientation, (aux)~Solution, Hyperparameters}}
	\Output{(Final) Solution $\Theta$ after backward phase}
	\Begin{
		$\Theta\leftarrow$\SetFrameFromParent{$k, \Theta$}\;
		\uIf{$\IsTwister{k}$}{
			\uIf{$\lnot\isEE{k}$}{
				$\Theta\leftarrow$\BackwardChildRoll{$k, \Theta, \Lambda$}\;
			}
			$\theta\leftarrow$\BackwardLocalTwist{$\tau_M, \solJOmat{k}$}$+\solAng{k}$\;
			\uIf{$\isEE{k}$\textbf{and}\ $\Lambda_{\Xi_{\text{SymmetricEndpoint}}}$}{
				$\theta'\leftarrow$\BackwardLocalTwist{$\tau_M\cdot\QAA{\uvecY}{\pi}, \solJOmat{k}$}$+\solAng{k}$\;
				\lIf{$\lvert\theta'\rvert < \lvert\theta\rvert$}{
					$\theta\leftarrow\theta'$
				}
			}
			$\solAng{k}\leftarrow\theta$
		}\uElse{
			$Q\leftarrow\Theta_{k_{Q^*}}\cdot \tau$\;
			$\lambda\leftarrow\TargetLatitude{k}{\vec{Q_y}}$\;
			$\theta\leftarrow\SafeAngleit{k}{\LALUT{k}{\lambda}}$\;
			\uIf{$\lnot\isEE{k}$}{
				$\theta'\leftarrow\SafeAngleit{k}{\LALUTit{k}{-\lambda}}$\;
				$\vec{\sigma}\leftarrow\RotVQ{\segment{k}}{\solGO{k}\cdot\QAA{\RA{k}}{\theta}}$\;
				$\vec{\sigma'}\leftarrow\RotVQ{\segment{k}}{\solGO{k}\cdot\QAA{\RA{k}}{\theta'}}$\;
				$\vec{d}\leftarrow\RotVQ{\segment{k}}{\solJO{\child{k}}}$\;
				\uIf{$\vec{\sigma'}\cdot\vec{d}>\vec{\sigma}\cdot\vec{d}$}{
					$\swap{\theta}{\theta'}$\;
					$\swap{\sigma}{\sigma'}$\;
				}
				$\vec{d}\leftarrow\RotVQ{\segment{k}}{\tau}$\;
				\uIf{$\vec{\sigma'}\cdot\vec{d}>\vec{\sigma}\cdot\vec{d}$}{
					$\theta\leftarrow\theta'$
				}
			}
			$\solAng{k}\leftarrow\theta$\;
			\uIf{$\lnot\isRoot{k}$ \textbf{and}\ $\IsTwister{\parent{k}}$}{
				$\vec{r_t}\leftarrow\RotVQ{\RA{k}}{\tau}$\;
				$\vec{r_k}\leftarrow\RotVQ{\RA{k}}{\solGO{k}}$\;
				$\vec{r_p}\leftarrow\RotVQ{\RA{\parent{k}}}{\solGO{\parent{k}}}$\;
				\uIf{$\lvert\vec{r_t}\cdot\vec{r_p}\rvert=1$ \textbf{or}\ $\lvert\vec{r_k}\cdot\vec{r_p}\rvert=1$}{
					$\vec{r_t}\leftarrow\RotVQ{\OA{k}}{\tau}$\;
					$\vec{r_k}\leftarrow\RotVQ{\OA{k}}{\solGO{k}}$\;
				}
				$\vec{p}\leftarrow\proj{\vec{r_t}}{\vec{r_p}}$\;
				\uIf{$\lnot\isEE{k}$}{
					$q\leftarrow\QAA{\solJD{\child{k}}}{\VAngleRA{\vec{r_t}}{\vec{p}}{\solJD{\child{k}}}}$\;
					$\vec{p}\leftarrow\proj{\RotVQ{\RA{k}}{q\cdot\tau}}{\vec{r_p}}$\;
				}
				$\gamma\leftarrow\VAngleRA{\proj{\vec{r_k}}{\vec{r_p}}}{\vec{p}}{\vec{r_p}}$\;
				$\solAng{\parent{k}}\leftarrow\gamma+\solAng{\parent{k}}$\;
				$\Theta\leftarrow$\SetFrameFromParent{$k, \Theta$}\;	
			}
		}
		\Return{\FinishBackward{$k, \Theta$}}
	}
\end{algorithm}

\begin{algorithm}
	\caption{BackwardChildRoll\label{alg:BackwardChildRoll}}
	\SetKwInOut{Input}{input}\SetKwInOut{Output}{output}
	\DontPrintSemicolon
	\Input{$k, \Theta, \Lambda$}
	\Output{$\Theta$ with $\Theta_{\RA{\child{k}}}$ aligned with plane $\perp$ to $\solJD{\child{k}}$ }
	\Begin{
		\If{$\lnot\RotVQ{\RA{\child{k}}}{\solGO{\child{k}}}\cdot \Theta_{k_{Q_y}} = 0$}{
			$\vec{c_D} \leftarrow \solJD{\child{k}}$\ \ \ $c_B \leftarrow \solGOmat{\child{k}}$\;
			$axis \leftarrow $'y'\;
			\uIf{$c_{B_{y}}\cdot \vec{c_D} \approx 1$ \textbf{or}\ $c_{B_{y}}\cdot \vec{c_D} \approx -1$ \textbf{or}\ \\ $\solJD{k}\cdot \vec{c_D} \approx 1$ \textbf{or}\ $\solJD{k}\cdot \vec{c_D} \approx -1$}{
				\lIf{$c_{B_{z}}\cdot \vec{c_D} \approx 1$ \textbf{or}\ $c_{B_{z}}\cdot \vec{c_D} \approx -1$ \textbf{or}\ $c_{B_{x}}\cdot \solJD{k} \approx 0$}{
					$axis \leftarrow $'x'
				}\lElse{
					$axis \leftarrow $'z'
				}
			}
			$\vec{c_v}\leftarrow c_{B_{axis}},$\ \ \ $\vec{p_v}\leftarrow\Theta_{k_{Q_{axis}}}$\;
			$\vec{c_p}\leftarrow\proj{\vec{c_v}}{\vec{c_D}},$\ \ \ $\vec{p_p}\leftarrow\proj{\vec{p_v}}{\vec{c_D}}$\;
			$q\leftarrow\QAA{\vec{c_D}}{\VAngleRA{\vec{c_p}}{\vec{p_p}}{\vec{c_D}}} \cdot \solGO{\child{k}}$\;
			$q'\leftarrow\QAA{\vec{c_D}}{\VAngleRA{\vec{c_p}}{\vec{p_p}}{-\vec{c_D}}} \cdot \solGO{\child{k}}$\;
			\uIf{$q_{axis}\cdot \vec{p_v} < q'_{axis}\cdot \vec{p_v}$}{
				$\solGO{\child{k}}\leftarrow q'$
			}\uElse{
				$\solGO{\child{k}} \leftarrow q$
			}
		}
		\Return{$\Theta$}
	}
\end{algorithm}

\begin{algorithm}
	\caption{BWTwist\label{alg:BWTwist}}
	\SetKwInOut{Input}{input}\SetKwInOut{Output}{output}
	\DontPrintSemicolon
	\Input{$\tau, \Omega$\hspace{-3pt}\tcp*[]{\hspace{-5pt}Target, Current Orientation}}
	\Output{Angle the joint should twist to achieve the given target $\tau$ based on its current orientation $\Omega$}
	\Begin{
		$d_{zy}\leftarrow\tau_z\cdot \Omega_y$ \ \ \ 
		$d_{xy}\leftarrow\tau_x\cdot \Omega_y$\; 
		\uIf{$d_{zy}\approx 1$ \textbf{or}\ $d_{zy}\approx -1$ \textbf{or}\ $\lvert d_{xy}\rvert <\lvert d_{zy}\rvert$}{
			\Return{\VAngleRA{$\Omega_x$}{$\proj{\tau_x}{\Omega_y}$}{$\Omega_y$}}
		}\uElse{
			\Return{return \VAngleRA{$\Omega_z$}{$\proj{\tau_z}{\Omega_y}$}{$\Omega_y$}}
		}
	}
\end{algorithm}

\begin{algorithm}
	\caption{FinishBackward\label{alg:FinishBackward}}
	\SetKwInOut{Input}{input}\SetKwInOut{Output}{output}
	\DontPrintSemicolon
	\Input{$k, \Theta$\tcp*[f]{Link, Partial Solution}}
	\Output{Solution $\Theta$ after BackwardPhase}
	\Begin{
		\uIf{$\lnot\isRoot{k}$ \textbf{and}\ $\IsTwister{\parent{k}}$}{
			$\Theta\leftarrow$\FinishBackward{$\parent{k}, \Theta$}\;
		}
		$\Theta\leftarrow\SetOriFromParent{k}{\Theta}$\;
	}
\end{algorithm}

\begin{algorithm}
	\caption{BWCD\_Posture\label{alg:BWCDPosture}}
	\SetKwInOut{Input}{input}\SetKwInOut{Output}{output}
	\DontPrintSemicolon
	\Input{$\Psi, \vec{\tau}, \Lambda$\tcp*[]{\small Posture to solve, Target Direction, Hyperparameters}}
	\Output{$\Psi$ result of the BWCD algorithm as a new posture.}
	\Begin{
		\For{$i\leftarrow 1$ \KwTo $\Lambda_{\text{MaxCCDIterations}}$}{
			\For{$k\leftarrow \Lambda_{\Root{\Sk}}$ \KwTo $\skEE$}{
				$\vec{pd} \leftarrow \RotVQ{\segmentNorm{\skEE}}{\postBasis{\text{Superpoint}}}$\;
				\If{\CCDTest{$\vec{\tau}, \vec{pd}, \Lambda$}}{
					\Return{$\Psi$}
				}
				$\vec{r} \leftarrow \RotVQ{\RA{k}}{\postBasis{k}}$\;
				$\vec{pdp}\leftarrow \proj{\vec{pd}}{\vec{r}}$\;
				$\vec{tdp}\leftarrow \proj{\vec{\tau}}{\vec{r}}$\;
				\If{$\norm{\vec{pdp}} \neq 0\ \textbf{and}\ \norm{\vec{tdp}} \neq 0$}{
					$\alpha\leftarrow \VAngleRA{\vec{pdp}}{\vec{tdp}}{\vec{r}}$\;
					\If{$\round{\abs{\alpha}}>0$}{
						$q\leftarrow \EPA{\QAA{\vec{r}}{\alpha}}{\RA{k}}$\;
						$\vec{p}\leftarrow \postPos{k}$\;
						$\postAng{k}\leftarrow \postAng{k} + \alpha$\;
						\For{$j\leftarrow k$ \KwTo $\skEE$}{
							$\postPos{j}\leftarrow \vec{p}$\;
							$c\leftarrow \child{j}\ \textbf{if not}\ \isEE{j}\ \textbf{else}\ \Superpoint{\Psi}$\;
							$\postBasis{c}\leftarrow q\cdot\postBasis{c}$\;
							\uIf{$\isEE{j}$}{
								$\postPos{\text{Superpoint}}\leftarrow \vec{p} + \RotVQ{\segment{j}}{\postBasis{c}}$\;
							}\uElse{
								$\vec{p}\leftarrow \vec{p} + \RotVQ{\segment{j}}{\postBasis{c}}$\;
							}
						}
					}
				}
			}
			$\vec{pd}\leftarrow \RotVQ{\segmentNorm{\skEE}}{\postBasis{\text{Superpoint}}}$\;
			\If{\CCDTest{$\vec{\tau}, \vec{pd}, \Lambda$}}{
				\Return{$\Psi$}
			}
		}
		\Return{$\Psi$}
	}
\end{algorithm}

\begin{algorithm}
	\caption{CCDTest\label{alg:CCDTest}}
	\SetKwInOut{Input}{input}\SetKwInOut{Output}{output}
	\DontPrintSemicolon
	\Input{$\vec{t}, \vec{ee}, \Lambda, \text{returnError=False}$\tcp*[]{\small Joint, Target, EndEffector, Hyperparameters, (optional)}}
	\Output{Current $\vec{ee}$ is an acceptable solution for a CCD-based algorithm.}
	\Begin{
		$\epsilon\leftarrow \round{-\frac{\vec{t}\cdot \vec{ee}-1}{2}}$\;
		\lIf{\text{returnError}}{
			\Return{$\epsilon \leq \Lambda_{\text{CCDPrecision}}, \epsilon$}
		}
		\Return{$\epsilon \leq \Lambda_{\text{CCDPrecision}}$}
	}
\end{algorithm}

\begin{algorithm}
	\caption{BWCD\_Solution\label{alg:BWCDSolution}}
	\SetKwInOut{Input}{input}\SetKwInOut{Output}{output}
	\DontPrintSemicolon
	\Input{$\Theta, \tau, \Lambda$\tcp*[]{\small Solution to solve, Target Direction, Hyperparameters}}
	\Output{$\Theta'$ result of the BWCD algorithm as a new solution.}
	\Begin{
		\For{$i\leftarrow 1$ \KwTo $\Lambda_{\text{MaxCCDIterations}}$}{
			$\vec{eod}\leftarrow \solJD{\text{EE}}$\;
			\For{$k\leftarrow \Lambda_{\Root{\Sk}}$ \KwTo $\skEE$}{
				\If{\CCDTest{$\vec{\tau}, \vec{eod}, \Lambda$}}{
					$\solAng{\text{EE}}\leftarrow\solAng{\text{EE}}+$\BackwardLocalTwist{$\tau_M, \solJOmat{\text{EE}}$}\;
					\Return{$\Theta$}
				}
				$\vec{r}\leftarrow \RotVQ{\RA{k}}{\solJO{k}}$\;
				$\vec{top}\leftarrow \proj{\vec{\tau}}{\vec{r}}$\;
				$\vec{eop}\leftarrow \proj{eod}{\vec{r}}$\;
				\If{$\norm{\vec{\vec{top}}} \neq 0\ \textbf{and}\ \norm{\vec{\vec{eop}}} \neq 0$}{
					$\solAng{k}\leftarrow \solAng{k}+\VAngleRA{\vec{eop}}{\vec{top}}{\vec{r}}$\;
					$\ApplyFK{\Theta}{k}$\;
					$\vec{eod}\leftarrow \solJD{\text{EE}}$\;
				}
			}
			\If{\CCDTest{$\vec{\tau}, \vec{eod}, \Lambda$}}{
				$\solAng{\text{EE}}\leftarrow\solAng{\text{EE}}+$\BackwardLocalTwist{$\tau_M, \solJOmat{\text{EE}}$}\;
				\Return{$\Theta$}
			}
		}
		$\solAng{\text{EE}}\leftarrow\solAng{\text{EE}}+$\BackwardLocalTwist{$\tau_M, \solJOmat{\text{EE}}$}\;
		\Return{$\Theta$}
	}
\end{algorithm}

\begin{algorithm}
	\caption{CCD\label{alg:CCD}}
	\SetKwInOut{Input}{input}\SetKwInOut{Output}{output}
	\DontPrintSemicolon
	\Input{$\Theta, \tau, \Lambda$\tcp*[]{\small Solution to solve, Target Orientation, Hyperparameters}}
	\Output{$\Theta$ result of the CCD algorithm as a new solution.}
	\Begin{
		$\epsilon \leftarrow 10000, \text{last}_\epsilon \leftarrow 0$\;
		$\vec{td}\leftarrow \vec{\tau}$\;
		$\vec{ed}\leftarrow \Theta_{\text{EE}_{\vec{d}}}$\;
		\For{$i\leftarrow 1$ \KwTo $\Lambda_{\text{MaxCCDIterations}}$}{
			$\text{last}_\epsilon \leftarrow \epsilon$\;
			\For{$k\leftarrow \skEE$ \KwTo $\Lambda_{\Root{\Sk}}$}{
				$\text{ok}, \epsilon \leftarrow$ \CCDTest{$\vec{td}, \vec{ed}, \Lambda, returnError=True$} \;
				\If{ok}{
					$\solAng{\text{EE}}\leftarrow\solAng{\text{EE}}+$\BackwardLocalTwist{$\tau_M, \solJOmat{\text{EE}}$}\;
					\Return{$\Theta$}
				}
				$r \leftarrow \RotVQ{\RA{k}}{\solGO{k}}$\;
				$\vec{tdp} \leftarrow \proj{\vec{td}}{r}$\;
				$\vec{edp} \leftarrow \proj{\vec{ed}}{r}$\;
				\If{$\norm{\vec{tdp}} \neq 0\ \textbf{and}\ \norm{\vec{edp}} \neq 0$}{
					$\solAng{k} \leftarrow \solAng{k} + \VAngleRA{\vec{edp}}{\vec{tdp}}{r}$\;
					\uIf{$\Lambda_{\Xi_{\text{AvoidJointEdges}}}$}{
						$\AvoidPostureJointEdges{\Theta}{\Lambda_{\text{Disturbance}\theta}}$\;
					}
					$\ApplyFK{k}$\;
					$\vec{ed}\leftarrow \Theta_{\text{EE}_{\vec{d}}}$\;
				}
			}
			$ok, \epsilon \leftarrow$ \CCDTest{$\vec{td}, \vec{ed}, \Lambda, returnError=True$}\;
			\If{ok}{
				$\solAng{\text{EE}}\leftarrow\solAng{\text{EE}}+$\BackwardLocalTwist{$\tau_M, \solJOmat{\text{EE}}$}\;
				\Return{$\Theta$}
			}
			\If{$\abs{\epsilon-\text{last}_\epsilon} \leq \Lambda_{Precision}$}{
				$\solAng{\text{EE}}\leftarrow\solAng{\text{EE}}+$\BackwardLocalTwist{$\tau_M, \solJOmat{\text{EE}}$}\;
				\Return{$\Theta$}
			}
		}
		$\solAng{\text{EE}}\leftarrow\solAng{\text{EE}}+$\BackwardLocalTwist{$\tau_M, \solJOmat{\text{EE}}$}\;
		\Return{$\Theta$}
	}
\end{algorithm}

\end{document}